\documentclass[a4paper,fleqn]{cas-sc}

\usepackage{amsmath}
\usepackage{graphicx}
\usepackage{rotating}
\usepackage{acronym}
\usepackage{enumitem}
\usepackage{caption}
\usepackage{comment}

\usepackage{multirow}

\usepackage{xspace}

\usepackage{booktabs}

\usepackage{subcaption}
\usepackage{svg}

\usepackage{placeins} 
\renewcommand*{\eqref}[1]{Eq.~\hyperref[#1]{\ref*{#1}}}
\newcommand*{\chref}[1]{Chapter~\hyperref[#1]{\ref*{#1}}}
\newcommand*{\secref}[1]{Sec.~\hyperref[#1]{\ref*{#1}}}
\newcommand*{\sectionref}[1]{Section~\hyperref[#1]{\ref*{#1}}}
\newcommand*{\figref}[1]{Fig.~\hyperref[#1]{\ref*{#1}}}
\newcommand*{\figureref}[1]{Figure~\hyperref[#1]{\ref*{#1}}}
\renewcommand*{\tabref}[1]{Tab.~\hyperref[#1]{\ref*{#1}}}
\newcommand*{\tableref}[1]{Table~\hyperref[#1]{\ref*{#1}}}
\newcommand*{\Subref}[1]{\hyperref[#1]{(\subref*{#1})}}
\newcommand*{\algref}[1]{Alg.~\hyperref[#1]{\ref*{#1}}}
\newcommand*{\algorithmref}[1]{Algorithm~\hyperref[#1]{\ref*{#1}}}
\newcommand*{\appendref}[1]{Appendix~\hyperref[#1]{\ref*{#1}}}

\newcolumntype{H}{>{\setbox0=\hbox\bgroup}c<{\egroup}@{}}

\newcommand{\dsname}{P$^3$~}

\makeatletter
\newcommand{\definetrim}[2]{%
  \define@key{Gin}{#1}[]{\setkeys{Gin}{trim=#2,clip}}%
}
\makeatother

\makeatletter
\newcommand{\definescale}[2]{%
  \define@key{Gin}{#1}[]{\setkeys{Gin}{scale=#2}}%
}
\makeatother

\makeatletter
\newcommand{\gridon}[2]{%
  \define@key{Gin}{#1}[]{\setkeys{Gin}{#2}}%
}
\makeatother

\newcommand{\etal}{\textit{et~al}\@\xspace}
\newcommand{\ie}{\textit{i.e.}\@\xspace}
\newcommand{\eg}{\textit{e.g.}\@\xspace}

\usepackage[authoryear]{natbib}

\def\tsc#1{\csdef{#1}{\textsc{\lowercase{#1}}\xspace}}
\tsc{WGM}
\tsc{QE}

\let\printorcid\relax
\begin{document}


\newacro{2d}[$2$D]{two-dimensional}
\newacro{3d}[$3$D]{three-dimensional}
\newacro{2dt}[$2$DT]{2D Delaunay triangulation}
\newacro{3dt}[$3$DT]{3D Delaunay tetrahedralisation}
\newacro{dt}[DT]{Delaunay triangulation}

\newacro{iou}[IoU]{intersection over union}
\newacro{cd}[CD]{Chamfer distance}
\newacro{nc}[NC]{normal consistency}
\newacro{noc}[NoC]{number of components}

\newacro{pca}[PCA]{principal component analysis}
\newacro{mlp}[MLP]{multilayer perceptron}

\newacro{bce}[BCE]{binary cross entropy}
\newacro{mse}[MSE]{mean square error}

\newacro{gsd}[GSD]{ground sampling distance}
\newacro{lidar}[LiDAR]{light detection and ranging}

\newacro{cnn}[CNN]{convolutional neural network}
\newacroplural{cnn}[CNNs]{convolutional neural networks}

\newacro{gnn}[GNN]{graph neural network}
\newacroplural{gnn}[GNNs]{graph neural networks}

\newacro{lod}[LOD]{level of detail}
\newacroplural{lod}[LODs]{levels of detail}
\newacro{mvs}[MVS]{multi-view stereo}
\newacro{sfm}[SfM]{structure from motion}
\newacro{lidar}[LiDAR]{Light Detection and Ranging}
\newacro{als}[ALS]{airborne laser scanning}

\newacro{nerf}[NeRF]{neural radiance field}

\newacro{sdf}[SDF]{signed distance function}
\newacro{of}[OF]{occupancy function}

\newacro{nma}[NMA]{National Mapping Agency}
\newacroplural{nma}[NMAs]{National Mapping Agencies}

\newacro{sota}[SOTA]{state-of-the-art}

\newacro{vit}[ViT]{Vision Transformer}

\newacro{ffl}[FFL]{Frame Field Learning}

\newacro{dof}[DoF]{degree of freedom}
\newacro{mta}[MTA]{maximum tangent angle}

\newacro{sm}[SM]{supplementary material}
\newacro{dsm}[DSM]{digital surface model}

\let\WriteBookmarks\relax
\def\floatpagepagefraction{1}
\def\textpagefraction{.001}

\shorttitle{The P$^3$ Dataset}    

\shortauthors{R. Sulzer et~al.}  

\title [mode = title]{Pixels, Points and Polygons:\\A Dataset and Benchmark for Multimodal Building Vectorization}  



%

\author[1,2]{Raphael Sulzer}
\cormark[1]




\affiliation[1]{organization={Université Côte d’Azur, INRIA},
                city={Sophia-Antipolis},
                country={France}}

\author[2]{Liuyun Duan}

\author[2]{Nicolas Girard}

\affiliation[2]{organization={LuxCarta Technology},
                city={Mouans-Sartoux},
                country={France}}

\cortext[1]{Corresponding author}

\author[1]{Florent Lafarge}


\begin{abstract}
Accurate and up-to-date building polygon maps are essential for urban planning, disaster response, and large-scale geospatial analysis, yet creating them automatically across diverse regions remains challenging. To address this challenge,
we present the \dsname dataset, a large-scale multimodal benchmark for building vectorization, constructed from aerial LiDAR point clouds, aerial images, and vectorized 2D building outlines, collected across three continents. The dataset contains over 10 billion LiDAR points with decimeter-level accuracy and RGB images at a ground sampling distance of 25 centimeters.
While many existing datasets primarily focus on the image modality, \dsname offers a complementary perspective by also incorporating dense 3D information.
We demonstrate that LiDAR point clouds serve as a robust modality for predicting building polygons, both in hybrid and end-to-end learning frameworks. Moreover, fusing aerial LiDAR and imagery further improves accuracy and geometric quality of predicted polygons. 
The \dsname dataset is publicly available, along with code and pretrained weights of three state-of-the-art models for building polygon prediction at \url{https://github.com/raphaelsulzer/PixelsPointsPolygons}.
\end{abstract}


\maketitle

\section{Introduction}

Cadastral maps that register the footprints of buildings as vector representations -- typically 2D polygons -- constitute an important source of information in numerous application fields, going from urban planning and navigation to environmental monitoring and disaster response through defense and intelligence. In the past decades, the construction and updates of these maps have been done mainly by human operators at some specific locations in the world only. Today, this tedious task is about to be replaced by automatic algorithms that learn to capture building outlines from satellite and aerial data, with the promise of vectorizing the several billion buildings in the world despite the strong variability of shape, density, and regularity of these objects. 

Several scientific communities such as Computer Vision, Computer Graphics, Geoscience, Remote Sensing and Photogrammetry have actively worked on the building vectorization problem during the past years. For fully automatic building vectorization, deep networks are commonly trained on datasets 
that (i) exploit the image modality only, (ii) have little architectural and regional variability, as well as little variability in terms of acquisition characteristics \citep{Roof3D2023,lineFit_eccv2024,roofvect_geoinfo,semcity_isprs20,isprs-annals-I-3-293-2012}, and (iii) use pixel masks instead of direct polygon annotations \citep{inria_dataset}.
Unfortunately, these restrictions tend to make the efficiency of recent \ac{sota} methods \citep{hisup,pix2poly,Zorzi_2023_ICCV} stagnant and, in the end, not ready to operate effectively, robustly and at large scale. (i) Using only overhead images can lead to false positive detections or missing building parts due to shadows and occlusions. Occlusions are particularly frequent in the form of trees hiding a building or part of it, or due to "leaning buildings" induced by image distortion. 
(ii) Small variability in the dataset can cause models to overfit to specific architectural styles of a country or region \citep{inria_dataset}, to specific acquisition geometries, or even to a ground truth annotation source, which is often a (semi-)automatic extraction pipeline itself. 
Finally, (iii) using pixel-level supervision only can lead to noisy, overly complex and geometrically inconsistent polygons compared to official cadastral maps \citep{hisup}.

\begin{figure}
	\definetrim{mytrim}{0 0 0 0}
	\newcommand{\mywidth}{0.24\linewidth}
	\newcommand{\myfontsize}{\scriptsize}
    \centering
    \resizebox{0.7\textwidth}{!}{
\includegraphics[width=\linewidth]{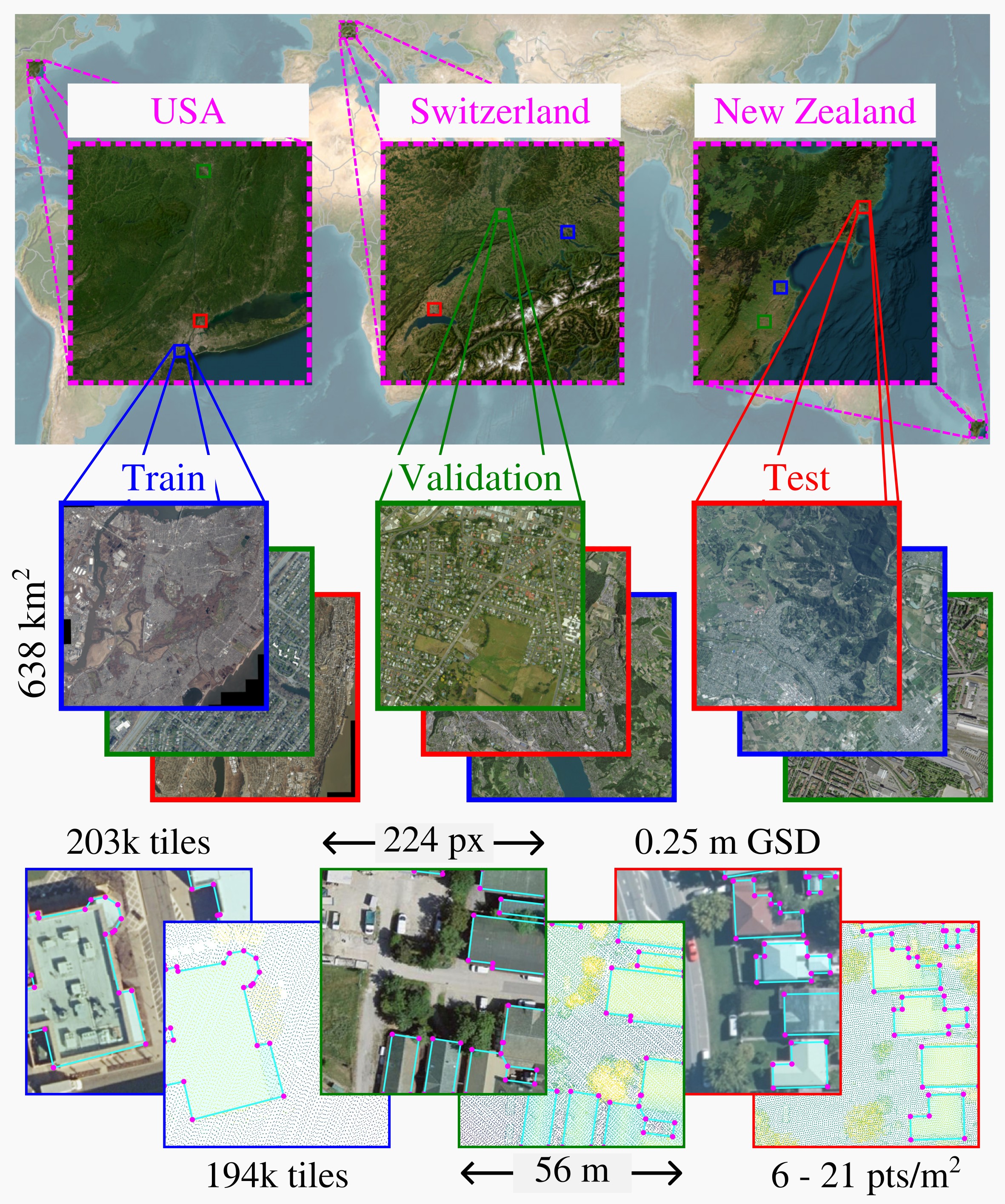} 
	}
\caption{\textbf{Overview of P$^3$.} We collect aerial images, aerial LiDAR point clouds and vectorized building outlines from the USA, Switzerland and New Zealand\protect\footnotemark[1]. We harmonize and tile the data to create a large benchmark dataset for building vectorization.}
\label{fig:map}
\end{figure}

To address these issues, we propose the \dsname dataset, a  
dataset that provides two input modalities from different regions and precise vectorized building outline annotations. 
\dsname includes the traditional image modality with aerial ortho- or near-nadir images, and an aerial LiDAR modality in the form of 
3D point clouds. Aerial LiDAR point clouds become increasingly openly available from \acp{nma} at the scale of entire countries \citep{lidarhd,swiss3d,ahn} and can thus constitute a complementary data source to imagery. In particular, aerial LiDAR point clouds provide direct height information with decimeter-level accuracy \citep{swiss3d}, and their acquisition is resilient to image distortion and meteorological and seasonal effects, as laser pulses can penetrate clouds and vegetation \citep{PEARSE2018257}. 

\footnotetext[1]{Licensed by Toitū Te Whenua Land Information New Zealand for re-use under CC-BY-4.0.} 

A second key specificity of our dataset is its scale and variability. The data cover a total area of 638km² 
across seven different regions on three different continents.
Lastly, the dataset provides ground truth polygon annotations with geometric guarantees. 
In particular, polygons do not overlap with neighboring polygons and are either simple, or contain holes, \eg to represent courtyards.  

In addition to the \dsname dataset, a second contribution of our work is a benchmark that analyzes the impact of the data modalities 
on the building vectorization problem. In particular, we propose a general pipeline to adapt the overpopulated image-based methods to the LiDAR modality and to the combination of image and LiDAR modalities. We conduct experiments with three recent \ac{sota} methods: FFL~\citep{ffl}, HiSup~\citep{hisup} and Pix2Poly~\citep{pix2poly}. 
Besides the traditional accuracy metrics used in the field, 
we also introduce a metric that measures the geometric quality of polygons in terms of symmetry and regularity. This aspect is often ignored in the commonly-used evaluation protocol, but is crucial for 
practitioners who want to exploit the potential of building maps.

Our dataset is designed to facilitate further research on multimodal building vectorization.
It is available to download and can be used directly  with a Python library that implements our benchmark with training, testing, and evaluation of the three building vectorization methods on our dataset.

\section{Related work}
Our review of previous work covers existing datasets for building detection and reconstruction, and methods for detecting and polygonizing building outlines.

\subsection{Related datasets}
\label{sec:rl_datasets}

Datasets for building detection can be grouped into three categories, with either the exploitation of 2D data modality, 3D data modality or a combination of the two. \tableref{tab:datasets} lists the existing datasets for each of these three categories and provides data and annotation specifications for each of them.
\paragraph{2D} datasets that exploit the image modality constitute the dominant categories. Inria \citep{inria_dataset}, CrowdAI \citep{mohanty2020deep}, WHU \citep{whu_dataset} and Shanghai \citep{devglobal2022shanghai} are the most popular datasets used in the literature. Data are typically satellite or aerial images acquired at near-nadir with a 
\ac{gsd} between 5 and 80~cm.
The Inria dataset \citep{inria_dataset} offers good urban diversity with a total area of 810km² covering different urban landscapes and countries, but requires a lossy conversion of the annotated pixel masks to polygons. Only the Microsoft \citep{Microsoft_dataset} and Google \citep{Open_buildings} datasets exhibit higher diversity by covering millions of square kilometers. However, these two datasets provide building polygons with a low precision and without the original images. Other datasets such as WBD \citep{VWB2020}, RoofSat \citep{lineFit_eccv2024} and RoofVect \citep{roofvect_geoinfo} provide a more complete 2D roof wireframe description in the form of planar graphs. These datasets are however small and poorly diversified. Note that several datasets on this list have been artificially enriched by data augmentation, such as CrowdAI \citep{mohanty2020deep}, RoofSat\citep{lineFit_eccv2024} or RoofVect \citep{roofvect_geoinfo}. In the case of the former, this enrichment strategy has been shown to cause 
biases in the results \citep{adimoolam2023efficientdeduplicationleakagedetection}. Note also that some datasets such as OpenCities \citep{gfdrr2020opencities} and SpaceNet2 \citep{spacenet} suffer from low-quality human annotations typically due to either the low resolution of the original data source or a lack of harmonization in between annotators. {Recent work Map2ImLas \citep{Map2ImLas2026} presents a multimodal dataset that aligns orthoimages, DSMs and 3D point clouds. It provides consistent 2D and 3D labels for 20 semantic classes while covering around 217~km² of the Netherlands. 
}

\paragraph{3D.} Building3D \citep{building3D_ICCV23}, City3D\citep{HuangCity3d_2022} and RoofN3D \citep{RoofN3D}, which exploit 3D data, mostly target the 3D building reconstruction problem from LiDAR point clouds \citep{simplicity}. Each point cloud represents a single building and not an urban scene or a part of it. Unfortunately, annotated 3D models that are created by automatic methods or human operators present strong geometric inconsistencies that prevent these datasets from being used effectively for the building footprint extraction task.

\paragraph{2D and 3D.} 
Only a few datasets present 2D and 3D modalities, \ie Roof3D \citep{Roof3D2023}, Potsdam, and Vaihingen \citep{isprs-annals-I-3-293-2012}. 
These datasets are small and exhibit low urban diversity. They also do not offer vectorized annotations, but pixel masks only. For Roof3D and Potsdam, the complementary modality is a \ac{dsm} and provides 2.5D information only. In contrast, our dataset combines both images and 3D LiDAR at larger scale in various urban landscapes. In particular, it is 400 times larger than the Vaihingen dataset and provides 2D outline polygons as annotation.    

\begin{table}
    \caption{\textbf{Related datasets.} Modalities typically include satellite images (SI), aerial images (AI), drone images (DI), airborne LiDAR scans (ALS) and digital suface models (DSM). SI$^*$ specfies datasets that use satellite imagery for their annotation but the imagery itself is not included. 
Tiling specifies whether data is a single building (SB) only, an image tile adapted to deep learning frameworks (T) or a large georeferenced tile bigger than 1000$\times$1000 px (T+). Diversity approximates the urban variability of the dataset with four categories: "- -" for a similar type of buildings from a single location, "-" for several types of buildings from a single location, "+" for several types of buildings from multiple locations in the world, and "++" if it covers at least 1$\%$ of the total urban scenes in the world. "H", "A" and "M" refer to the annotation origin, \ie either from human operators, automatic algorithms or a mix of these two.
}
\setlength{\tabcolsep}{1pt}
\newcommand{\rotangle}{62} 
\newcommand{\rot}[1]{\rotatebox{\rotangle}{#1}}
\centering
\small
\resizebox{\textwidth}{!}{
\begin{tabular}{@{}l@{\hspace{3pt}}l |c c c c c |c@{}c@{}c@{}c@{}c@{}}

&\bf Dataset&\multicolumn{5}{c}{\bf Data specifications}& \multicolumn{5}{c}{\bf Annotation}\\
& & \rot{Modality} & \rot{Ground sampling} & \rot{Area (km²)} & \rot{Tiling} & \rot{Diversity} & \rot{Pixel label} & \rot{2D outline polygon} & \rot{2D roof wireframe} & \rot{3D building wireframe} & \rot{Origin} \\
\hline
\multirow{11}{*}{\rotatebox[origin=c]{90}{\hspace{-18mm}2D}}
&Inria \citep{inria_dataset}              & AI & 0.3m    & 810   &T+& + & \checkmark & & & & M \\
&SemiCity \citep{semcity_isprs20}         & SI & 0.5m    & 50    &T+& - & \checkmark & & & & H \\
&WHU \citep{whu_dataset}              & AI & 0.075m  & 450       &T& - & \checkmark &  & & & H \\
&CrowdAI \citep{mohanty2020deep}          & SI & 0.3m    & 2.3k  &T& - &  & \checkmark & & & H\\
& OpenCities \citep{gfdrr2020opencities} & DI & 0.4-0.8m& 415   &T& + & &\checkmark&&& H \\
& Shanghai \citep{devglobal2022shanghai}  & SI & 0.3m    & 21    &T& - &  & \checkmark & & & H\\
&WHU \citep{whu_dataset}             & SI & 0.45m   & 550       &T& + &  & \checkmark &  & & H \\
&SpaceNet2 \citep{spacenet}               & SI & 0.3m    & 3k    &T& + &  & \checkmark & & & H \\
&UBC \citep{Huang_2022_CVPRW}             & SI & 0.5-0.8m& 66.1  &T& + &  & \checkmark & & & M\\
&Microsoft \citep{Microsoft_dataset}      & SI$^*$ & 0.3-1.0m&N/A&-& ++ & & \checkmark & & &A\\
&Google \citep{Open_buildings}            & SI$^*$ & 0.5m & 58M  &-& ++ &  & \checkmark & & &M\\
&WBD \citep{VWB2020}                      & SI & 0.3m & 32 &SB& - - & & & \checkmark & &H \\
&RoofSat \citep{lineFit_eccv2024}         & SI & 0.3m & 14.9 &T& - & & & \checkmark & &H \\
&RoofVect \citep{roofvect_geoinfo}        & AI & 0.1m & 2 &SB& - &  & & \checkmark& &H\\ \hline
\multirow{3}{*}{\rotatebox[origin=c]{90}{3D}}
&Building3D \citep{building3D_ICCV23} & ALS & 30pts/m² & 998 & SB&+ &  &  & & \checkmark & M \\
&City3D \citep{HuangCity3d_2022} & ALS & 4-50 pts/m² & N/A &SB& + &  &  & & \checkmark & A \\
&RoofN3D \citep{RoofN3D} & ALS & 4.7 pts/m² & 1010 & SB & - &   &  & & \checkmark & A \\ \hline
\multirow{5}{*}{\rotatebox[origin=c]{90}{2D+3D}}
&Roof3D \citep{Roof3D2023} & AI+DSM & 0.3m & 22.4 &T& - &  \checkmark & & & & M \\
&Potsdam \citep{isprs-annals-I-3-293-2012} & AI+DSM & 0.05m & 3.4            &T& - &  \checkmark & & & & H \\
&Vaihingen \citep{isprs-annals-I-3-293-2012} & AI+ALS & 0.08m, 4pts/m² & 1.5 &T& - &  \checkmark & & & & H \\
&Map2ImLas \citep{Map2ImLas2026} & AI+ALS & 0.075m, 10-14pts/m² & 217 &T+& - &  \checkmark & \checkmark & & & A \\
&\textbf{P$^3$ (ours)} & AI+ALS & 0.25m, 16pts/m² & 638                     &T& + & & \checkmark & & &M \\ \hline
\end{tabular}
}
\label{tab:datasets}
\end{table}

\subsection{Building outline vectorization methods}
\label{sec:rl_methods}

Methods for detecting and polygonizing building outlines largely exploit the image modality. They generally fall into two main categories: hybrid methods and direct building polygon prediction. 
\paragraph{Hybrid methods} typically involve multiple stages, often combining segmentation or detection with polygon extraction and refinement steps. Many recent approaches in this category incorporate structured priors, semantics, or preexisting databases. For instance, semantically encriched point clouds can be used to validate and correct existing building databases of \acp{nma} \citep{ign_semantic_footprint}, and elevation information provides geometric primitive clues for building rectangles extraction \citep{ortner2007IJCV}. ASIP \citep{Mu_cvpr2020} splits and merges cells of a polygonal partition according to a deep semantic segmentation map to generate compact polygons. Frame Field Learning \citep{ffl} directly learns the directional fields that align with building boundaries, and obtains more regular polygons for urban scenes. HiSup \citep{hisup} tackles this challenge by integrating directional fields, semantics, and geometry learning with hierarchical supervision to enhance the geometry precision. The sophisticated hierarchical design preserves a good balance between complexity and fidelity. 

\paragraph{Direct building polygon prediction} approaches typically leverage advanced neural network architectures to process aerial or satellite imagery and directly output vectorized building polygons without intermediate raster representations. Popular methods such as PolyMapper \citep{polymapper_2019}], PolyWorld\citep{zorzi2022polyworld}, and Re-PolyWorld \citep{Zorzi_2023_ICCV} directly extract vertices via convolutional or recurrent neural networks and apply a graph neural network to form polygons. PolyBuilding \citep{HU202315} shows promising performance by applying transformers that reduces the vertex redundancy in CNN-GCN paradigm. Based on the success of \citep{topdig}, the image-to-vertex sequence methods \citep{p2pformer,geoformer,pix2poly} represented by Pix2Poly employ a transformer to produce building vertex tokens as sequence, eliminating post-processing steps by using permutation matrices for connections.  Other methods such as GAST \citep{Khomiakov_2024_WACV} and PolyR-CNN \citep{PolyR-CNN} focus inside a region of interest. They thereby avoid complex designs but mainly handle relatively simple and small urban areas.
 
Most of the aforementioned methods are designed to work with images and do not leverage 3D information. We present a method to adapt existing neural networks in the field to exploit both image and LiDAR modalities, making them more accurate in complex urban environments.
\section{The \dsname dataset}

We collect data from nine cities across seven regions in three countries. Our dataset covers a total area of 638km² and includes approximately 10B pixels, 10B LiDAR points, and 224k individual building polygon annotations. 
\tableref{tab:source} and \tableref{tab:details} 
provide detailed characteristics of the collected input data,  including licensing, download links, region names, image resolution, etc. We download the data from national or regional mapping agencies. All data is publicly available and free to download, re-use, modify and redistribute. Only the data from New Zealand requires an attribution under the CC-BY-4.0 license.

To prepare the raw remote sensing and cadastral data for deep learning frameworks, we resample, tile and harmonize images, point clouds and building polygons. We then perform a qualitative analysis to ensure overall data quality. The following sections describe all steps in detail.

\begin{table}
\caption{\textbf{Data sources}. We show the different licenses and restrictions of the collected data. We also provide links to the download pages of the original data source. All data can be downloaded and used for free without any restrictions. Only the New Zealand data requires attribution to the source.}
\label{tab:source}
    \centering
    \resizebox{0.87\textwidth}{!}{
    \begin{tabular}{@{}cccccc@{}}
    \toprule
    \bf Country     &\bf Split           &\bf Modality  &\bf License                  &\bf Restrictions         & \bf Link                                                                                                     \\ \midrule
    \multirow{4}{*}{\shortstack{United \\ States}}          & train, val, test & Images    & Custom                                                   & None                 & \href{https://data.gis.ny.gov/maps/sharegisny::nys-orthoimagery-latest/about}{data.gis.ny.gov}      \\

    & train, val, test & LiDAR     & None                                                     & None                 & \href{https://data.gis.ny.gov/maps/sharegisny::nys-latest-lidar-collections/about}{data.gis.ny.gov} \\

    & train, test      & Buildings & Custom                                                   & None                 & \href{https://nycmaps-nyc.hub.arcgis.com/datasets/nyc::building/about}{nycmaps-nyc.hub.arcgis.com}             \\

    & val              & Buildings & Custom                                                   & None                 & \href{https://data.gis.ny.gov/maps/a6bbc64e38f04c1c9dfa3c2399f536c4/about}{data.gis.ny.gov}         \\

    \midrule

    \multirow{3}{*}{\shortstack{Switzer- \\ land}} & train, val, test & Images    & None                                                     & None                 & \href{https://www.swisstopo.admin.ch/en/orthoimage-swissimage-10}{swisstopo.admin.ch}                  \\
    & train, val, test & LiDAR     & None                                                     & None                 & \href{https://www.swisstopo.admin.ch/en/height-model-swisssurface3d}{swisstopo.admin.ch}               \\
    & train, val, test & Buildings & None                                                     & None                 & \href{https://www.swisstopo.admin.ch/en/landscape-model-swisstlm3d}{swisstopo.admin.ch}               \\

\midrule
\multirow{3}{*}{\shortstack{New \\ Zealand}}          & train, val, test & Images    & \cellcolor[HTML]{FFFFFF}{\color[HTML]{202020} CC-BY-4.0} & Attribution required & \href{https://data.linz.govt.nz/data/category/aerial-photos/}{data.linz.govt.nz}                      \\
    & train, val, test & LiDAR     & \cellcolor[HTML]{FFFFFF}{\color[HTML]{202020} CC-BY-4.0} & Attribution required & \href{https://data.linz.govt.nz/layers/category/elevation/?kind=pointcloud}{data.linz.govt.nz}        \\
    & train, val, test & Buildings  & \cellcolor[HTML]{FFFFFF}{\color[HTML]{202020} CC-BY-4.0} & Attribution required & \href{https://data.linz.govt.nz/data/category/topographic/building-outlines/}{data.linz.govt.nz}      \\ \bottomrule
    \end{tabular}
    }
\end{table}
\begin{table}
  \caption{\textbf{Data details}. We collect data from nine different cities in three countries, ranging from urban to rural areas, and including a variety of building types. The dataset is split into training, validation, and test sets. The image GSD varies between 10 and 15~cm, while the average LiDAR point density varies between 6 and 21 points per square meter. Note that, we downsample the images to a GSD of 25 cm for the benchmark dataset. The annotations are either semi-automatically extracted from orthorectified images or constitute cadasteral data.}
\label{tab:details}
    \centering
    \resizebox{\textwidth}{!}{
\begin{tabular}{@{}ccccccccc@{}}
\toprule
  \multicolumn{4}{c}{\emph{Region}} & \emph{Image} & \emph{LiDAR}        & \multicolumn{3}{c}{\emph{Building Annotations}}      \\
  \midrule
  \textbf{Country}                   &  \textbf{City}                & \textbf{Type}    & \textbf{Split}                     & \textbf{GSD}~[cm]  & \textbf{Density}~[pts/m²] & \textbf{Source}                 & \textbf{Type}   & \textbf{Count} \\
\midrule
\multirow{3}{*}{\shortstack{United \\ States}}& New York City  & city        & train   & \multirow{3}{*}{15}          & 21 & \multirow{3}{*}{\shortstack{semi-automatic  \\ image extraction}} & \multirow{3}{*}{\shortstack{2D outline \\ polygon}}                                                                & 88k \\
                            & Albany                   & residential  & val                        &           & 6  &  &   & 1.3k \\
                            & Yonkers                           & industrial & test                       &           & 6  &  &   & 31k \\

\midrule
\multirow{3}{*}{\shortstack{Switzer- \\ land}}& Zurich         & city        & train   & \multirow{3}{*}{10}          & 17 & \multirow{3}{*}{\shortstack{manual \\ image extraction}} & \multirow{3}{*}{\shortstack{2D roof \\ wireframe}}                                             & 43k \\
                            & Basel                            & city        & val                        &           & 11 &  &   & 0.3k \\
                            & Jouxtens-Mézery                  & residential & test                       &           & 13 &  &   & 11k \\

\midrule
\multirow{3}{*}{\shortstack{New \\ Zealand}}& Twyford           & rural      & train   & \multirow{3}{*}{12.5}        &  16 & \multirow{3}{*}{\shortstack{semi-automatic  \\ image extraction}} & \multirow{3}{*}{\shortstack{2D outline \\ polygon}}                                                          & 31k \\
                            & Waipukurau                        & rural      & val                        &           & 19 &  &   & 1.6k \\
                            & Gisborne                          & city       & test                       &           & 19 &  &   & 16k \\\bottomrule
\end{tabular}
}
\end{table}

\subsection{Aerial images} The raw input images are near-nadir acquisition with a \ac{gsd} between 10 and 15~cm. The images are not orthorectified, resulting in some distortion in off-nadir image regions (see \figureref{fig:polygon_a}). To standardize the images across all regions, we resample them to a \ac{gsd} of 25~cm and clip them to non-overlapping tiles of 224$\times$224~pixels, equivalent to 56$\times$56~m on the ground. This process yields a total of 203k image tiles. The radiometry of the images is not altered. The images from the three countries exhibit different spectral characteristics due to varying acquisition conditions and sensors (see \eg \figureref{fig:all_countries}).


\subsection{Aerial LiDAR point clouds} The LiDAR point clouds exhibit a density ranging from 6 to 21 points per square meter across different regions. 
The point clouds are georeferenced to the same coordinate system as the images 
{and are spatially well alligned with the images. The data comes from national mapping agencies with documented quality specifications (linked in \tabref{tab:source}); for example, swisstopo reports a planimetric accuracy of $\pm$20~cm ($1\sigma$) and an altimetric accuracy of $\pm$10~cm ($1\sigma$).} We clip them to equally sized non-overlapping tiles of 56~m$\times$56~m. This results in 194k LiDAR tiles -- slightly fewer than image tiles due to missing LiDAR data in certain areas, such as deep water. On average, the LiDAR tiles contain between 20k and 70k points. We preserve the original point cloud density without resampling. Additionally, we do not color the point clouds, as it would first require to orthorectify the images before projecting the image colors on them.


\subsection{Building outline polygon annotations} 
\begin{figure}
    \centering
    \resizebox{\textwidth}{!}{
    \setlength{\tabcolsep}{2pt}
    \newcommand{\mywidth}{0.48\linewidth}
    \begin{subfigure}[t]{0.48\textwidth}
        \centering
		\begin{tabular}{cc}
            \includegraphics[width=\mywidth]{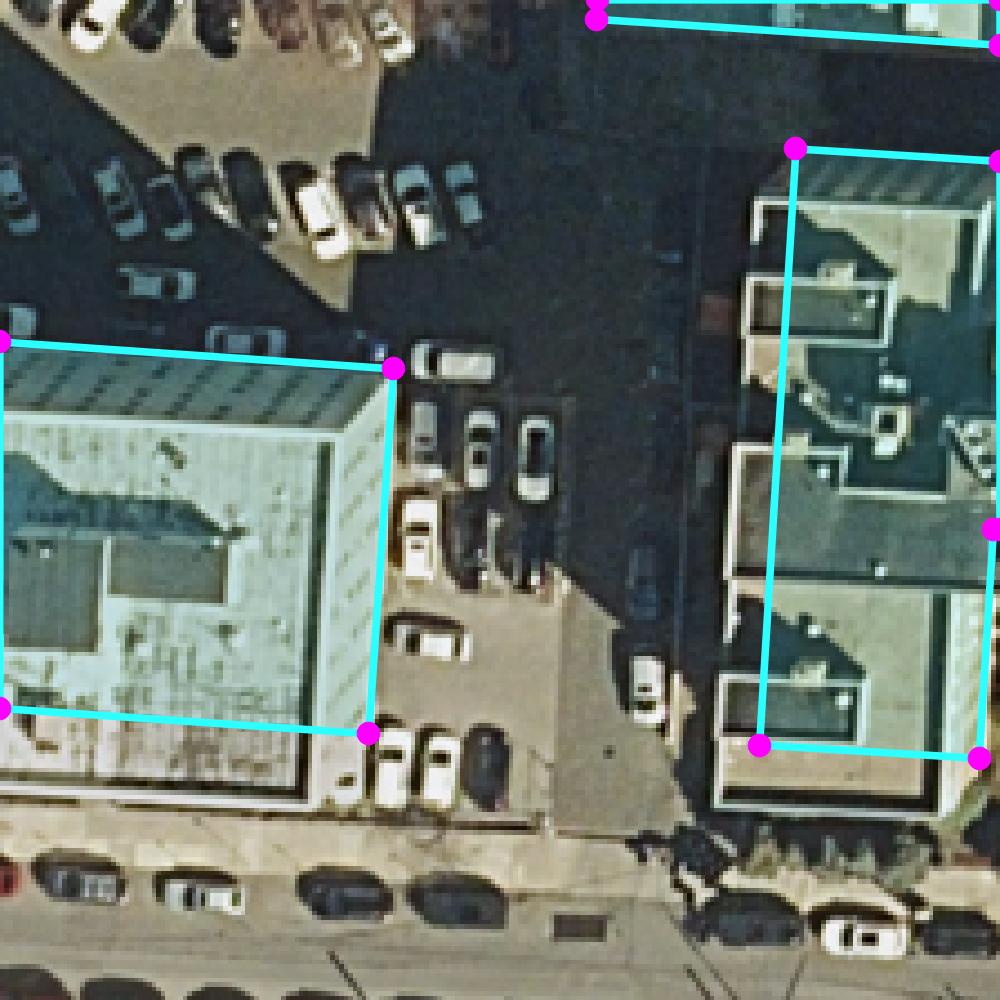} &
			\includegraphics[width=\mywidth]{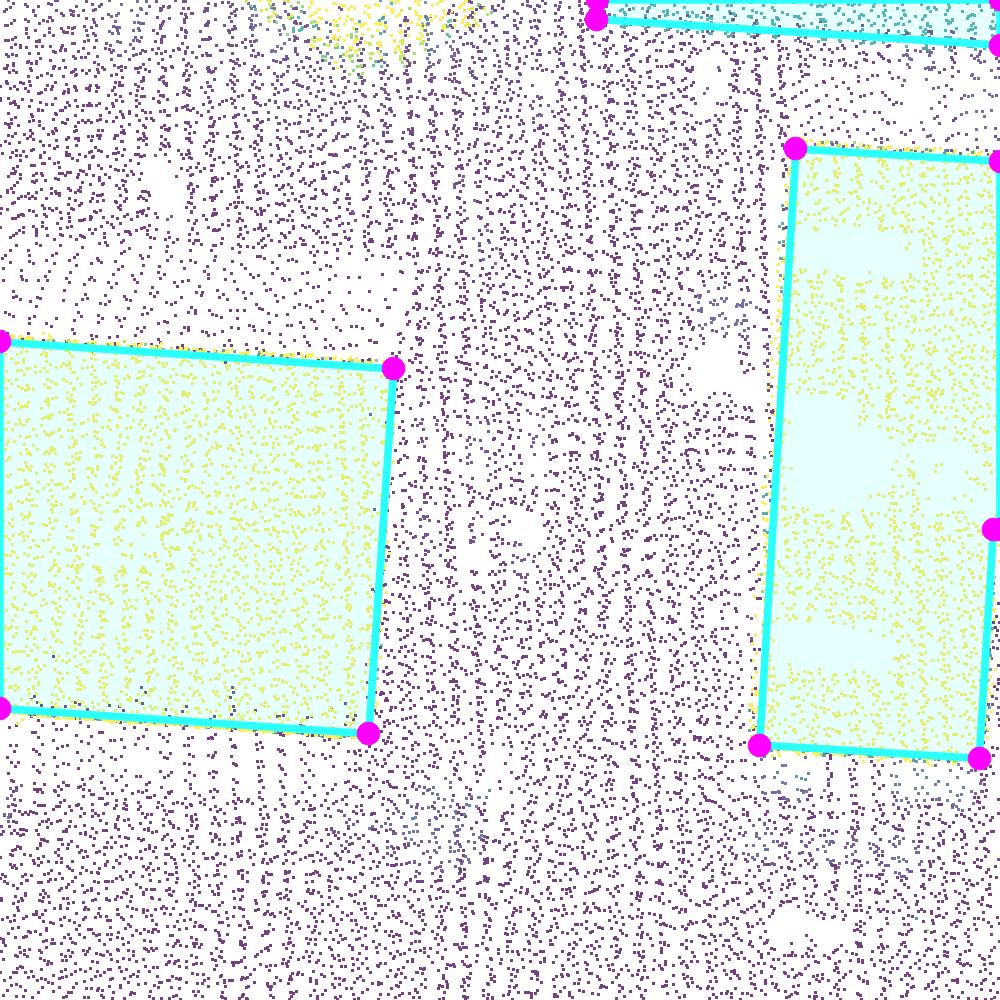}
		\end{tabular}
        \vspace{-2mm}
		\caption{Building outline polygon}
        \label{fig:polygon_a}
    \end{subfigure}
    \hspace{5pt}
    \begin{subfigure}[t]{0.48\textwidth}
        \centering
		\begin{tabular}{cc}
			\includegraphics[width=\mywidth]{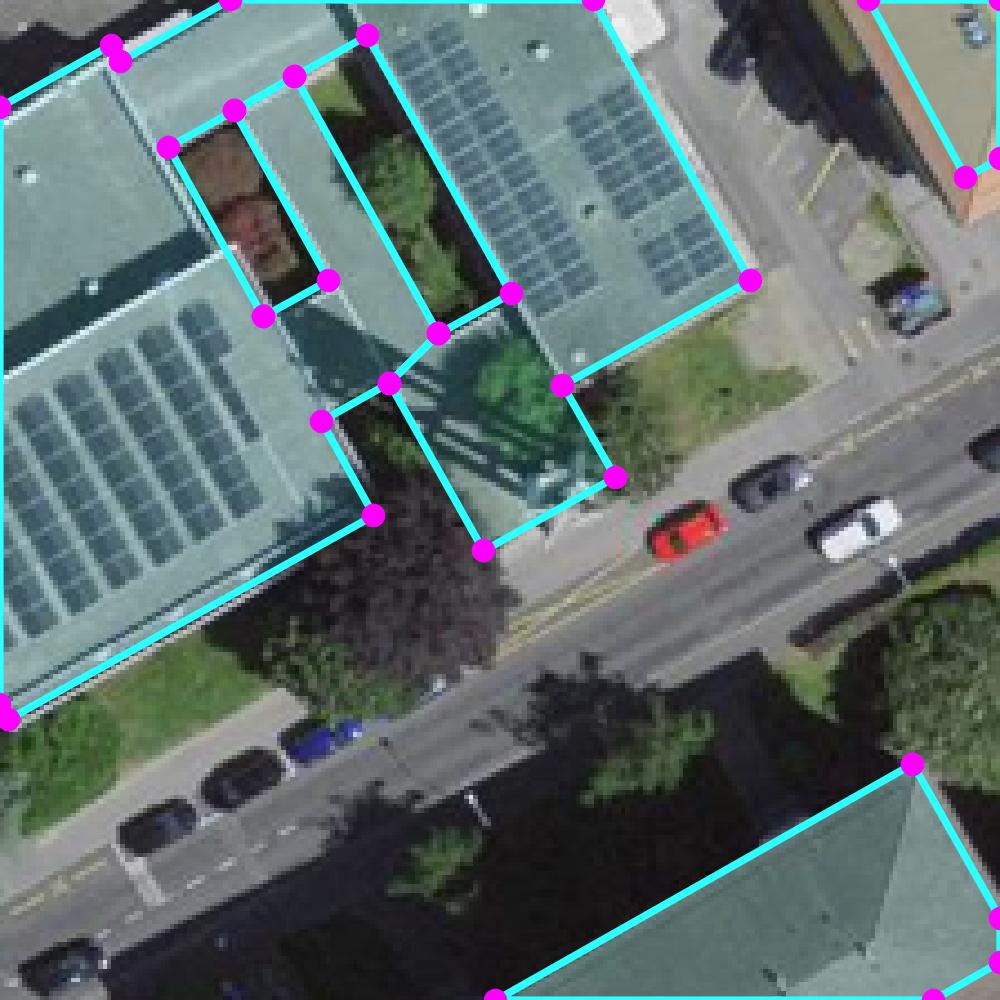} &
			\includegraphics[width=\mywidth]{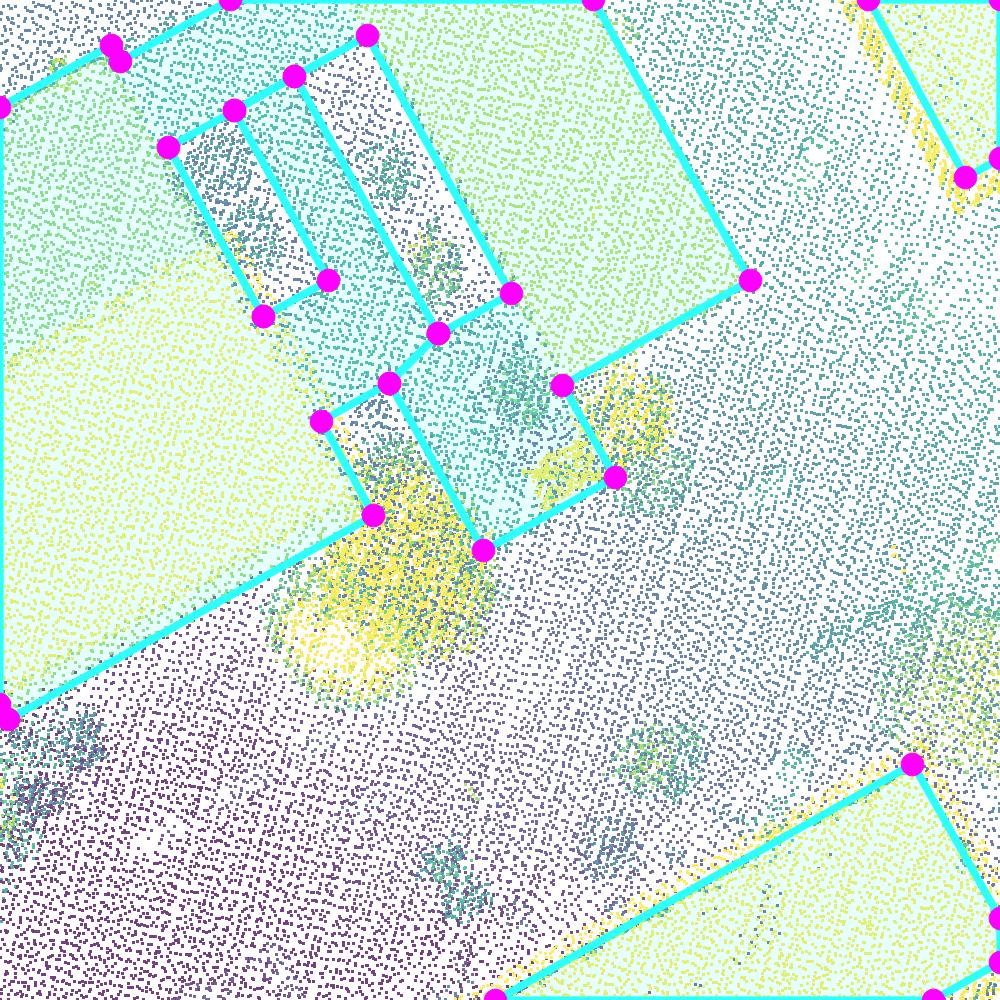}
		\end{tabular}
        \vspace{-2mm}
		\caption{Building outline polygon with holes}
        \label{fig:polygon_b}
    \end{subfigure}
    }
    \caption{\textbf{Image/LiDAR tiles with polygon annotations.} In (\subref{fig:polygon_a}), we show an example tile of Jersey, NY, US. The image exhibits distortion leading to the \emph{leaning building} effect. The annotation is nonetheless accurately placed on the base of the building. The LiDAR acquisition does not suffer from the \emph{leaning building} effect. In (\subref{fig:polygon_b}), we show an example tile of Zurich, Switzerland. The building has two interior cutouts. The annotation has thus two interior rings connected to the exterior ring at their closest vertex.}
    \label{fig:polygon_figure}
\end{figure}

Our dataset includes annotations as 2D polygons representing the orthogonal projections of the complete building perimeter onto the ground. For buildings without roof overhang, these polygons typically coincide with the building footprint -- the 2D polygon where the building touches the ground. In off-nadir image regions, roof outlines may appear slightly offset from the building footprint due to image distortion (see \figureref{fig:polygon_a}). {By providing annotations at the building base together with non-orthorectified images, our dataset reflects common real-world conditions and encourages models to learn relief displacement implicitly, reducing reliance on an additional orthorectification step. As shown in our experiments, the trained models are indeed capable of predicting polygons at the building base despite the image distortion.} We do not provide annotations for roof partitions or interior walls, as this information is only available for the Switzerland part of the dataset. Where two or more building outline polygons overlap in the input data we merge them into a single polygon. 
We clip the building polygons to match the same tile dimensions as the image and LiDAR tiles. Polygons spanning multiple tiles are split at the tile boundaries. We then convert the polygon annotations to the widely adapted MS-COCO format \citep{mscoco}.

In the urban environment, larger buildings commonly feature interior cutouts such as courtyards, lightwells, air shafts, atriums, passageways, or alleys. The MS-COCO format does not directly support storing interior rings as polygons with holes in vector format. Additionally, most current deep learning based vectorization methods do not specifically address interior ring prediction. We thus adopt the convention established by Xu \etal \citep{hisup} to connect interior and exterior rings of a polygon at their closest vertex (see \figureref{fig:polygon_b}). The upside of this approach is that it allows deep networks to predict interior rings without architecture modifications. However, it can create invalid geometry through self-intersection and complicate the explicit distinction between exterior and interior rings. 
To facilitate more robust solutions for future work, we include an additional attribute \emph{hashole}, in the MS-COCO annotation, enabling distinction between interior and exterior rings.

\subsection{Dataset quality control} We performe a broad visual inspection along with random checks to ensure overall quality of the ground truth polygons and their alignment with the input images and LiDAR point clouds. We found two types of errors: \emph{functional ambiguity} and \emph{temporal mismatch}.

\begin{figure}
    \centering
    \resizebox{\textwidth}{!}{
    \setlength{\tabcolsep}{2pt}
    \newcommand{\mywidth}{0.48\linewidth}
    \begin{subfigure}[t]{0.48\textwidth}
        \centering
		\begin{tabular}{cc}
			\includegraphics[width=\mywidth]{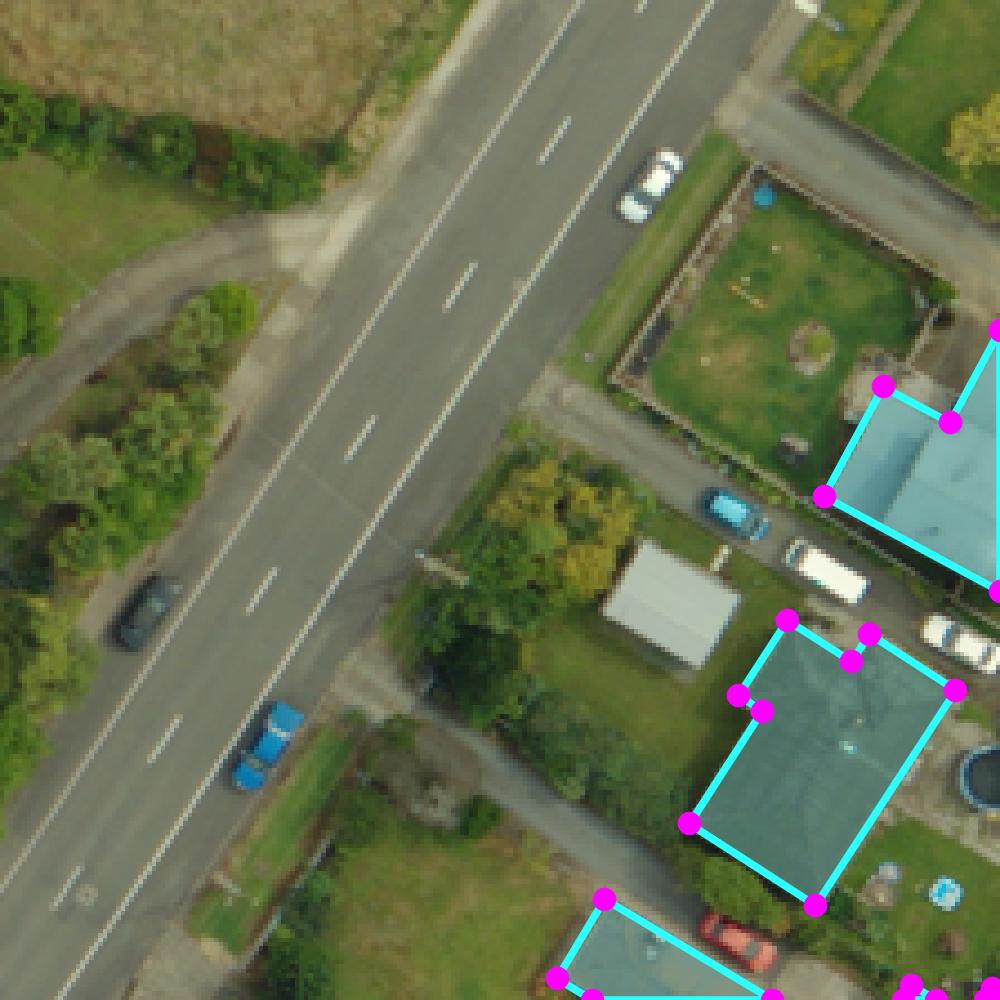} &
			\includegraphics[width=\mywidth]{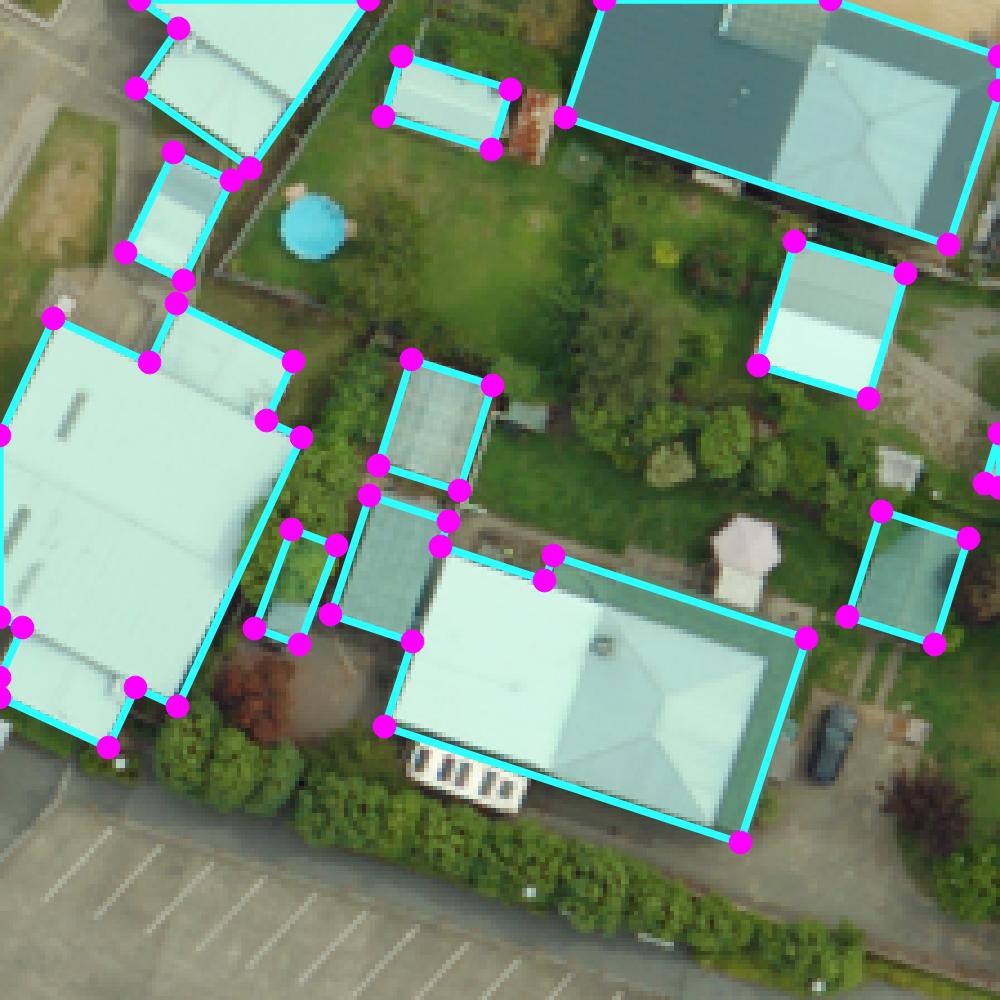}
		\end{tabular}
        \vspace{-2mm}
        		\caption{Functional ambiguity}
        \label{fig:ambiguity_a}
    \end{subfigure}
    \hspace{5pt}
    \begin{subfigure}[t]{0.48\textwidth}
        \centering
		\begin{tabular}{cc}
                        \includegraphics[width=\mywidth]{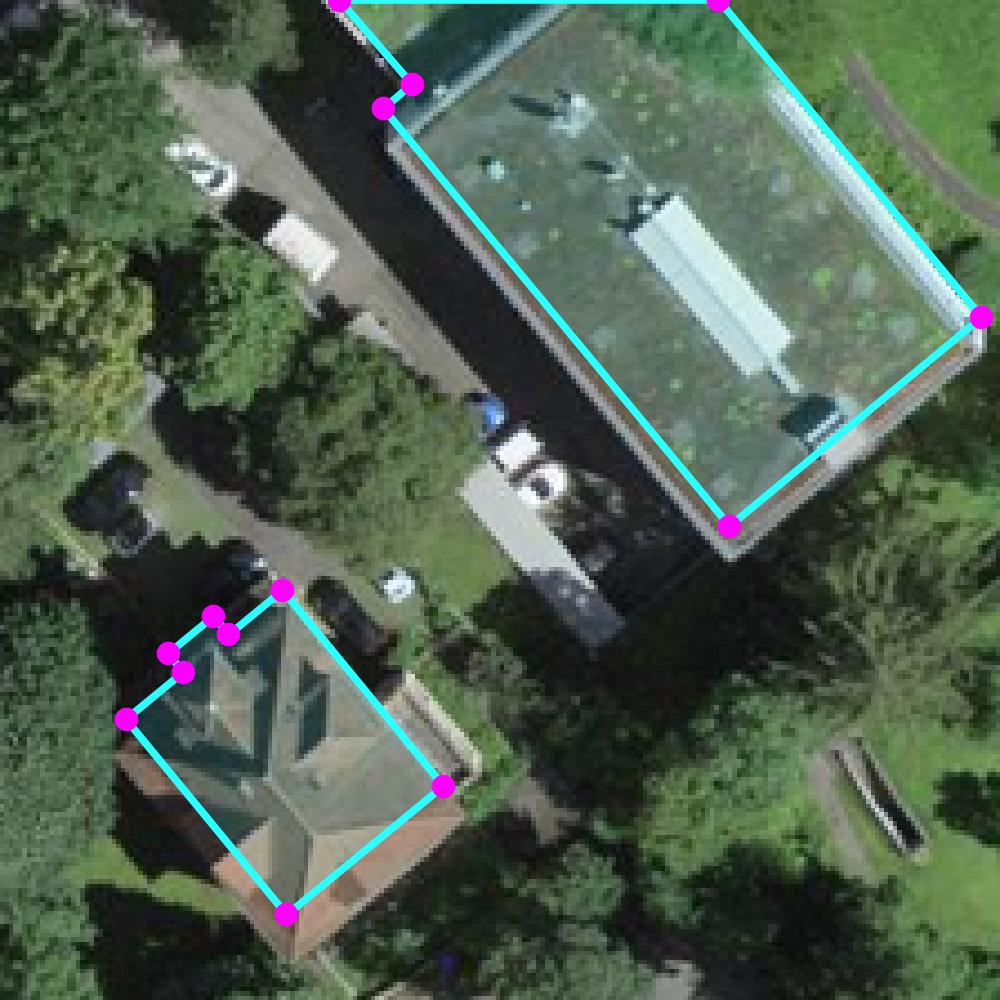} &
			\includegraphics[width=\mywidth]{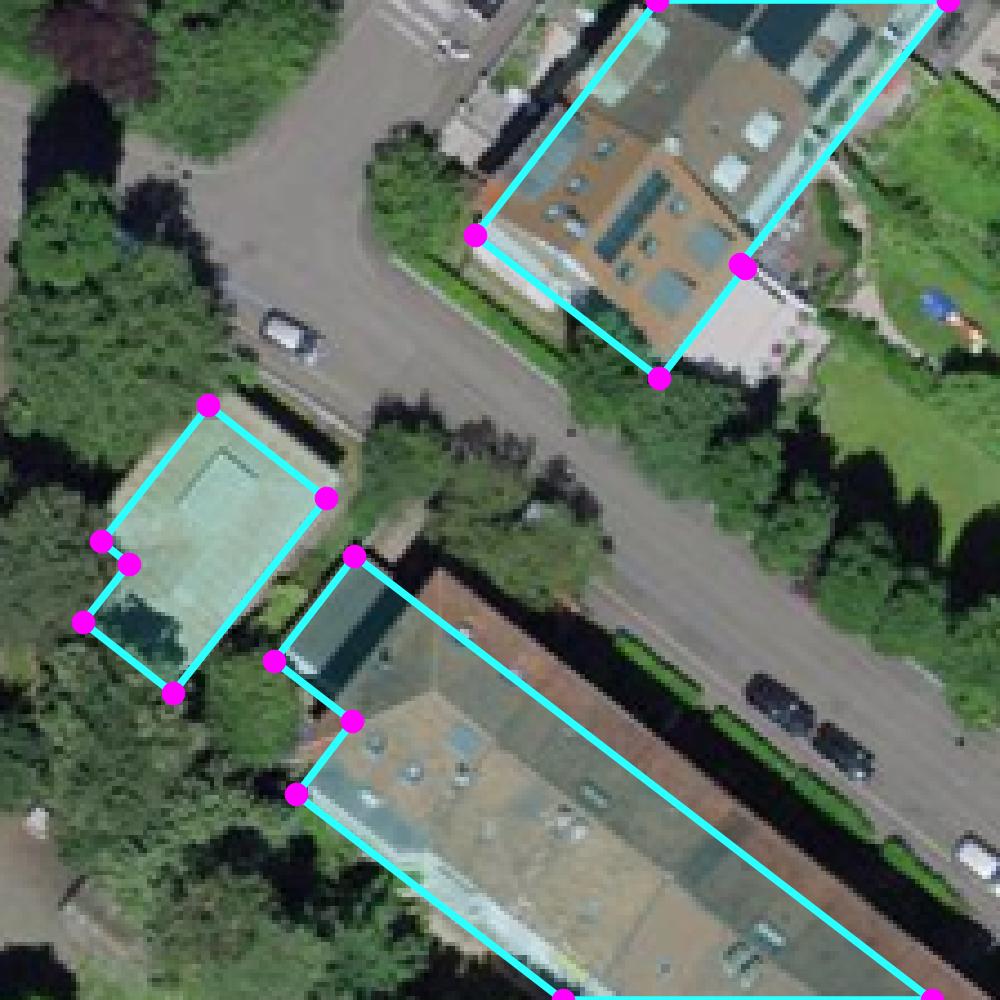}
		\end{tabular}
        \vspace{-2mm}
        		\caption{Level-of-detail ambiguity}

        \label{fig:ambiguity_b}
    \end{subfigure}
    }
    \caption{\textbf{Annotation ambiguity.} In (\subref{fig:ambiguity_a}), we show that a garage is not labelled as a building in the left tile, while similar structures are labelled as buildings in the right tile. In (\subref{fig:ambiguity_b}), we show how a small annex structure is not considered part of the left building in the left tile. In the right tile a similar annex is considered part of the bottom right building.}
    \label{fig:ambiguity_figure}
\end{figure}

\paragraph{Functional and level-of-detail ambiguity.}
The first type of errors corresponds to functional ambiguities, \ie situations where the definition of what is a building is not clear and subject to interpretation. This typically happens on (i) certain annex structures to a building like a garage or a garden shed that leads to over- or under detection (\figureref{fig:ambiguity_a}), and (ii) buildings with fine details such as small roof overhangs or sub-metric façade features that lead to a level-of-detail gap (\figureref{fig:ambiguity_b}). We manually count the tiles of the validation set in which such ambiguities occur and find that approximately 6\% of tiles are affected. While this is a relatively high percentage, this error only impacts buildings or building parts with a small area.
{In practice, we did not observe this being an issue during training as we observe that the models learn to predict the finer level-of-detail (see \eg \figureref{fig:ambiguity_b}, top image, the top-left building has more details in the prediction on its south-east façade than in the ground truth). The impact on evaluation metrics is also minor, since the affected building parts have a small area and thus contribute little to area-based metrics such as IoU.}

\paragraph{Temporal mismatch.}

\begin{figure}
    \centering
    \resizebox{\textwidth}{!}{
    \setlength{\tabcolsep}{2pt}
    \newcommand{\mywidth}{0.48\linewidth}
    \begin{subfigure}[t]{0.48\textwidth}
        \centering
		\begin{tabular}{cc}
            \includegraphics[width=\mywidth]{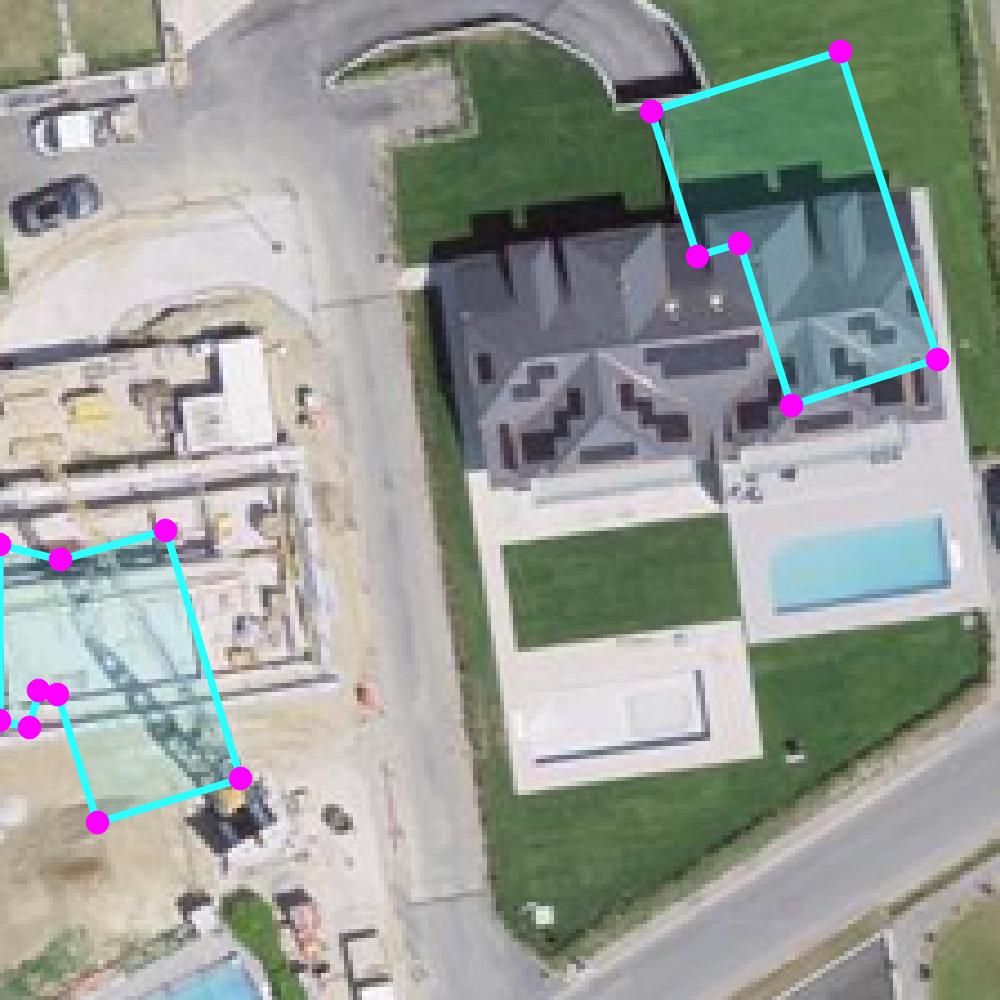} &
			\includegraphics[width=\mywidth]{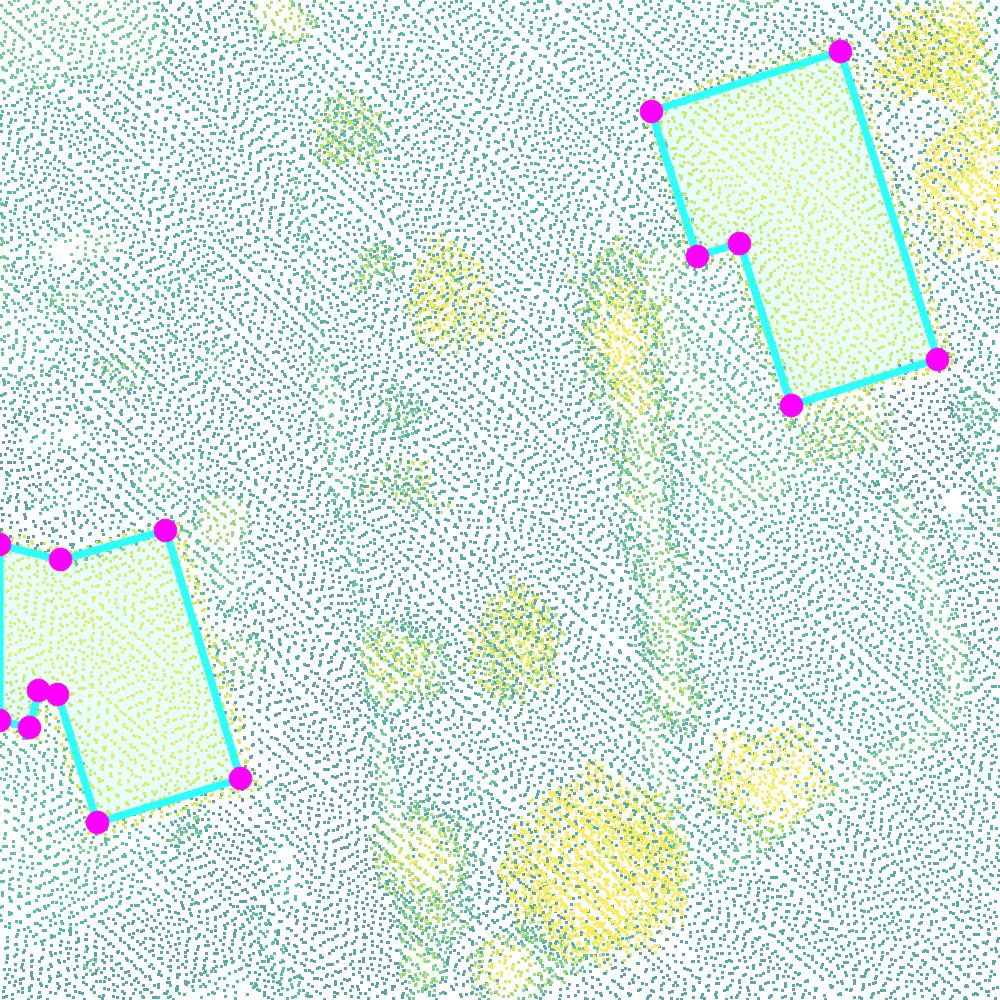}
		\end{tabular}
        \vspace{-2mm}
		\caption{Ground truth temporal mismatch}
        \label{fig:temp_a}
    \end{subfigure}
    \hspace{5pt}
    \begin{subfigure}[t]{0.48\textwidth}
        \centering
		\begin{tabular}{cc}
            \includegraphics[width=\mywidth]{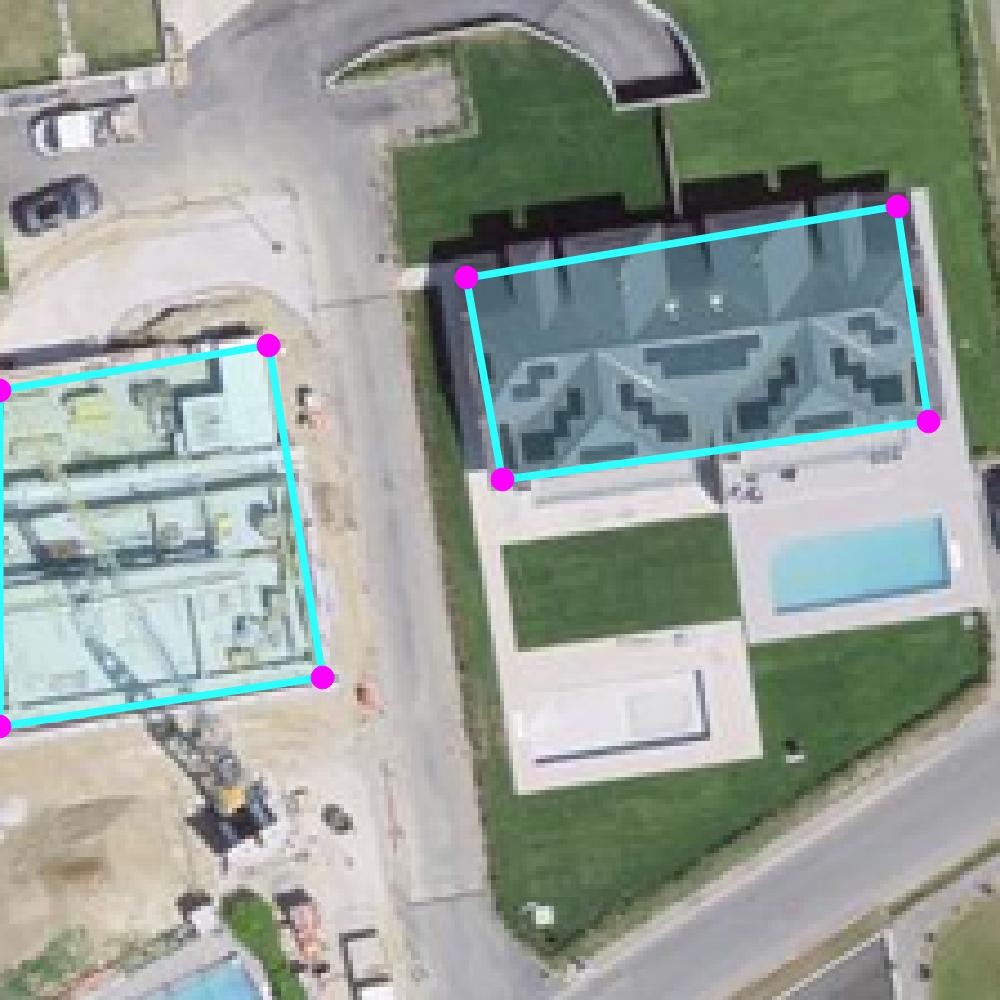} &
			\includegraphics[width=\mywidth]{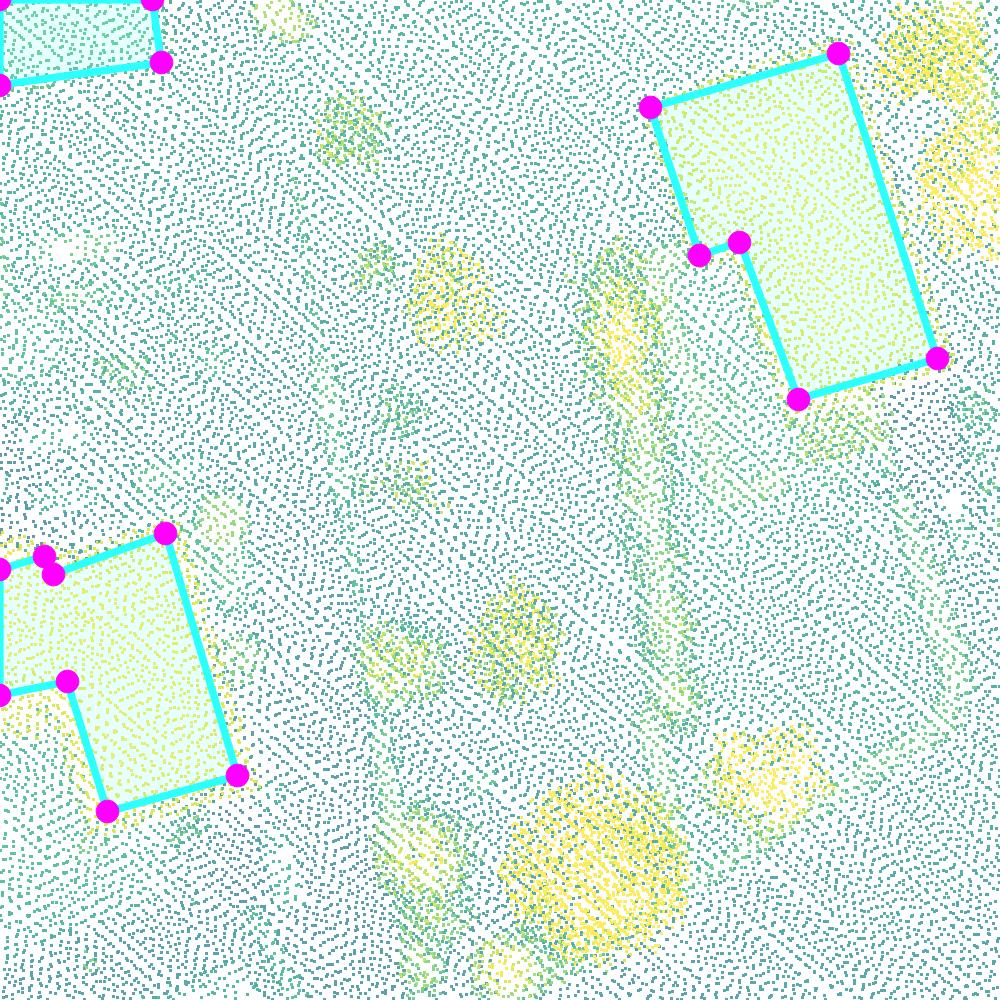}
		\end{tabular}
        \vspace{-2mm}
		\caption{Prediction temporal mismatch}
        \label{fig:temp_b}
    \end{subfigure}
    }
\caption{\textbf{Temporal mismatch.} In this area of our dataset, the image and LiDAR data have a temporal mismatch. The image shows two newly constructed buildings, while the LiDAR data is from a previous flight that shows buildings that are most likely demolished. The ground truth polygons (\subref{fig:temp_a}) align with the LiDAR data and not with the image. In (\subref{fig:temp_b}) we only use the image modality (left) to predict polygons  corresponing to the newly constructed buildings. When using only the LiDAR modality (right), the predicted polygons correspond to the demolished buildings. This strategy can help to identify such temporal mismatches.}
\label{fig:temporal_mismatch}
\end{figure}

The second type of errors corresponds to temporal mismatches between the input modalities and the ground truth polygons. Such mismatches occur when either the image or the LiDAR point cloud were acquired at a different time than the building outline annotations. This can lead to situations where buildings are newly constructed or demolished in one modality but not in the other (see \figureref{fig:temp_a}).

To identify areas with a temporal mismatch we train two versions of a polygon prediction model, one taking only images and one taking only LiDAR point clouds as input (see \figureref{fig:temp_b}). 
We then search for areas with a large predictive difference between the two modalities in the validation set. We manually inspect these areas and find that less than 1\% of annotations in the validation have a temporal mismatch. In most of these cases, either the LiDAR or the imagery is temporally misaligned with the ground truth polygon, but rarely both. We chose to retain these examples in the dataset to encourage the development of models that can learn to handle such inconsistencies by leveraging both modalities.
\section{Baseline methods}
\label{sec:baselines}

{In this section, we present the baseline methods competing in our polygon prediction benchmark.
We reimplement all methods in a unified, publicly available library.}

\begin{figure}
	\centering
	\begin{subfigure}[b]{0.49\textwidth}
		\centering
		\includegraphics[width=\linewidth]{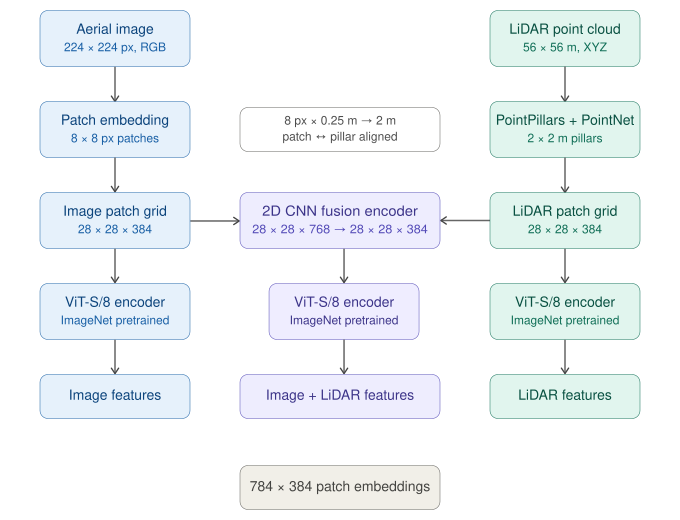}
		\vspace{3mm}	
		\caption{Encoder architectures and fusion design}
		\label{fig:baseline_networks_encoding_fusion}
	\end{subfigure}
	\hfill
	\begin{subfigure}[b]{0.49\textwidth}
		\centering
		\includegraphics[width=\linewidth]{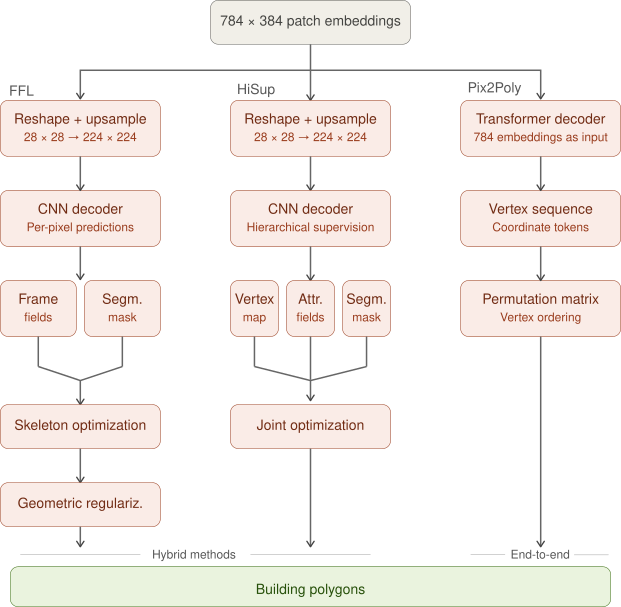}
		\caption{Decoder architectures}
		\label{fig:baseline_networks_decoders}
	\end{subfigure}
	\caption{\textbf{Baseline network components.} In (\subref{fig:baseline_networks_encoding_fusion}) we show the encoder and fusion design. In (\subref{fig:baseline_networks_decoders}) we show the decoder variants used in the benchmark.}
	\label{fig:baseline_networks}
\end{figure}

\subsection{Encoding architectures}

{In the following paragraphs, we describe the image, point cloud and fusion encoder illustrated in \figureref{fig:baseline_networks_encoding_fusion}.
The output of all encoders is a sequence of 784 patch embeddings, which are directly input to the Pix2Poly~\citep{pix2poly} decoder, or reshaped and upsampled to a spatial map for processing with the CNN-based decoders of \ac{ffl}~\citep{ffl} and HiSup~\citep{hisup}.}

\paragraph{Image encoder.}
{For the image encoder, we use a standard \ac{vit}~\citep{vit} pretrained on ImageNet~\citep{imagenet} with image input size of 224$\times$224 pixels and a patch size of 8$\times$8 pixels. The transformer extracts patch-level features without global pooling, thus producing a sequence of 784 patch embeddings.}

\paragraph{Point cloud encoder.}
{We design a novel point cloud encoder to process aerial LiDAR point clouds with varying point counts and densities, combining PointPillars \citep{pointpillars} and \ac{vit}. We first scale the point cloud to a fixed height of 1~m.
We then voxelize the point cloud into pillars with size 2$\times$2$\times$1~m.  
We use a maximum number of points per voxel of 64. We then extract pillar features for each voxel \citep{pointpillars} and process them with a small PointNet~\citep{pointnet}.
Choosing a 2$\times$2m pillar footprint makes the LiDAR representation compatible with a pretrained ViT, yielding 784 pillar embeddings as transformer tokens.
We then reuse the same \ac{vit} architecture as in the image encoder, replacing only the image patch-embedding layer with the point-cloud embedding module.
The point-cloud encoder thus outputs a sequence of 784 patch embeddings.}

\paragraph{Fusion encoder.} {We implement a fusion encoder that concatenates image and point-cloud features channel-wise by aligning the point-pillar size with the image patch size (8$\times$8 pixels at 25 cm GSD, i.e., 2$\times$2 m pillars). A lightweight 2D CNN then processes the fused representation to handle small image-LiDAR misalignments and compress the channel dimension, enabling the use of the same \ac{vit} architecture as above.}

\subsection{Decoding architectures}

{We test the three following baseline decoders on our dataset. The decoder architectures are illustrated in \figureref{fig:baseline_networks_decoders}.}

\paragraph{FFL}{~\citep{ffl} employs a CNN-based decoder to predict per-pixel orientation (frame) fields and segmentation masks.Polygonization is performed by tracing the predicted frame fields to extract structured, direction-aligned polygon boundaries, followed by geometric regularization.}
\paragraph{HiSup.} {The CNN-based decoder of HiSup~\citep{hisup} predicts (i) a point feature map for detecting convex and concave polygon vertices, (ii) an edge feature map in the form of attraction field maps, (iii) and a segmentation mask to represent the area enclosed by the polygon. These outputs are jointly processed and input into an optimization pipeline to reconstruct the final polygons.
\paragraph{Pix2Poly}~\citep{pix2poly} directly uses the ViT patch embeddings as input to a transformer-based decoder that autoregressively predicts vertex coordinates. Self-attention enables the decoder to model dependencies between vertices, while cross-attention over the ViT image patch embeddings grounds predictions in the input features. Additionally, a permutation matrix orders and groups the vertices to produce polygons. This enables direct, end-to-end polygon reconstruction without relying on intermediate per-pixel segmentations.}

\section{Experiments}

In this section, we introduce a benchmark of multimodal building outline prediction on the P$^3$ dataset. We also present several ablation studies on the model architecture and input modalities using a subset of P$^3$ and other datasets.

\subsection{Metrics}
\label{sec:metrics}

We use three different types of metrics to evaluate the performance of the baseline methods: (i) boundary- and area-based accuracy and completeness metrics, (ii) complexity metrics, and (iii) efficiency metrics, \ie average prediction time and the number of trainable parameters of the models. All metrics are computed per tile and then averaged over all tiles. Formal definitions of the metrics can be found in \appendref{sm:evaluation_metrics}.

\paragraph{Boundary- and area-based metrics.}

(i) The \ac{iou} metric provides a combined measure of accuracy and completeness of the predicted polygons. However, it does not provide a good measure of the boundary distance between the predicted and ground truth polygons, as it is mainly influenced by the polygon area \citep{boundary_iou}. (ii) Average Precision (AP) and average recall (AR) computed over different \ac{iou} thresholds measure the accuracy and completeness of the predicted polygon sets \citep{mscoco}. (iii) The POLIS metric \citep{polis} measures the symmetric distance between each predicted polygon vertex and its closest point on the ground truth polygon boundary, and vice versa. (iv, v) The Hausdorff and Chamfer distance (HD and CD) measure the average and maximum boundary distance of predicted and ground truth polygons, irrespective of vertex locations. (vi) Finally, the \ac{mta}~\citep{ffl} measures the maximum angular error between the predicted and ground truth polygon edges. 

POLIS, HD, CD and MTA are computed for ground truth and predicted polygon pairs with a minimum \ac{iou} of 0.5. They are important measures for practitioners because they correlate with a high visual similarly of predicted and ground truth polygons.

\paragraph{Complexity metrics.}
(i) NR measures the relative difference of the number of predicted and ground truth vertices. (ii) C-IoU, defined as the product of NR and \ac{iou} measures the similarity between predicted and ground truth polygons in terms of both shape \emph{and} complexity. And finally, (iii) the normalized \ac{dof} measure the regularity and symmetry of polygon edges \citep{lineFit_eccv2024}. 
A lower DoF score indicates that the predicted polygons are simpler and more regular, while a higher DoF score indicates that the predicted polygons are more complex and less regular.

\subsection{Experimental setup}
\label{sm:experimental_setup}

We perform data augmentation on the images and point clouds in the form of random horizontal and vertical flips, and random 90$^\circ$ rotations. We also apply random color jitter, brightness and contrast changes to the images.
We train all models with a batch size of 16 and an initial learning rate of 0.0003, using the Adam optimizer.
The learning rate is linearly decayed to zero over 200 epochs following a warmup phase.
We train on four NVIDIA V100 GPUs with 16GB VRAM each. Training takes between 20 and 40 hours for 200 epochs. The prediction times are measured on a single NVIDIA A6000 GPU.


\subsection{Multimodal building outline prediction}

We first evaluate the performance of the baseline methods on the full P$^3$ dataset. Each model is trained with our fusion encoder using both images and LiDAR point clouds as input. In \tableref{tab:all_countries} we show the quantitative results. 
Pix2Poly produces the best polygons in terms of complexity and accuracy, exhibiting the best or second best results for all metrics. 
\figureref{fig:all_countries}, \figureref{fig:all_countries_sm1} and \figureref{fig:all_countries_sm2} show examples of Pix2Poly's more compact and accurate polygon predictions compared to other methods.
However, the figures also show that Pix2Poly still has some shortcomings such as incorrect connections in the building outline which lead to significant boundary errors, missing interior cutouts or entirely missing building outlines.
The HiSup-predicted polygons are the only ones that accurately capture interior cutouts.
A common limitation of the CNN-based methods HiSup and FFL is their tendency to produce excessive vertices for building polygons, falling to fully address the regularity and symmetry present in urban areas.

Compared to existing datasets, our dataset is more challenging due to the presence of complex building types and its greater variability, leading to a drop in all metrics. For example, the mean \ac{iou} decreases by about 5 percentage points compared to the results reported on the WHU dataset \citep{whu_dataset}. In terms of relative performance of the competing methods, our results align with those reported on other datasets \citep{inria_dataset,devglobal2022shanghai,gfdrr2020opencities,whu_dataset}. However, using identical encoders across all methods results in smaller performance gaps between the tested methods.

\begin{table}
    \caption{\textbf{Multimodal polygon prediction} of baseline models with our fusion encoder trained and tested on the full P$^3$ dataset. For each metric, we highlight the \colorbox{blue!25}{best} and \colorbox{blue!10}{second best} scores.}
    \setlength{\tabcolsep}{3pt}
    \centering
    \resizebox{0.9\textwidth}{!}{
    \begin{tabular}{@{}lH|cccccccccc@{}}
    \toprule
        &   & \multicolumn{4}{c}{\emph{Boundary}}  &  \multicolumn{3}{c}{\emph{Area}} &  \multicolumn{3}{c}{\emph{Complexity}}  \\
    \midrule
    \textbf{Model} & Unnamed: 0 & \textbf{POLIS [m]} $\downarrow$ & \textbf{CD [m]} $\downarrow$ & \textbf{HD [m]} $\downarrow$ & \textbf{MTA [$^\circ$]} $\downarrow$ & \textbf{AP} $\uparrow$ & \textbf{AR} $\uparrow$ & \textbf{IoU} $\uparrow$ & \textbf{C-IoU} $\uparrow$ & \textbf{NR=1} & \textbf{DoF} $\downarrow$ \\
    \midrule
    \textbf{FFL}~\citep{ffl} & \detokenize{ffl/v0_all_bs4x16} & 2.41 & 2.11 & 9.34 & 40.4 & 0.286 & 0.43 & 0.832 & 0.763 & 0.848 & 0.808 \\
    \textbf{HiSup}~\citep{hisup} & \detokenize{hisup/v0_all_bs4x16} & \cellcolor{blue!10} 2.07 & \cellcolor{blue!10} 2.03 & \cellcolor{blue!10} 9.1 & \cellcolor{blue!10} 36.5 & \cellcolor{blue!10} 0.309 & \cellcolor{blue!10} 0.474 & \cellcolor{blue!25} 0.844 & \cellcolor{blue!10} 0.789 & \cellcolor{blue!10} 0.881 & \cellcolor{blue!10} 0.758 \\
    \textbf{Pix2Poly}~\citep{pix2poly} & \detokenize{pix2poly/v0_all_bs4x16} & \cellcolor{blue!25} 1.91 & \cellcolor{blue!25} 1.92 & \cellcolor{blue!25} 8.17 & \cellcolor{blue!25} 34.4 & \cellcolor{blue!25} 0.347 & \cellcolor{blue!25} 0.49 & \cellcolor{blue!10} 0.842 & \cellcolor{blue!25} 0.801 & \cellcolor{blue!25} 0.901 & \cellcolor{blue!25} 0.741 \\
    \bottomrule
    \end{tabular}
    }
    \label{tab:all_countries}
\end{table}

\begin{figure}
    \centering
	\definetrim{mytrim}{0 0 0 0}
	\newcommand{\mywidth}{0.25\linewidth}
	\newcommand{\myfontsize}{\scriptsize}
	\setlength{\tabcolsep}{0mm}
    \newcommand{\baseimg}{./images/all_countries_figure}
    \newcommand{\imgpath}[1]{\baseimg/#1} 

\newcommand{\inputImages}[4]{%
    &
    \includegraphics[width=\mywidth,mytrim]{\imgpath{#1_test_#2_#3_#4.jpg}} &
    \includegraphics[width=\mywidth,mytrim]{\imgpath{#1_test_#2_pred_lidar.jpg}} &
    \includegraphics[width=\mywidth,mytrim]{\imgpath{#1_test_#2_pred_both.jpg}} &
    \includegraphics[width=\mywidth,mytrim]{\imgpath{#1_test_#2_gt.jpg}} \\
}
\newcommand{\inputPred}[4]{%
    \includegraphics[width=\mywidth,mytrim]{\imgpath{#1_test_#2_#3_#4.jpg}}
}
\newcommand{\inputGt}[3]{%
    \includegraphics[width=\mywidth,mytrim]{\imgpath{#1_test_#2_gt_#3}}
}

\centering
\resizebox{0.98\textwidth}{!}{

\begin{tabular}{@{}l@{\hspace{3pt}}cccc@{}}

\rotatebox{90}{\hspace{13mm}Switzerland} &
\inputGt{all}{3739}{both} &
\inputPred{all}{3739}{ffl}{v0_all_bs4x16} &
\inputPred{all}{3739}{hisup}{v0_all_bs4x16} &
\inputPred{all}{3739}{pix2poly}{v0_all_bs4x16} \\

\rotatebox{90}{\hspace{12mm}New Zealand} &
\inputGt{all}{22222}{both} &
\inputPred{all}{22222}{ffl}{v0_all_bs4x16} &
\inputPred{all}{22222}{hisup}{v0_all_bs4x16} &
\inputPred{all}{22222}{pix2poly}{v0_all_bs4x16} \\

\rotatebox{90}{\hspace{17mm}USA} &
\inputGt{all}{36151}{both} &
\inputPred{all}{36151}{ffl}{v0_all_bs4x16} &
\inputPred{all}{36151}{hisup}{v0_all_bs4x16} &
\inputPred{all}{36151}{pix2poly}{v0_all_bs4x16} \\

&\makebox[\mywidth][c]{Ground truth} &
\makebox[\mywidth][c]{FFL \citep{ffl}} &
\makebox[\mywidth][c]{HiSup \citep{hisup}} &
\makebox[\mywidth][c]{Pix2Poly \citep{pix2poly}}\\

\end{tabular}
}
\caption{\textbf{Multimodal polygon prediction.} 
We show ground truth and predicted building outlines of P$^3$, for Swizerland, New Zealand\protect\footnotemark[1]~and the USA (from top to bottom), with LiDAR point clouds superimposed on aerial images. 
The first column shows the ground truth reference polygons, while the second to forth column show polygons predicted using both input modalities.  FFL predicts the most complete polygon set, but also the polygons with the most noise. HiSup predicts more regular polygons and is the only method which accurately reconstructs the inner courtyard in the example in row three. Pix2Poly predicts the most regular polygons, but also produces gross errors, such as a missing vertex in row one, a missing building in row two and a missing interior structure in row three. 
}
\label{fig:all_countries}
\end{figure}

\subsection{Ablation studies}

In the following section, we conduct several ablation studies on a subset of our dataset that includes only data from Switzerland. The Switzerland subset is less challenging compared to the full dataset due to its lower variability. However, it allows us to investigate the impact of different model components more rapidly. We first test different backbones for the polygon prediction models and then assess the influence of the different input modalities on the polygon prediction task.

\paragraph{Backbone architecture.}

We train a Pix2Poly model with a ConvNeXt~\citep{convnext} and a \ac{vit}~\citep{vit} backbone pretrained using the DINO~\citep{dinov1,dinov2,dinov3} self-supervised learning strategies.

\tableref{tab:backbone_ablation_table} shows the different backbone configurations and their performance. The small ConvNeXt backbone has the fastest inference time with only 1.03 seconds per image, but achieves the lowest performance in all other metrics compared to the \ac{vit} backbones.
While DINOv2 and v3 pretraining generally results in stronger features for many vision tasks compared to DINOv1, our results indicate that a finer spatial resolution of 8x8 pixel patches leads to more accurate and complete polygon predictions. However, the smaller patch size also leads to a slower inference time due to the larger number of patches that need to be processed.

\begin{table}
    \caption{\textbf{Backbone ablation}. We compare a Pix2Poly~\citep{pix2poly} model trained and tested on the Switzerland subset with different backbones. We use the DINOv1 \ac{vit}-S model with patch size 8$\times$8 as the default backbone for all other experiments due to its strong performance on the boundary, area and complexity metrics.}
    \label{tab:backbone_ablation_table}
    \setlength{\tabcolsep}{3pt}
    \centering
\resizebox{\textwidth}{!}{
\begin{tabular}{@{}llc@{}H|cccccc@{}}
\toprule
    & & & & \multicolumn{2}{c}{\emph{Boundary}}& \multicolumn{1}{c}{\emph{Area}}  & \emph{Complexity} &  \multicolumn{2}{c}{\emph{Efficiency}} \\
\midrule
\textbf{Backbone} & \textbf{Pretraining} & \textbf{Patch size} & Unnamed: 0 & \textbf{POLIS [m]} $\downarrow$ & \textbf{MTA [$^\circ$]} $\downarrow$ & \textbf{IoU} $\uparrow$ & \textbf{NR=1} & \textbf{Time [s]} $\downarrow$ & \textbf{Params [$\times 10^6$]} $\downarrow$ \\
\midrule
\textbf{ConvNeXt-T}~\citep{convnext} & DINOv3~\citep{dinov3} & -- & \detokenize{pix2poly/image_bs2x16_convnext_tiny_fd256} & 2.99 & 34.2 & 0.817 & 0.897 & \cellcolor{blue!25} 1.03 & 37.9 \\
\textbf{ViT-S}~\citep{vit} & DINOv1~\citep{dinov1} & 8x8 & \detokenize{pix2poly/v4_image_vit_bs4x16} & \cellcolor{blue!25} 2.46 & 34.3 & \cellcolor{blue!25} 0.845 & \cellcolor{blue!25} 0.906 & 1.44 & \cellcolor{blue!25} 31.9 \\
\textbf{ViT-S}~\citep{vit} & DINOv2~\citep{dinov2} & 14x14 & \detokenize{pix2poly/dinov2} & 3.3 & \cellcolor{blue!25} 34.1 & 0.805 & 0.883 & 1.13 & \cellcolor{blue!10} 32.2 \\
\textbf{ViT-S+}~\citep{vit} & DINOv3~\citep{dinov3} & 16x16 & \detokenize{pix2poly/image_bs2x16_dinov3} & \cellcolor{blue!10} 2.58 & \cellcolor{blue!25} 34.1 & \cellcolor{blue!10} 0.836 & \cellcolor{blue!10} 0.902 & \cellcolor{blue!10} 1.11 & 38.8 \\
\bottomrule
\end{tabular}
}
\end{table}

\paragraph{Modality ablation.} To evaluate the impact of different input modalities on the polygon prediction task, we train each baseline method with a point, an image, and a fusion encoder. \tableref{tab:modality_ablation} demonstrates that models utilizing LiDAR data consistently predict more complete ($\sim$0.55 AR) and accurate polygons (0.35-0.38 AP) compared to models using only images (0.46-0.49 AR, 0.28-0.32 AP). The fusion encoder efficiently integrates the strengths of both input modalities to achieve the best performance for all models.
Interestingly, for angular error MTA and vertex ratio NR, FFL and HiSup perform better for image-predicted polygons. This can be attributed to the presence of more prominent corner and edge features in the images compared to LiDAR point clouds.
The recently proposed end-to-end polygon prediction approach, Pix2Poly, exhibits significantly slower runtime compared to hybrid methods.
\figureref{fig:modality_ablation}, \figureref{fig:modality_ablation_sm1} and \figureref{fig:modality_ablation_sm2} confirm the higher accuracy and geometric simplicity of LiDAR- and multimodal-predicted polygons compared to those predicted from images alone. 
{One possible explanation for the consistently better boundary metrics of LiDAR-based models is that LiDAR features are inherently aligned with the building base, whereas image-based models must learn to predict building base polygons despite the relief displacement present in non-orthorectified images (see \figureref{fig:polygon_a}).}

\begin{figure}
    \centering
    \definetrim{mytrim}{0 0 0 0}
    \newcommand{\mywidth}{0.25\linewidth}
    \newcommand{\myfontsize}{\scriptsize}
    \setlength{\tabcolsep}{0mm}
    \newcommand{\baseimg}{./images/modality_ablation_figure}
    \newcommand{\imgpath}[1]{\baseimg/#1}

    \newcommand{\inputImages}[4]{%
        &
        \includegraphics[width=\mywidth,mytrim]{\imgpath{#1_val_#2_#3_#4.jpg}} &
        \includegraphics[width=\mywidth,mytrim]{\imgpath{#1_val_#2_pred_lidar.jpg}} &
        \includegraphics[width=\mywidth,mytrim]{\imgpath{#1_val_#2_pred_both.jpg}} &
        \includegraphics[width=\mywidth,mytrim]{\imgpath{#1_val_#2_gt.jpg}} \\
    }
    \newcommand{\inputPred}[4]{%
        \includegraphics[width=\mywidth,mytrim]{\imgpath{#1_val_#2_#3_#4.jpg}}
    }
    \newcommand{\inputGt}[3]{%
        \includegraphics[width=\mywidth,mytrim]{\imgpath{#1_val_#2_gt_#3}}
    }

    \resizebox{0.98\textwidth}{!}{%
        \begin{tabular}{@{}c@{}c@{}c@{}c@{\hspace{2pt}}l@{}}

        \inputGt{Switzerland}{0}{image} &
        \inputPred{Switzerland}{0}{ffl}{v4_image_bs4x16} &
        \inputPred{Switzerland}{0}{hisup}{v3_image_vit_cnn_bs4x12} &
        \inputPred{Switzerland}{0}{pix2poly}{v4_image_vit_bs4x16} &
        \rotatebox{-90}{\hspace{-28mm}Pred. Image} \\

        \inputGt{Switzerland}{0}{lidar} &
        \inputPred{Switzerland}{0}{ffl}{v5_lidar_bs2x16_mnv64} &
        \inputPred{Switzerland}{0}{hisup}{lidar_pp_vit_cnn_bs2x16_mnv64} &
        \inputPred{Switzerland}{0}{pix2poly}{lidar_pp_vit_bs2x16_mnv64} &
        \rotatebox{-90}{\hspace{-28mm}Pred. LiDAR} \\

        \inputGt{Switzerland}{0}{both} &
        \inputPred{Switzerland}{0}{ffl}{v4_fusion_bs4x16_mnv64} &
        \inputPred{Switzerland}{0}{hisup}{early_fusion_vit_cnn_bs2x16_mnv64} &
        \inputPred{Switzerland}{0}{pix2poly}{early_fusion_bs2x16_mnv64} &
        \rotatebox{-90}{\hspace{-28mm}Pred. Fusion} \\

        \makebox[\mywidth][c]{Ground truth} &
        \makebox[\mywidth][c]{FFL \citep{ffl}} &
        \makebox[\mywidth][c]{HiSup \citep{hisup}} &
        \makebox[\mywidth][c]{Pix2Poly \citep{pix2poly}} & \\
        \end{tabular}
    }

    \caption{\textbf{Modality ablation.} We present a sample tile displaying predicted and ground truth building polygons from the Switzerland subset. The first column shows ground truth polygons, while subsequent columns show predicted polygons from baseline models trained on different input modalities, \ie images only (first row), LiDAR only (second row), and the fusion of image and LiDAR data (third row).
    Note, in the bottom right corner, where a tree obscuring a building corner adversely affects the image-only prediction, while LiDAR-only and multimodal polygon predictions remain unaffected by this occlusion. Across all models, polygons predicted using multimodal inputs demonstrate superior simplicity and accuracy, especially with Pix2Poly.}
    \label{fig:modality_ablation}
\end{figure}

\begin{table}
            \caption{\textbf{Modality ablation}. We compare the baseline models with our image, LiDAR and fusion encoders trained and tested on the different modalities of the Switzerland subset.}
    \setlength{\tabcolsep}{2pt}
    \centering
    \resizebox{\textwidth}{!}{
        \begin{tabular}{@{}cc@{}H|cccccccccc@{}}
            \toprule
                & & & \multicolumn{4}{c}{\emph{Boundary}}& \multicolumn{3}{c}{\emph{Area}}  & \emph{Complexity} &  \multicolumn{2}{c}{\emph{Efficiency}} \\
            \midrule
            \textbf{Model} & \textbf{Modality} & Unnamed: 0 & \textbf{POLIS [m]} $\downarrow$ & \textbf{CD [m]} $\downarrow$ & \textbf{HD [m]} $\downarrow$ & \textbf{MTA [$^\circ$]} $\downarrow$ & \textbf{AP} $\uparrow$ & \textbf{AR} $\uparrow$ & \textbf{IoU} $\uparrow$ & \textbf{NR=1} & \textbf{Time [s]} $\downarrow$ & \textbf{Params} $\downarrow$ \\
            \midrule
            \multirow{3}{*}{ \textbf{FFL}~\citep{ffl}} & \textbf{Image} & \detokenize{ffl/v4_image_bs4x16} & 3 & 2.7 & 12 & 41 & 0.275 & 0.46 & 0.839 & 0.847 & 0.532 & \cellcolor{blue!25} 23.7M \\
             & \textbf{LiDAR} & \detokenize{ffl/v5_lidar_bs2x16_mnv64} & 2.35 & 1.94 & 9.66 & 44.8 & 0.359 & 0.545 & 0.87 & 0.829 & 0.582 & \cellcolor{blue!25} 23.7M \\
             & \textbf{Fusion} & \detokenize{ffl/v4_fusion_bs4x16_mnv64} & 2.14 & \cellcolor{blue!10} 1.88 & 8.9 & 39.7 & 0.376 & \cellcolor{blue!10} 0.569 & \cellcolor{blue!25} 0.877 & 0.874 & 0.58 & 26.4M \\
            \midrule
            \multirow{3}{*}{\textbf{HiSup}~\citep{hisup}} & \textbf{Image} & \detokenize{hisup/v3_image_vit_cnn_bs4x12} & 2.46 & 2.48 & 11.4 & 35.2 & 0.287 & 0.493 & 0.85 & 0.885 & \cellcolor{blue!25} 0.124 & 30.8M \\
             & \textbf{LiDAR} & \detokenize{hisup/lidar_pp_vit_cnn_bs2x16_mnv64} & 2.05 & 2.01 & 9.58 & 37 & 0.347 & 0.54 & 0.87 & 0.882 & 0.165 & 30.8M \\
             & \textbf{Fusion} & \detokenize{hisup/early_fusion_vit_cnn_bs2x16_mnv64} & 1.91 & 1.9 & 8.94 & 35.2 & 0.355 & 0.568 & \cellcolor{blue!10} 0.872 & 0.89 & \cellcolor{blue!10} 0.129 & 33.5M \\
            \midrule
            \multirow{3}{*}{\textbf{Pix2Poly}~\citep{pix2poly}} & \textbf{Image} & \detokenize{pix2poly/v4_image_vit_bs4x16} & 2.46 & 2.5 & 10.8 & 34.3 & 0.317 & 0.492 & 0.845 & 0.906 & 1.44 & 31.9M \\
             & \textbf{LiDAR} & \detokenize{pix2poly/lidar_pp_vit_bs2x16_mnv64} & \cellcolor{blue!10} 1.88 & 1.9 & \cellcolor{blue!10} 8.5 & \cellcolor{blue!10} 34.1 & \cellcolor{blue!10} 0.379 & 0.552 & 0.869 & \cellcolor{blue!10} 0.913 & 1.15 & 31.9M \\
             & \textbf{Fusion} & \detokenize{pix2poly/early_fusion_bs2x16_mnv64} & \cellcolor{blue!25} 1.8 & \cellcolor{blue!25} 1.82 & \cellcolor{blue!25} 8.16 & \cellcolor{blue!25} 33.4 & \cellcolor{blue!25} 0.398 & \cellcolor{blue!25} 0.578 & 0.87 & \cellcolor{blue!25} 0.915 & 1.15 & 34.6M \\
            \bottomrule
            \end{tabular}
    }
            \label{tab:modality_ablation}
    \end{table}

\paragraph{Temporal mismatch.}
In \figureref{fig:temporal_mismatch_prediction}, we show exemplary polygon predictions of P$^3$ in which the image and LiDAR data have a temporal mismatch. 
While this case is rare in our dataset, it shows that the tested fusion models are not yet able to handle temporal mismatches between the image and LiDAR data.

\begin{figure}
\centering
    \definetrim{mytrim}{0 0 0 0}
	\newcommand{\mywidth}{0.25\linewidth}
	\newcommand{\myfontsize}{\scriptsize}
	\setlength{\tabcolsep}{0mm}
    \newcommand{\baseimg}{./images/modality_ablation_figure}
    \newcommand{\imgpath}[1]{\baseimg/#1} 

\newcommand{\inputImages}[4]{%
    &
    \includegraphics[width=\mywidth,mytrim]{\imgpath{#1_test_#2_#3_#4.jpg}} &
    \includegraphics[width=\mywidth,mytrim]{\imgpath{#1_test_#2_pred_lidar.jpg}} &
    \includegraphics[width=\mywidth,mytrim]{\imgpath{#1_test_#2_pred_both.jpg}} &
    \includegraphics[width=\mywidth,mytrim]{\imgpath{#1_test_#2_gt.jpg}} \\
}
\newcommand{\inputPred}[4]{%
    \includegraphics[width=\mywidth,mytrim]{\imgpath{#1_test_#2_#3_#4.jpg}}
}
\newcommand{\inputGt}[3]{%
    \includegraphics[width=\mywidth,mytrim]{\imgpath{#1_test_#2_gt_#3}}
}

    \newcommand{\tileID}{1659}

    \resizebox{0.98\textwidth}{!}{%
        \begin{tabular}{@{}c@{}c@{}c@{}c@{\hspace{2pt}}l@{}}

    \inputGt{Switzerland}{\tileID}{image} &
    \inputPred{Switzerland}{\tileID}{ffl}{v4_image_bs4x16} &
    \inputPred{Switzerland}{\tileID}{hisup}{v3_image_vit_cnn_bs4x12} &
    \inputPred{Switzerland}{\tileID}{pix2poly}{v4_image_vit_bs4x16} &
    \rotatebox{-90}{\hspace{-28mm}Pred. Image} \\
    
    \inputGt{Switzerland}{\tileID}{lidar} &
    \inputPred{Switzerland}{\tileID}{ffl}{v5_lidar_bs2x16_mnv64} &
    \inputPred{Switzerland}{\tileID}{hisup}{lidar_pp_vit_cnn_bs2x16_mnv64} &
    \inputPred{Switzerland}{\tileID}{pix2poly}{lidar_pp_vit_bs2x16_mnv64} &
    \rotatebox{-90}{\hspace{-27mm}Pred. LiDAR} \\
    
    \inputGt{Switzerland}{\tileID}{both} &
    \inputPred{Switzerland}{\tileID}{ffl}{v4_fusion_bs4x16_mnv64} &
    \inputPred{Switzerland}{\tileID}{hisup}{early_fusion_vit_cnn_bs2x16_mnv64} &
    \inputPred{Switzerland}{\tileID}{pix2poly}{early_fusion_bs2x16_mnv64} &
    \rotatebox{-90}{\hspace{-28mm}Pred. Fusion} \\

\makebox[\mywidth][c]{Ground truth} &
\makebox[\mywidth][c]{FFL \citep{ffl}} &
\makebox[\mywidth][c]{HiSup \citep{hisup}} &
\makebox[\mywidth][c]{Pix2Poly \citep{pix2poly}}&\\

\end{tabular}
}
\caption{\textbf{Polygon prediction with temporal mismatch.} We show a tile in which the image and LiDAR data have a temporal mismatch. The image shows two newly constructed buildings, while the LiDAR data is from a previous flight that shows buildings that are most likely demolished. The ground truth polygons align with the LiDAR data and not with the image. The predicted polygons from the LiDAR and image modality only are consistent, while the fused prediction are not accurate and do not match any ground truth polygons.}
\label{fig:temporal_mismatch_prediction}
\end{figure}

\paragraph{LiDAR point density.}

\begin{figure}

    \renewcommand{\arraystretch}{0.75}

	\definetrim{mytrim}{0 0 0 0}
	\newcommand{\mywidth}{0.24\linewidth}
	\newcommand{\myfontsize}{\scriptsize}
	\setlength{\tabcolsep}{0mm}
    \newcommand{\baseimg}{./images/lidar_density_ablation_figure}
    \newcommand{\imgpath}[1]{\baseimg/#1} 

\newcommand{\inputImages}[4]{%
    &
    \includegraphics[width=\mywidth,mytrim]{\imgpath{#1_test_#2_#3_#4.jpg}} &
    \includegraphics[width=\mywidth,mytrim]{\imgpath{#1_test_#2_pred_lidar.jpg}} &
    \includegraphics[width=\mywidth,mytrim]{\imgpath{#1_test_#2_pred_both.jpg}} &
    \includegraphics[width=\mywidth,mytrim]{\imgpath{#1_test_#2_gt.jpg}} \\
}
\newcommand{\inputPred}[3]{%
    \includegraphics[width=\mywidth,mytrim]{\imgpath{#1_test_#2_#3.png}}
}
\newcommand{\inputGt}[3]{%
    \includegraphics[width=\mywidth,mytrim]{\imgpath{#1_test_#2_gt_#3}}
}
    \centering
    \resizebox{\textwidth}{!}{
    \begin{tabular}{@{}ccccc@{}}
    
    \inputGt{Switzerland}{9722}{both} &
    \inputPred{Switzerland}{9722}{0} & 
    \inputPred{Switzerland}{9722}{2} & 
    \inputPred{Switzerland}{9722}{4} & 
    \inputPred{Switzerland}{9722}{5} 
    \\
    
    Ground truth & 1 pt/m$^2$ & 16 pt/m$^2$ & 64 pt/m$^2$ & 128 pt/m$^2$ \\

\end{tabular}
    }
\caption{\textbf{LiDAR density ablation.} We present a sample tile displaying predicted and ground truth building polygons from the Switzerland subset with different point densities. Only LiDAR points are used for the prediction.}
\label{fig:lidar_density_ablation}
\end{figure}

The point clouds from Switzerland have an average density between 11 and 17 pts/m$^2$. However, locally, the density can be much higher, \eg on vertical features such as building walls. We train a FFL model with a LiDAR point encoder and input 1, 4, 16, 32, 64 and 128 pts/m$^2$. We achieve this by adapting the maximum number of points during the voxelization step in our point cloud encoder. Voxels with fewer points are upsampled to the maximum number of points, while voxels with more points are downsampled \citep{pointpillars}. \figureref{fig:lidar_density_ablation} shows example point clouds and predicted polygons for different point densities. The model trained with 1 pt/m$^2$ is able to predict the building outlines, but the polygons are less accurate and more complex than the ones predicted with higher point densities. Higher point densities lead to more accurate and simpler polygons.

\begin{table}
    \caption{\textbf{Density ablation}. We compare a FFL~\citep{ffl} model trained and tested on increasingly dense LiDAR point clouds. We use 16 pts/m$^2$ as the default density for all other experiments.}
    \label{tab:density_ablation}
    \setlength{\tabcolsep}{3pt}
    \centering
    \resizebox{0.85\textwidth}{!}{
        \begin{tabular}{@{}cH|cccccccccc@{}}
            \toprule
                & & \multicolumn{4}{c}{\emph{Boundary}}  & \multicolumn{3}{c}{\emph{Area}}  &  \multicolumn{3}{c}{\emph{Complexity}} \\
            \midrule
            \textbf{Density [pts/m$^2$]} & Unnamed: 0 & \textbf{POLIS [m]} $\downarrow$ & \textbf{CD [m]} $\downarrow$ & \textbf{HD [m]} $\downarrow$ & \textbf{MTA [$^\circ$]} $\downarrow$ & \textbf{AP} $\uparrow$ & \textbf{AR} $\uparrow$ & \textbf{IoU} $\uparrow$ & \textbf{C-IoU} $\uparrow$ & \textbf{NR=1} & \textbf{DoF} $\downarrow$ \\
            \midrule
            1 & \detokenize{ffl/v5_lidar_bs2x16_mnv4} & 2.77 & 2.24 & 11.2 & 48.2 & 0.293 & 0.487 & 0.847 & 0.714 & 0.78 & 0.868 \\
            4 & \detokenize{ffl/v5_lidar_bs2x16_mnv16} & 2.55 & 2.03 & 10.1 & 46.5 & 0.34 & 0.519 & 0.865 & 0.746 & 0.812 & 0.869 \\
            16 & \detokenize{ffl/v5_lidar_bs2x16_mnv64} & 2.35 & 1.94 & 9.66 & 44.8 & 0.359 & 0.545 & 0.87 & 0.765 & 0.829 & 0.866 \\
            32 & \detokenize{ffl/v5_lidar_bs2x16_mnv128} & 2.23 & \cellcolor{blue!10} 1.85 & \cellcolor{blue!10} 9.2 & 43.1 & \cellcolor{blue!10} 0.363 & 0.553 & 0.874 & 0.774 & 0.838 & \cellcolor{blue!10} 0.859 \\
            64 & \detokenize{ffl/v5_lidar_bs2x16_mnv256} & \cellcolor{blue!25} 2.17 & \cellcolor{blue!25} 1.83 & \cellcolor{blue!25} 9.07 & \cellcolor{blue!10} 41.9 & \cellcolor{blue!25} 0.373 & \cellcolor{blue!25} 0.56 & \cellcolor{blue!25} 0.877 & \cellcolor{blue!25} 0.785 & \cellcolor{blue!25} 0.849 & \cellcolor{blue!10} 0.859 \\
            128 & \detokenize{ffl/v5_lidar_bs2x16_mnv512} & \cellcolor{blue!10} 2.21 & 1.9 & 9.24 & \cellcolor{blue!25} 41 & 0.362 & \cellcolor{blue!25} 0.56 & \cellcolor{blue!10} 0.876 & \cellcolor{blue!10} 0.783 & \cellcolor{blue!10} 0.846 & \cellcolor{blue!25} 0.842 \\
            \bottomrule
            \end{tabular}
    }
    \end{table}
However, even a model trained with 1 pt/m$^2$ leads to a mean IoU of 84.7$\%$ (see \tableref{tab:density_ablation}). This shows that even the sparse 3D information is suFfficient to detect most buildings in our dataset. Increasing the point density until 64 pts/m$^2$ improves all metrics, \ie it allows the model to predict more complete and accurate, but also simpler polygons, with a more accurate number of vertices compared to the ground truth. At 128 pts/m$^2$, the model performance starts to decline again for some of the metrics. We attribute this to the fact that only very few voxels have a density of 128 pts/m$^2$. Our point cloud encoder thus upsamples most voxels to this point density by duplicating existing points. This may lead to a slower convergence of the model and a less accurate prediction.

\paragraph{Image ground sampling distance.}

The original Switzerland data contains images with a \ac{gsd} of 10~cm. We prepare a subset of the data with a \ac{gsd} of 15~cm and compare the results with a model trained on images with a \ac{gsd} of 25~cm. The \ac{gsd} of 25~cm is the one used in our benchmark dataset. We train a FFL model with a image encoder and input images with a \ac{gsd} of 15 and 25~cm. \tableref{tab:gsd_ablation} shows the results of the ablation study. The model trained on images with a \ac{gsd} of 15~cm achieves better results on the area-based metrics while the one trained on images with 25~cm GSD achieves better results on all other metrics. We attribute this to the fact that the model trained on 15~cm GSD images has access to a smaller context window (see \secref{fig:gsd_ablation}).

\begin{table}
        \caption{\textbf{Ground sampling distance ablation}. We compare a FFL~\citep{ffl} model trained and tested on aerial images with a GSD of 15 and 25~cm. We use images with a 25~cm GSD as the default resolution for all other experiments.}
    \label{tab:gsd_ablation}
    \setlength{\tabcolsep}{3pt}
    \centering
    \resizebox{0.8\textwidth}{!}{
        \begin{tabular}{@{}cH|cccccccccc@{}}
            \toprule
                & & \multicolumn{4}{c}{\emph{Boundary}}  & \multicolumn{3}{c}{\emph{Area}}  &  \multicolumn{3}{c}{\emph{Complexity}} \\
            \midrule
            \textbf{GSD [cm]} & Unnamed: 0 & \textbf{POLIS [m]} $\downarrow$ & \textbf{CD [m]} $\downarrow$ & \textbf{HD [m]} $\downarrow$ & \textbf{MTA [$^\circ$]} $\downarrow$ & \textbf{AP} $\uparrow$ & \textbf{AR} $\uparrow$ & \textbf{IoU} $\uparrow$ & \textbf{C-IoU} $\uparrow$ & \textbf{NR=1} & \textbf{DoF} $\downarrow$ \\
            \midrule
            15 & \detokenize{ffl/ffl_image_015} & \cellcolor{blue!10} 3.98 & \cellcolor{blue!10} 3.48 & \cellcolor{blue!10} 15.9 & \cellcolor{blue!10} 44.5 & \cellcolor{blue!25} 0.279 & \cellcolor{blue!25} 0.475 & \cellcolor{blue!25} 0.862 & \cellcolor{blue!25} 0.764 & \cellcolor{blue!10} 0.823 & \cellcolor{blue!10} 0.842 \\
            25 & \detokenize{ffl/v4_image_bs4x16} & \cellcolor{blue!25} 3 & \cellcolor{blue!25} 2.7 & \cellcolor{blue!25} 12 & \cellcolor{blue!25} 41 & \cellcolor{blue!10} 0.275 & \cellcolor{blue!10} 0.46 & \cellcolor{blue!10} 0.839 & \cellcolor{blue!10} 0.759 & \cellcolor{blue!25} 0.847 & \cellcolor{blue!25} 0.826 \\
            \bottomrule
            \end{tabular}
    }
    \end{table}

\begin{figure}

	\definetrim{mytrim}{0 0 0 0}
	\newcommand{\mywidth}{0.24\linewidth}
	\newcommand{\myfontsize}{\scriptsize}
	\setlength{\tabcolsep}{2mm}
    \newcommand{\baseimg}{./images/gsd_ablation_figure}
    \newcommand{\imgpath}[1]{\baseimg/#1} 

\newcommand{\inputImages}[4]{%
    &
    \includegraphics[width=\mywidth,mytrim]{\imgpath{#1_val_#2_#3_#4.jpg}} &
    \includegraphics[width=\mywidth,mytrim]{\imgpath{#1_val_#2_pred_lidar.jpg}} &
    \includegraphics[width=\mywidth,mytrim]{\imgpath{#1_val_#2_pred_both.jpg}} &
    \includegraphics[width=\mywidth,mytrim]{\imgpath{#1_val_#2_gt.jpg}} \\
}
\newcommand{\inputPred}[4]{%
    \includegraphics[width=\mywidth,mytrim]{\imgpath{#1_val_#2_#3_#4.jpg}}
}
\newcommand{\inputGt}[3]{%
    \includegraphics[width=\mywidth,mytrim]{\imgpath{#1_val_#2_gt_#3}}
}
    \centering
    \resizebox{0.8\textwidth}{!}{
    \begin{tabular}{@{}ccc@{}}
    
    \inputGt{Switzerland}{0}{image} &
    \inputPred{Switzerland}{0}{ffl}{image_015} &
    \inputPred{Switzerland}{0}{ffl}{v4_image_bs4x16}
    \\
    
    Ground truth & 15~cm GSD & 25~cm GSD \\

\end{tabular}
    }
\caption{\textbf{Ground sampling distance ablation.} We present a sample tile displaying predicted and ground truth building polygons from the Switzerland subset with different point densities. The models are trained on images with a GSD of 15 and 25 cm which leads to different input tiles and predictions.}
\label{fig:gsd_ablation}
\end{figure}

\subsection{Comparison with existing image datasets}

\begin{figure}
	\centering
	\begin{subfigure}[b]{0.63\textwidth}
		\centering
		\vspace{0pt}
		\includegraphics[width=\linewidth]{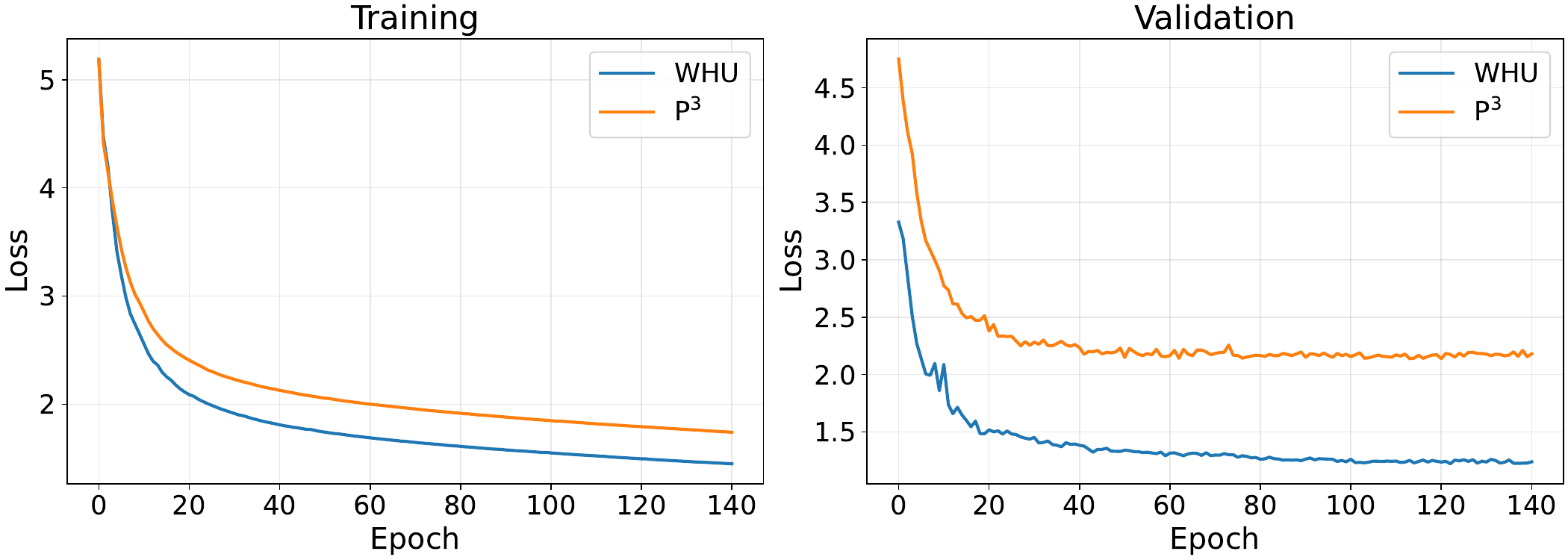}
		\vspace{-4mm}
		\caption{Training and validation loss}\label{fig:loss_curves}
	\end{subfigure}
	\hfill
	\begin{subfigure}[b]{0.33\textwidth}
		\centering
		\vspace{0pt}
		\small
		\resizebox{\linewidth}{!}{\begin{tabular}{@{}l|cccc@{}}
\toprule
 & \multicolumn{4}{c}{\emph{Tested}} \\
\cmidrule(lr){2-5}
\emph{Trained} & \textbf{WHU} & \textbf{P$^3$} & \textbf{Inria} & \textbf{SpaceNet2} \\
\midrule
\textbf{WHU} & \cellcolor{blue!25} 0.941 & \cellcolor{blue!10} 0.681 & \cellcolor{blue!10} 0.653 & \cellcolor{blue!10} 0.522 \\
\textbf{P$^3$} & \cellcolor{blue!10} 0.813 & \cellcolor{blue!25} 0.777 & \cellcolor{blue!25} 0.695 & \cellcolor{blue!25} 0.538 \\
\bottomrule
\end{tabular}
}
		\vspace{10mm}
		\caption{IoU scores}\label{fig:datasets_iou}
	\end{subfigure}
	\caption{\textbf{Related dataset training.} We train Pix2Poly on images of \dsname and the WHU dataset \citep{whu_dataset}. We then test the methods on those datasets, as well as the INRIA dataset \citep{inria_dataset} and the SpaceNet2 Vegas subset \citep{spacenet}. In (a), we show the training and validation loss curves. In (b), we show the IoU scores. The model trained on \dsname predicts the other datasets with a higher IoU score compared to the models trained on the WHU dataset, except for the WHU dataset itself.}\label{fig:training_comparison}
\end{figure}

{In this experiment, we investigate how our dataset compares to existing image datasets. We train Pix2Poly on images of \dsname and the WHU dataset \citep{whu_dataset}. \figref{fig:loss_curves} shows the train and validation loss curves. The model trained on the WHU dataset reaches a lower training and validation loss compared to the model trained on \dsname, which indicates that the WHU dataset is easier to learn for the model. We now test the two trained models on those datasets, as well as the INRIA dataset \citep{inria_dataset} and the SpaceNet2 Vegas subset \citep{spacenet} to investigate how they perform on out-of-distribution data. \figref{fig:datasets_iou} shows the IoU scores of the two models on the four datasets. This shows that our dataset is more challenging than the WHU dataset, but also that it provides a better training set for generalization to other datasets.} 

\begin{figure}
    \centering
    \definetrim{mytrim}{0 0 0 0}
    \setlength{\tabcolsep}{0mm}

    \begin{subfigure}[t]{0.49\textwidth}
        \centering
        \newcommand{\datasetcompimgwidth}{0.32\linewidth}
\newcommand{\datasetcompcolsep}{1mm}
\begin{tabular}{@{}c@{\hspace{\datasetcompcolsep}}c@{\hspace{\datasetcompcolsep}}c@{}}
\scriptsize

\includegraphics[width=\datasetcompimgwidth,mytrim]{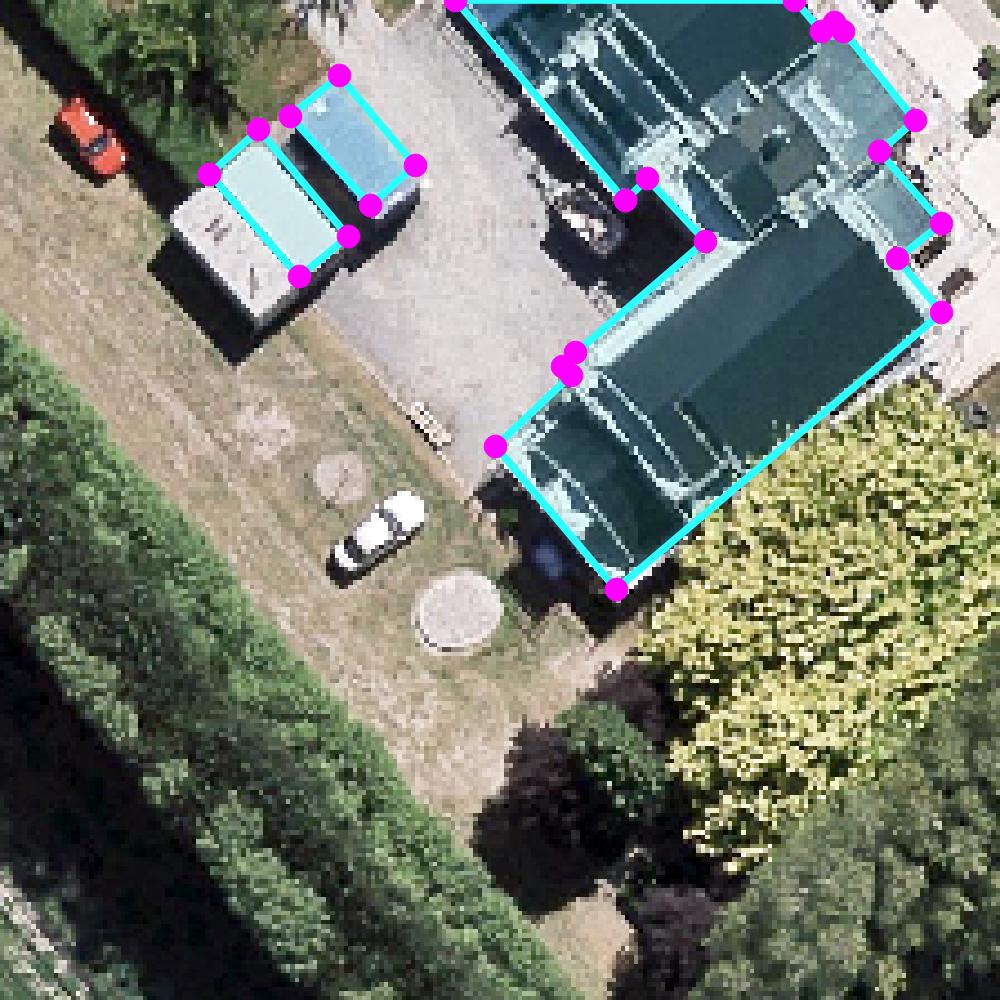} &
\includegraphics[width=\datasetcompimgwidth,mytrim]{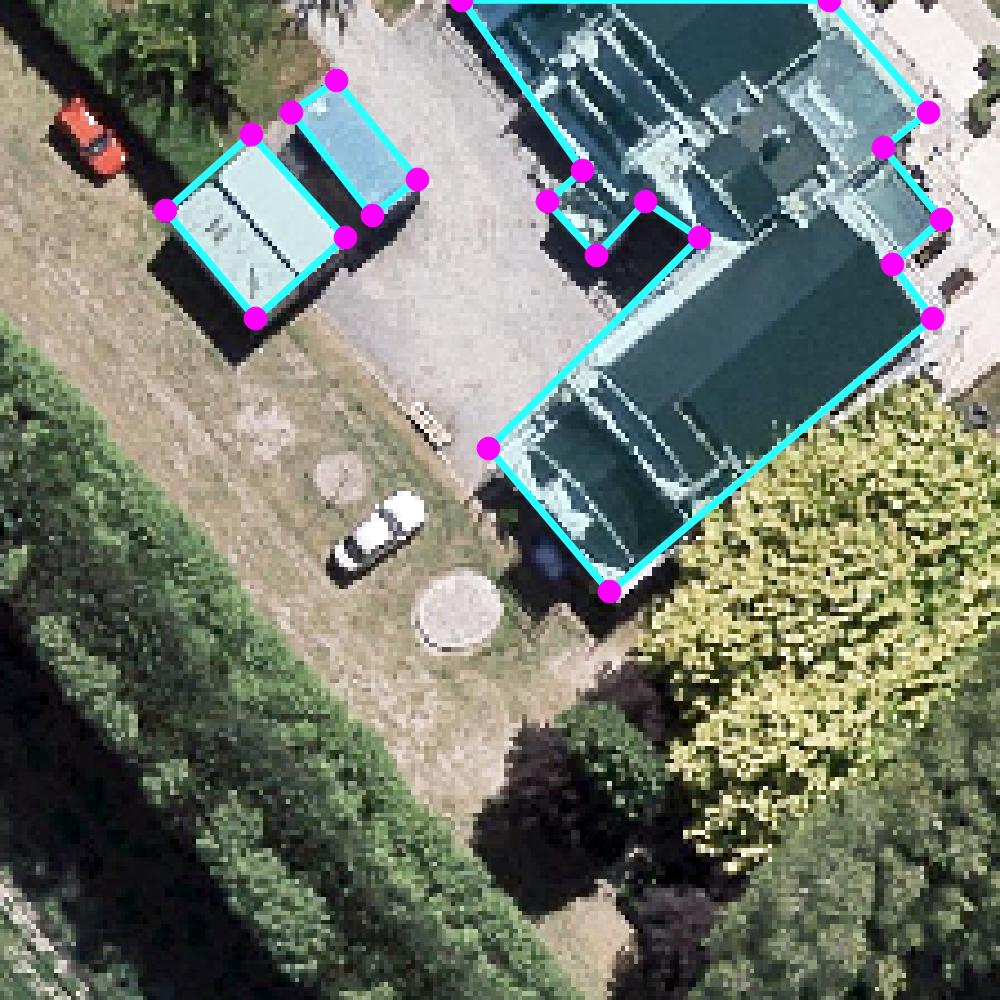} &
\includegraphics[width=\datasetcompimgwidth,mytrim]{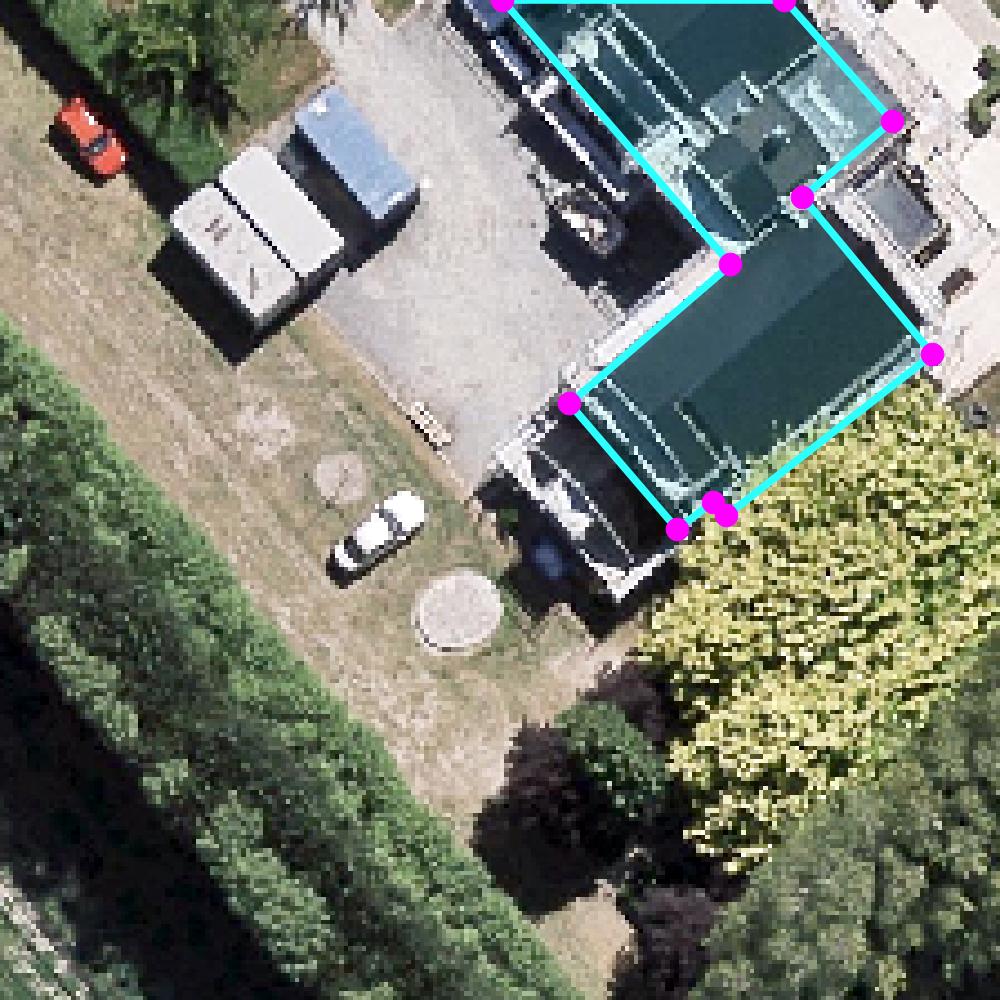} \\

\includegraphics[width=\datasetcompimgwidth,mytrim]{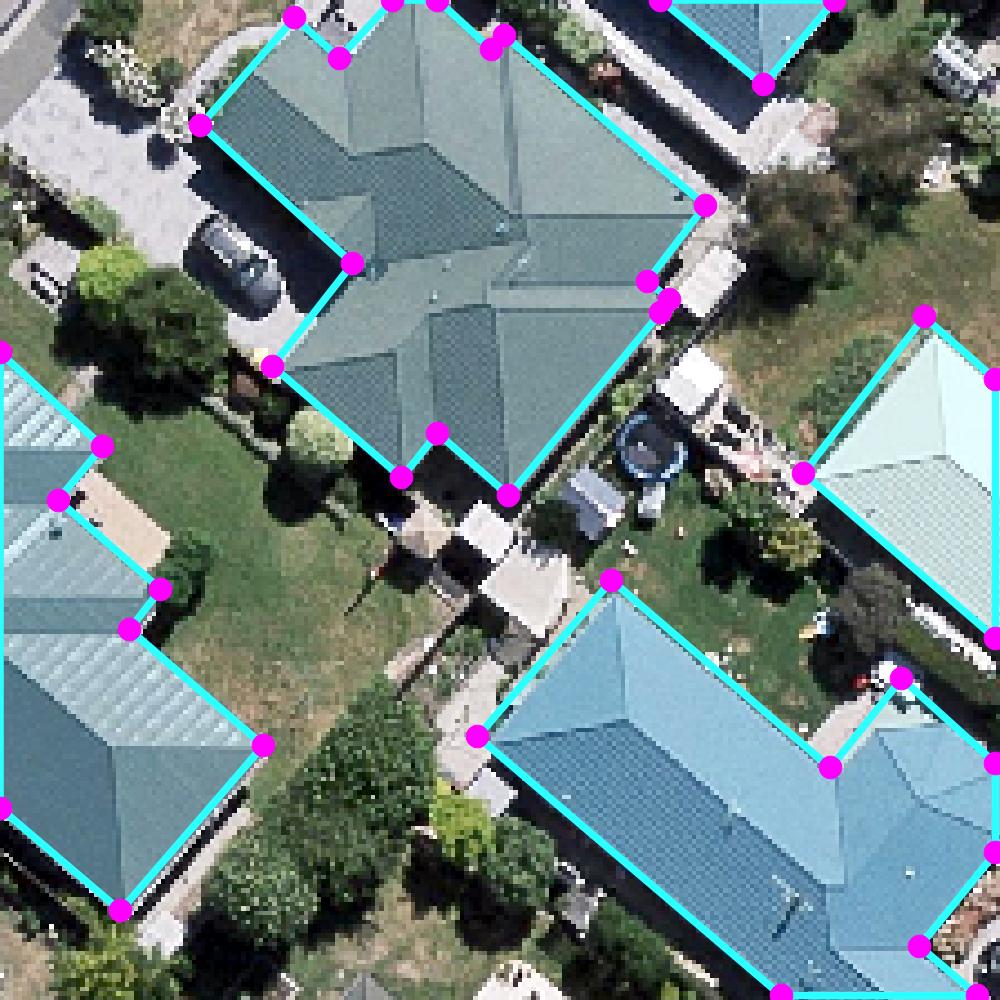} &
\includegraphics[width=\datasetcompimgwidth,mytrim]{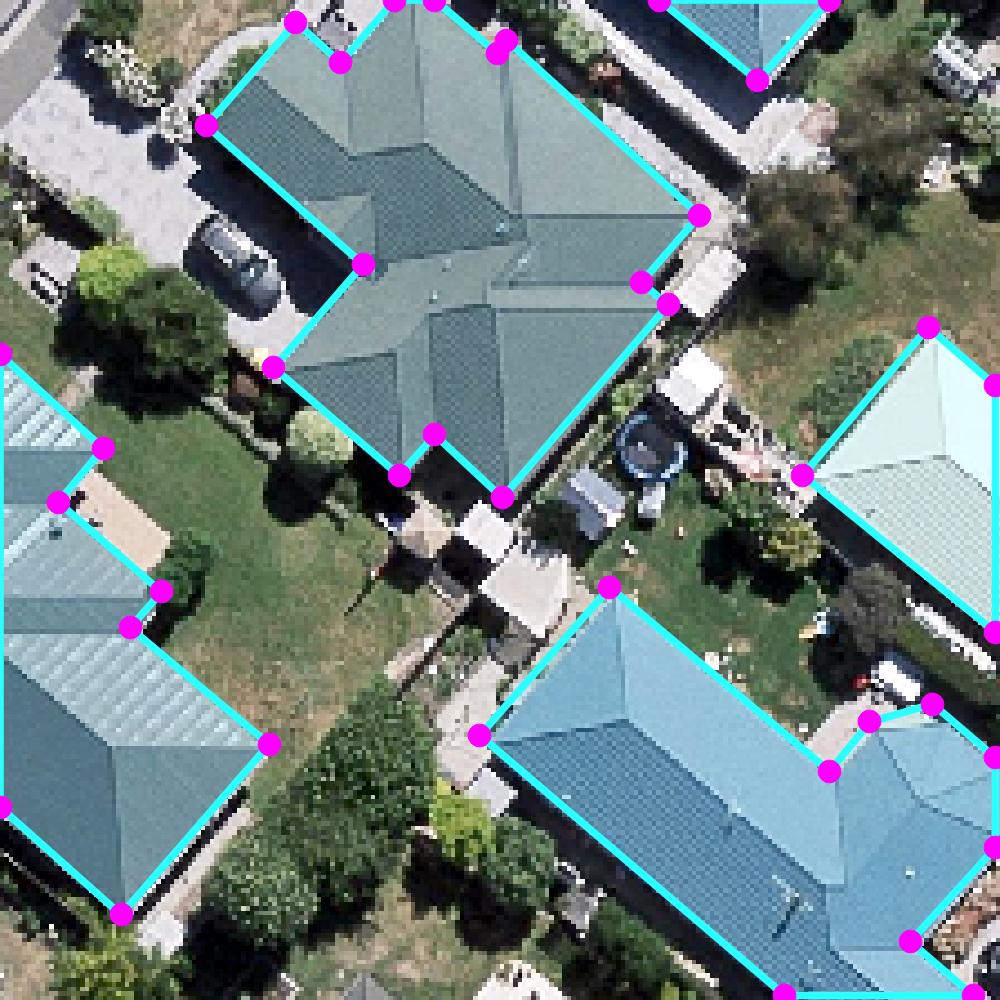} &
\includegraphics[width=\datasetcompimgwidth,mytrim]{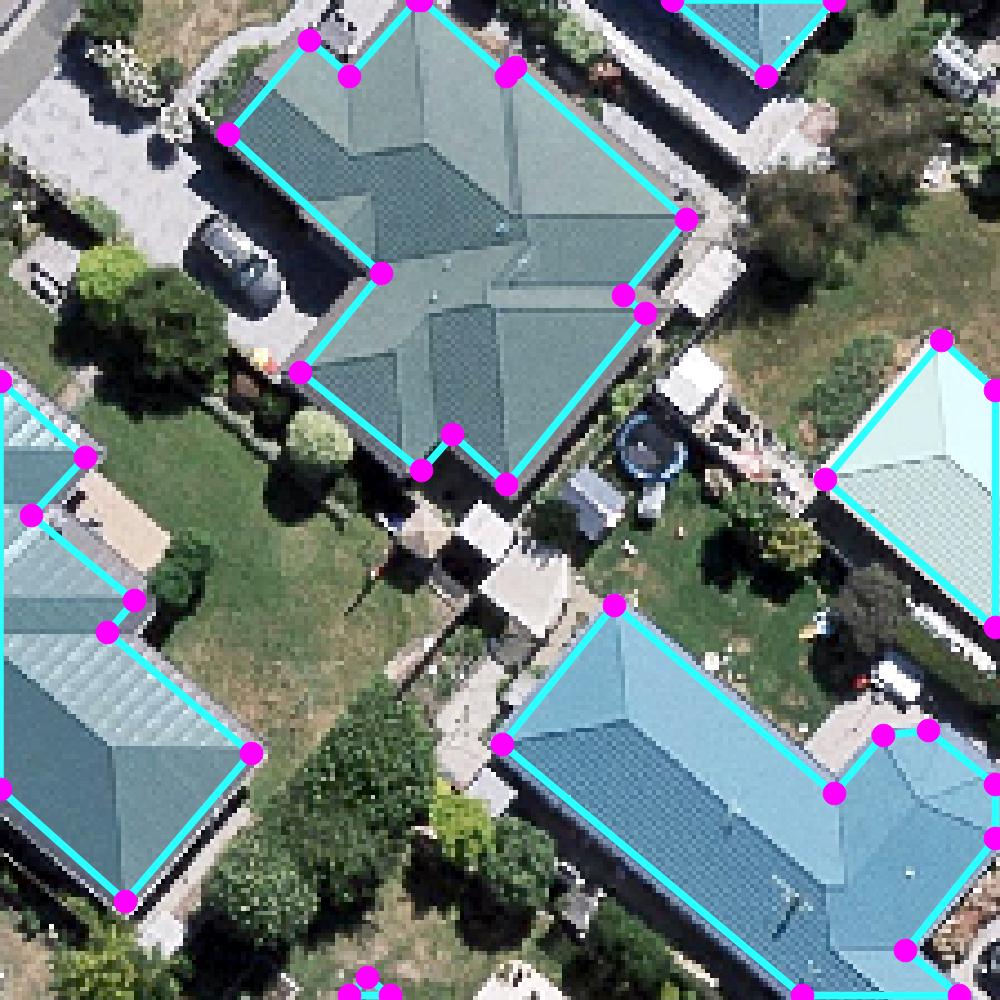} \\

\makebox[\datasetcompimgwidth][c]{Ground truth} &
\makebox[\datasetcompimgwidth][c]{Trained on WHU} &
\makebox[\datasetcompimgwidth][c]{Trained on \dsname}\\
\end{tabular}

        \caption{WHU~\citep{whu_dataset}}\label{fig:dataset_comparison_whu}
    \end{subfigure}
    \begin{subfigure}[t]{0.49\textwidth}
        \centering
        \newcommand{\datasetcompimgwidth}{0.32\linewidth}
\newcommand{\datasetcompcolsep}{1mm}
\begin{tabular}{@{}c@{\hspace{\datasetcompcolsep}}c@{\hspace{\datasetcompcolsep}}c@{}}
\scriptsize

\includegraphics[width=\datasetcompimgwidth,mytrim]{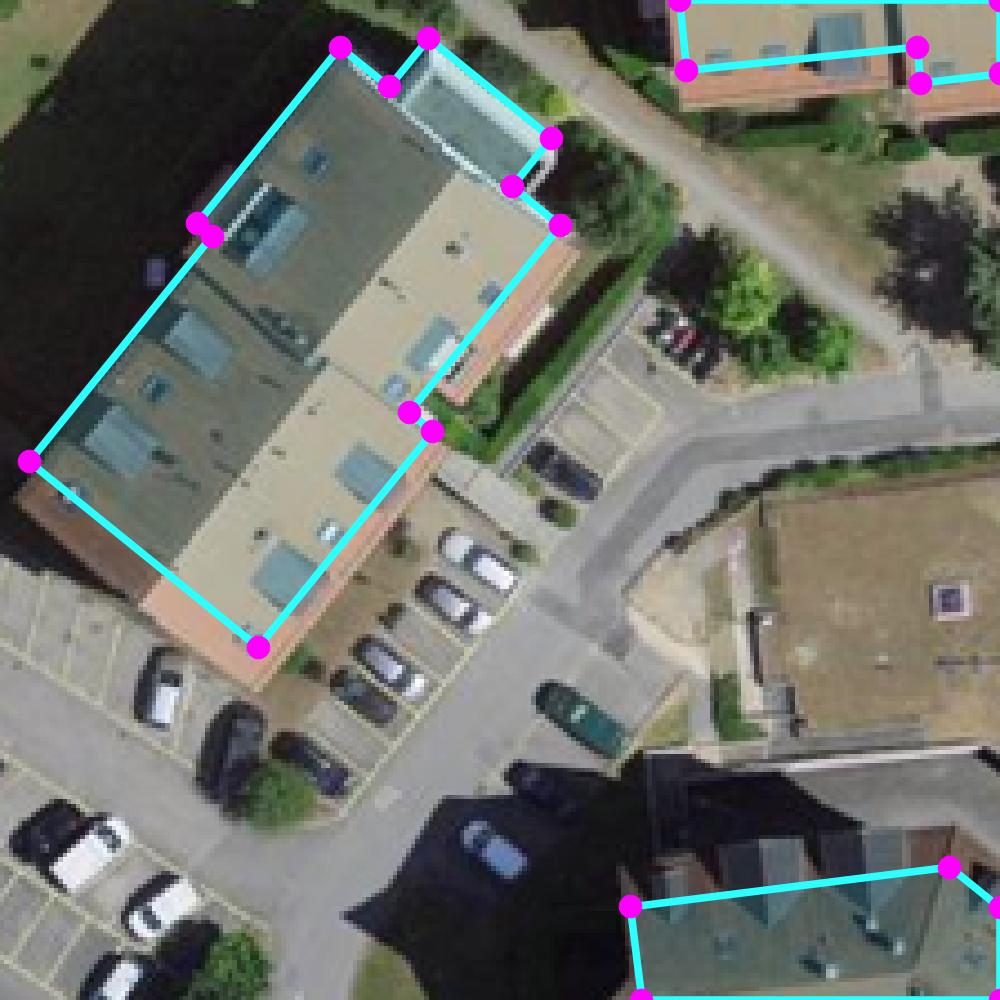} &
\includegraphics[width=\datasetcompimgwidth,mytrim]{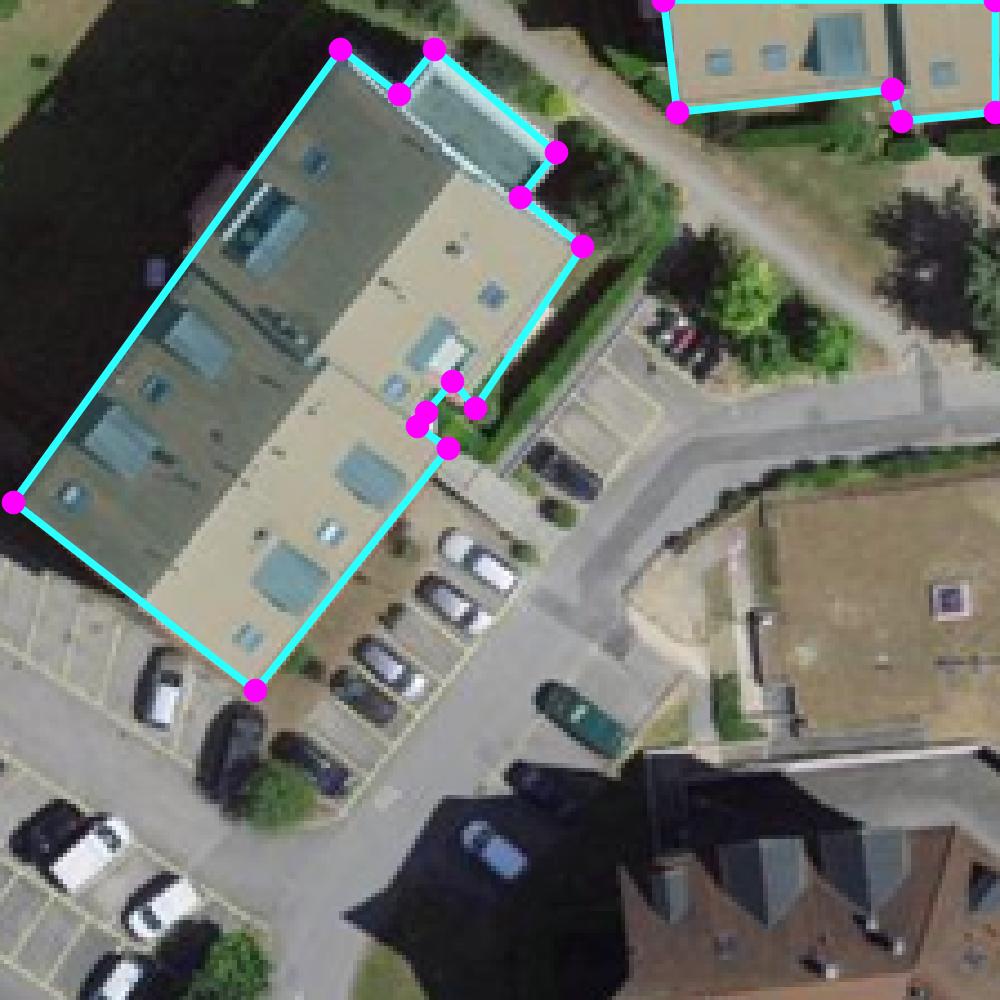} &
\includegraphics[width=\datasetcompimgwidth,mytrim]{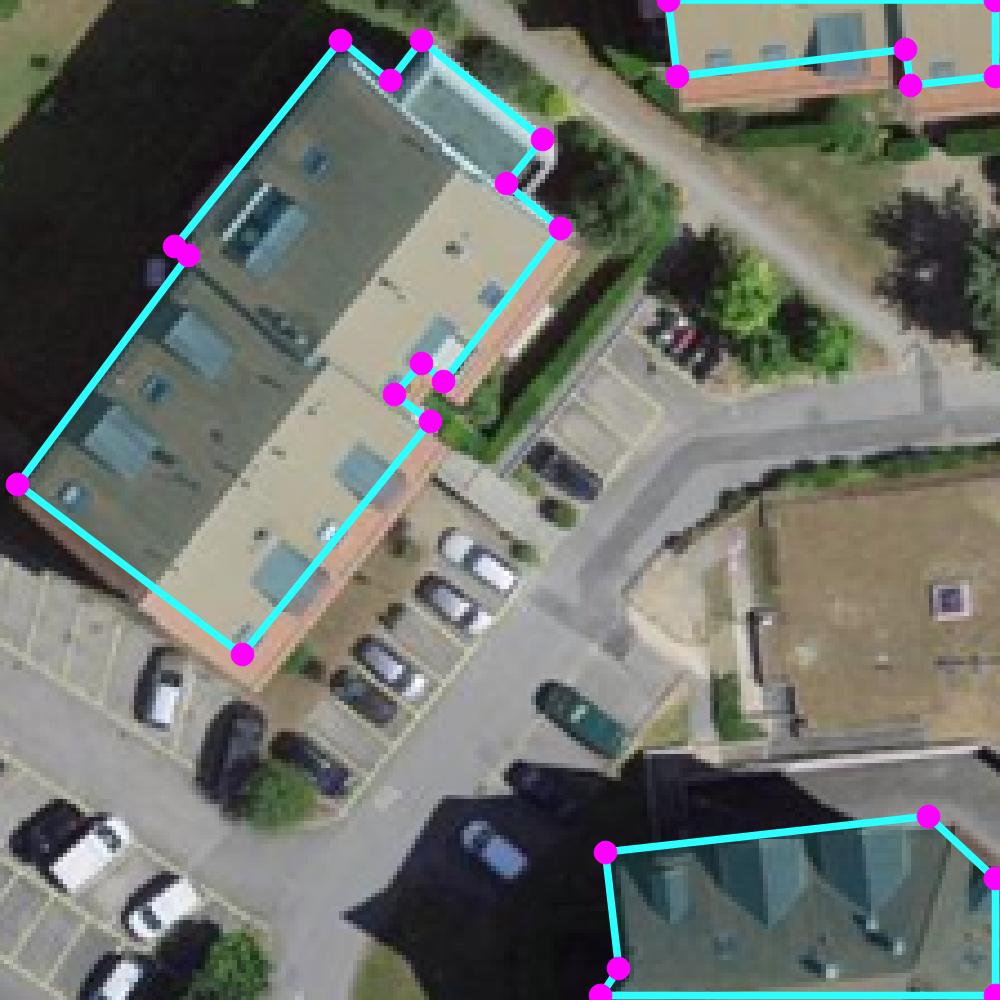} \\

\includegraphics[width=\datasetcompimgwidth,mytrim]{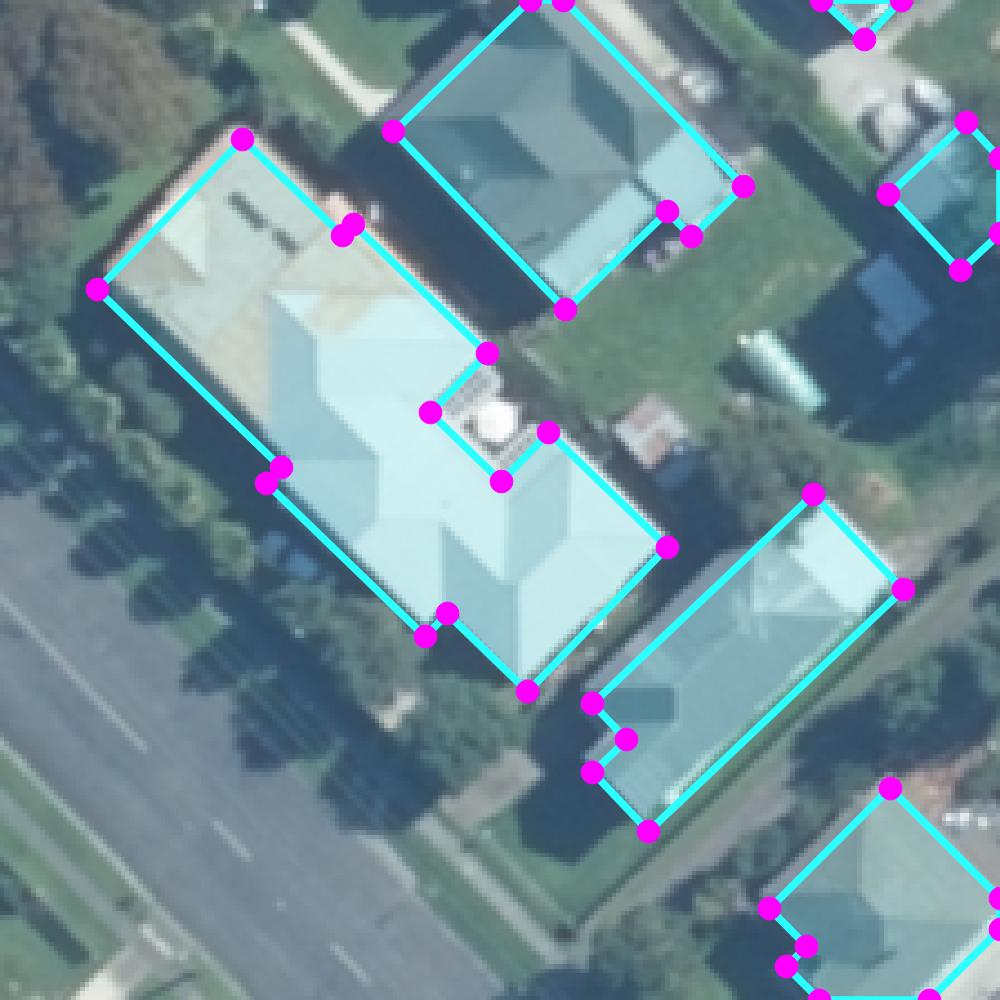} &
\includegraphics[width=\datasetcompimgwidth,mytrim]{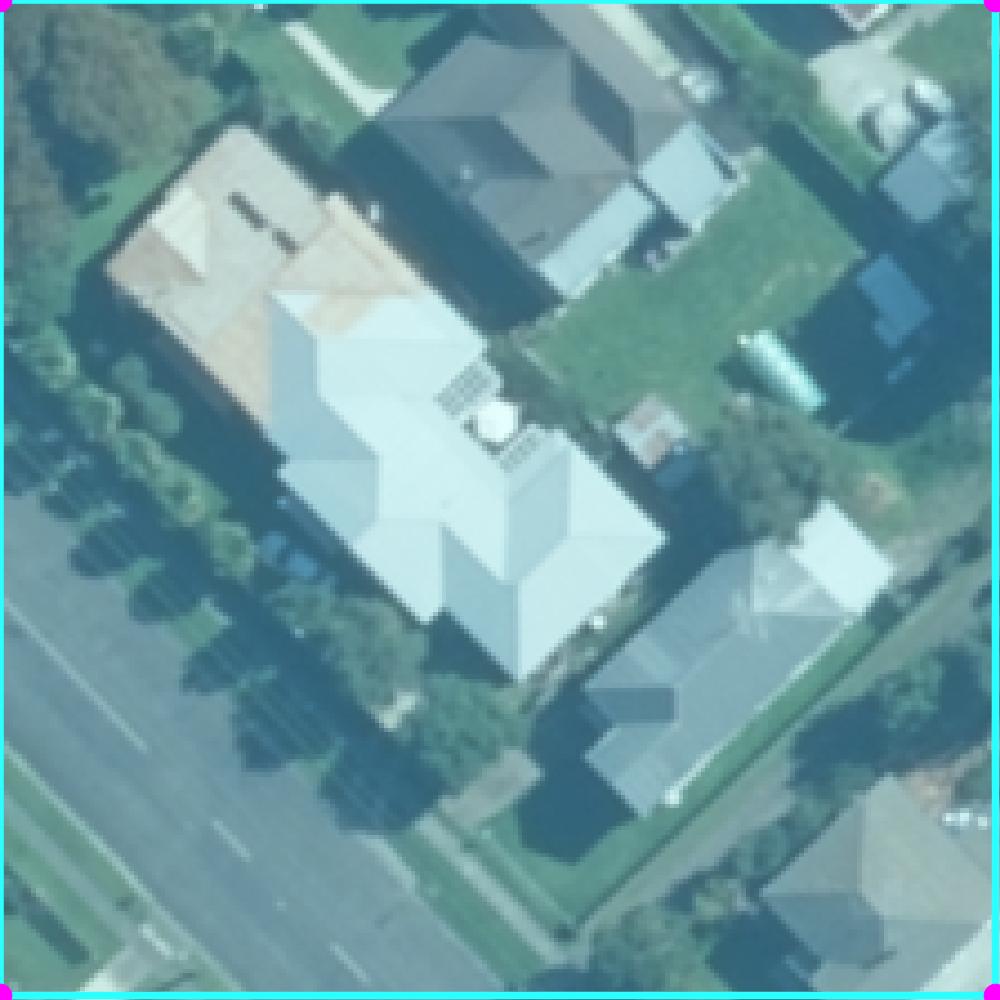} &
\includegraphics[width=\datasetcompimgwidth,mytrim]{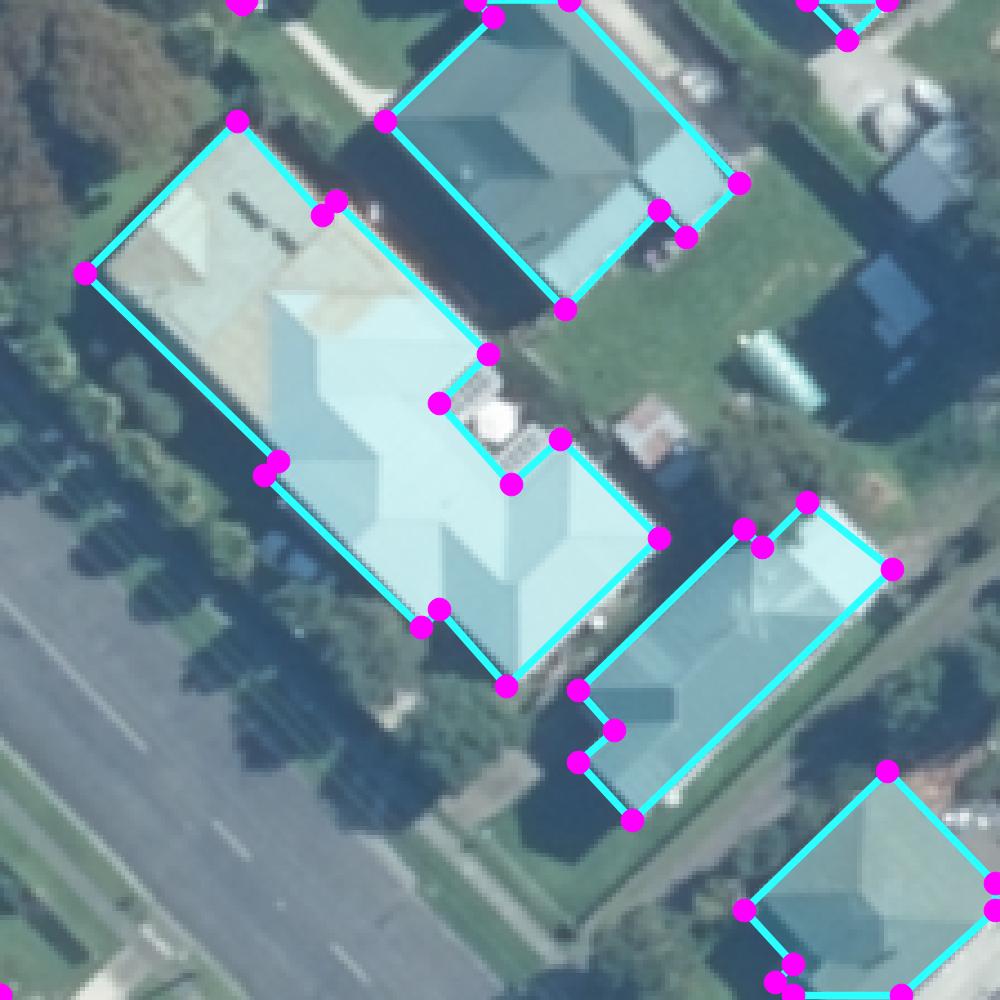} \\

\makebox[\datasetcompimgwidth][c]{Ground truth} &
\makebox[\datasetcompimgwidth][c]{Trained on WHU} &
\makebox[\datasetcompimgwidth][c]{Trained on \dsname}\\
\end{tabular}

        \caption{P$^3$}\label{fig:dataset_comparison_p3}
    \end{subfigure}
\par\vspace{5mm}
    \begin{subfigure}[t]{0.49\textwidth}
        \centering
        \newcommand{\datasetcompimgwidth}{0.32\linewidth}
\newcommand{\datasetcompcolsep}{1mm}
\begin{tabular}{@{}c@{\hspace{\datasetcompcolsep}}c@{\hspace{\datasetcompcolsep}}c@{}}
\scriptsize

\includegraphics[width=\datasetcompimgwidth,mytrim]{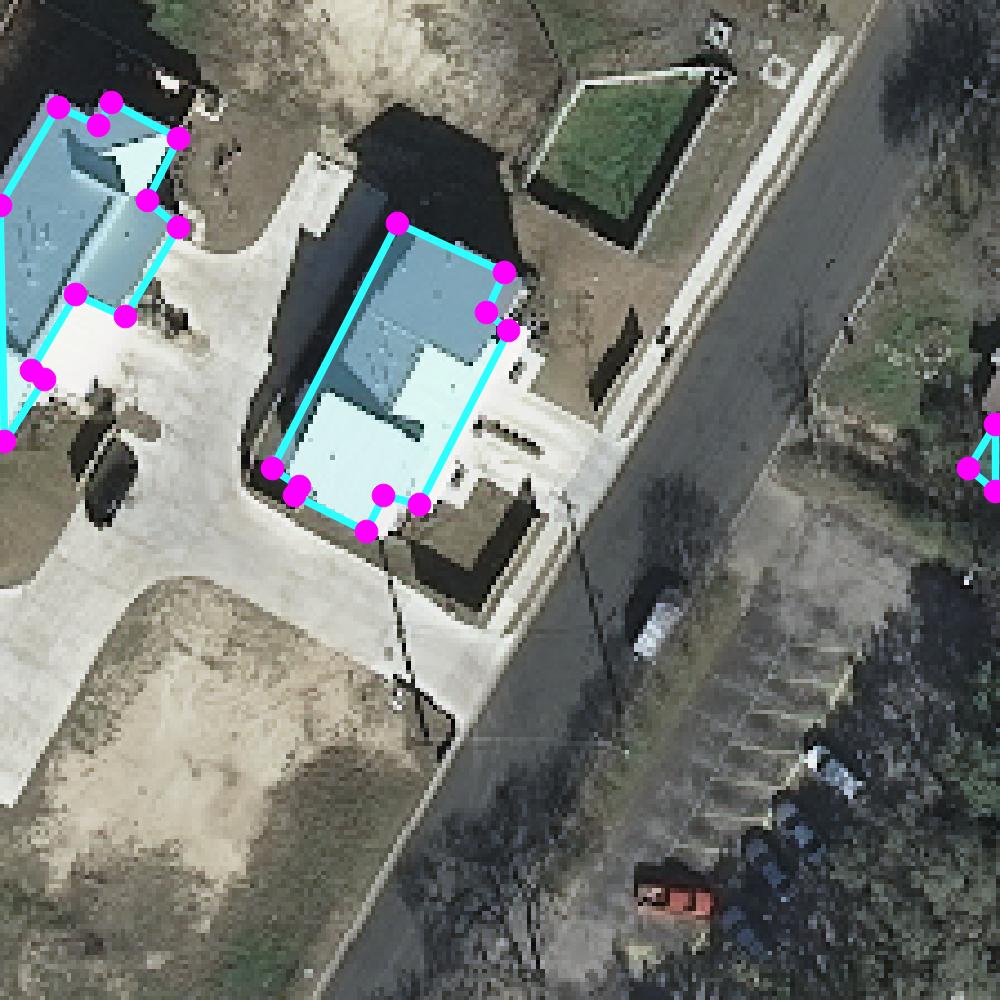} &
\includegraphics[width=\datasetcompimgwidth,mytrim]{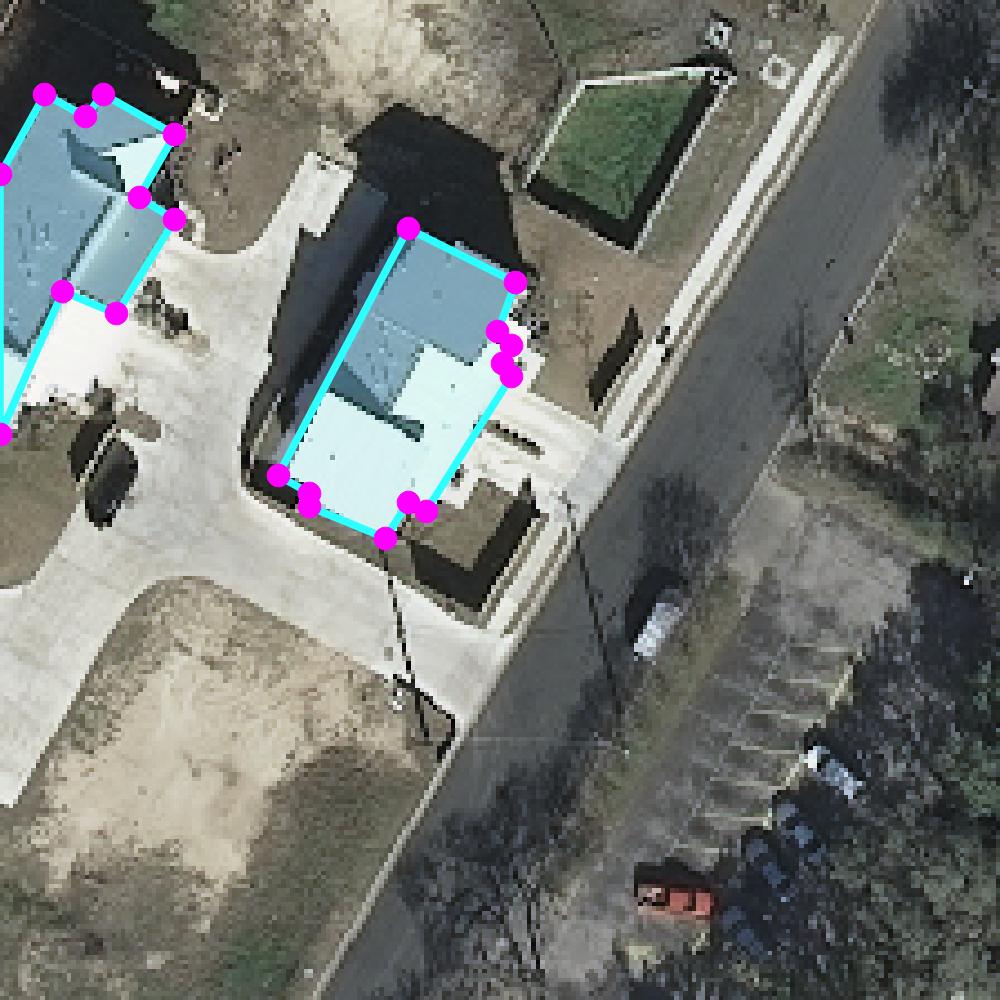} &
\includegraphics[width=\datasetcompimgwidth,mytrim]{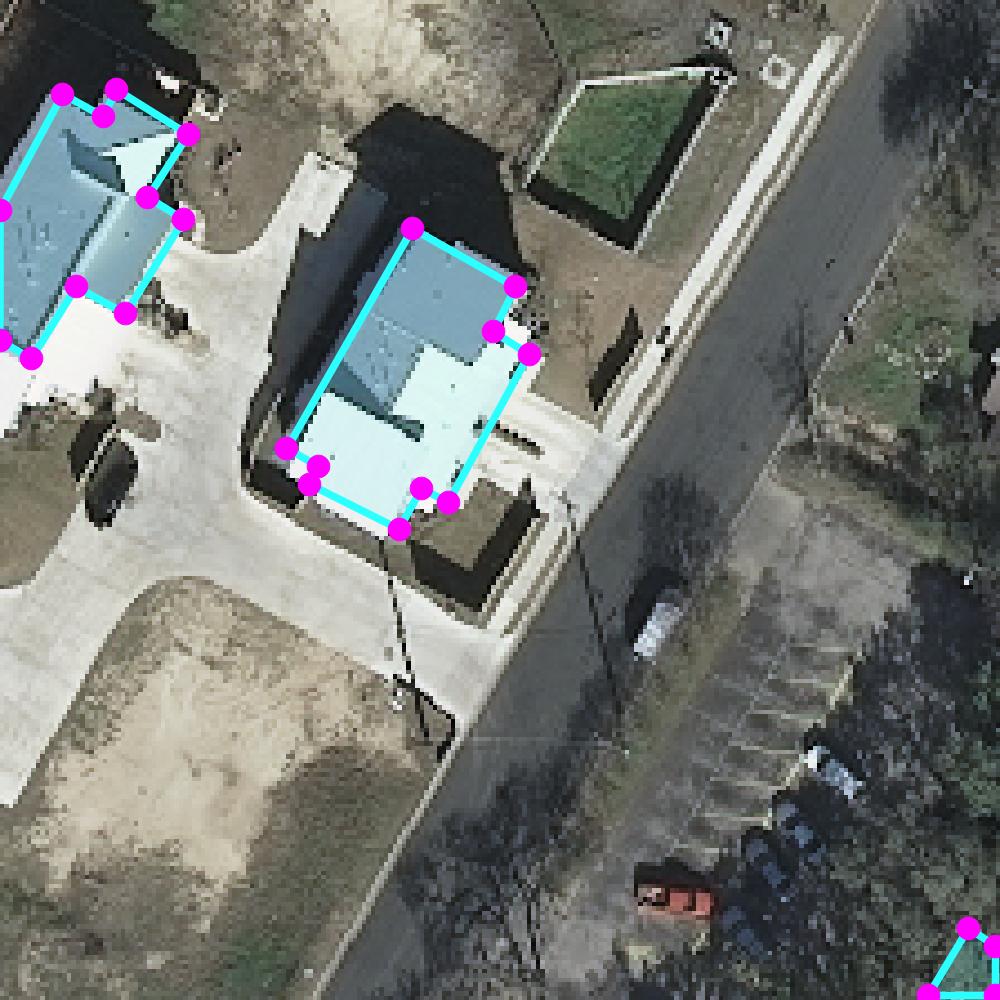} \\

\includegraphics[width=\datasetcompimgwidth,mytrim]{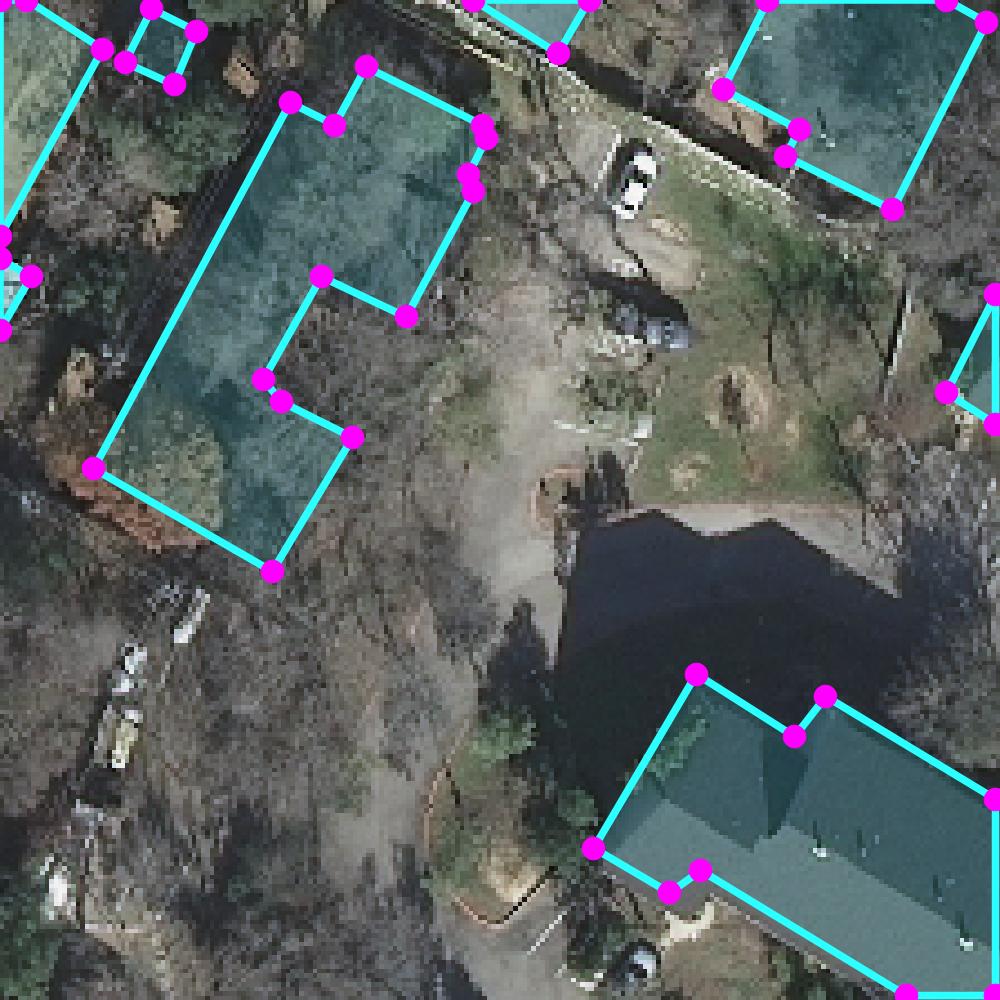} &
\includegraphics[width=\datasetcompimgwidth,mytrim]{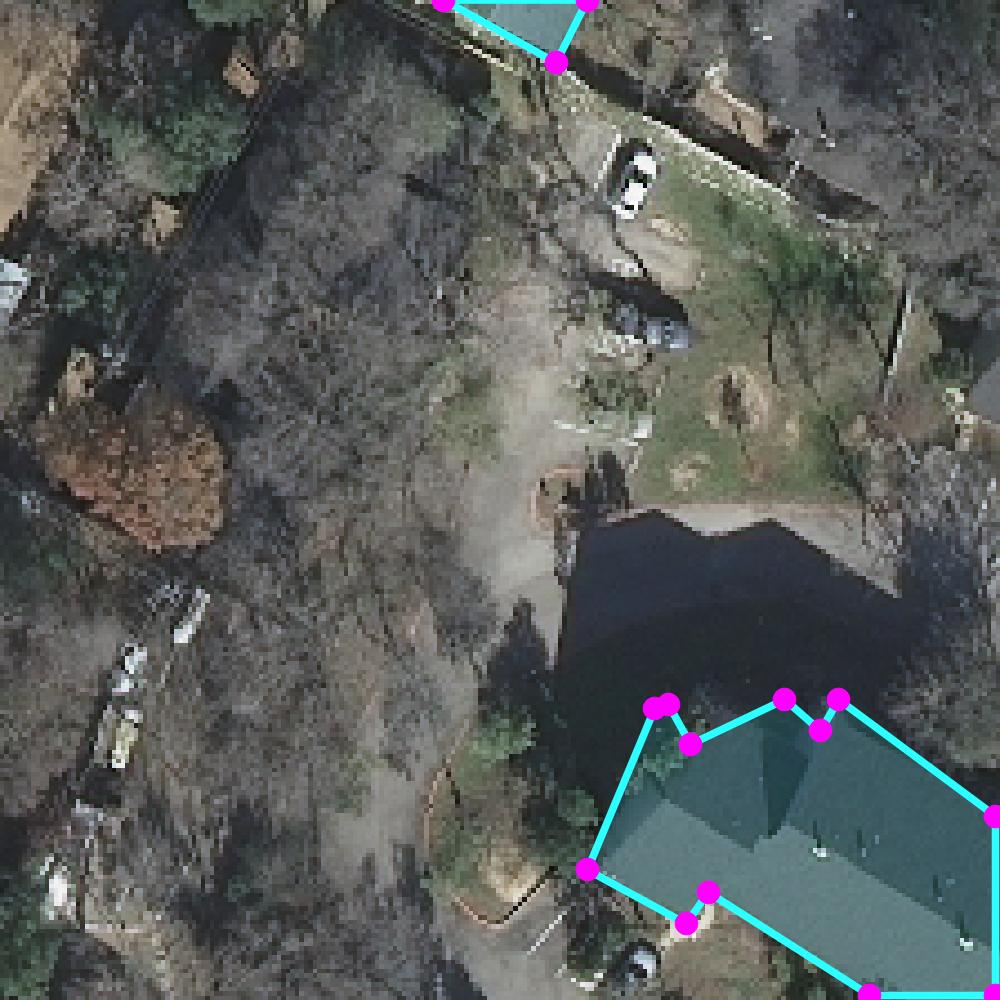} &
\includegraphics[width=\datasetcompimgwidth,mytrim]{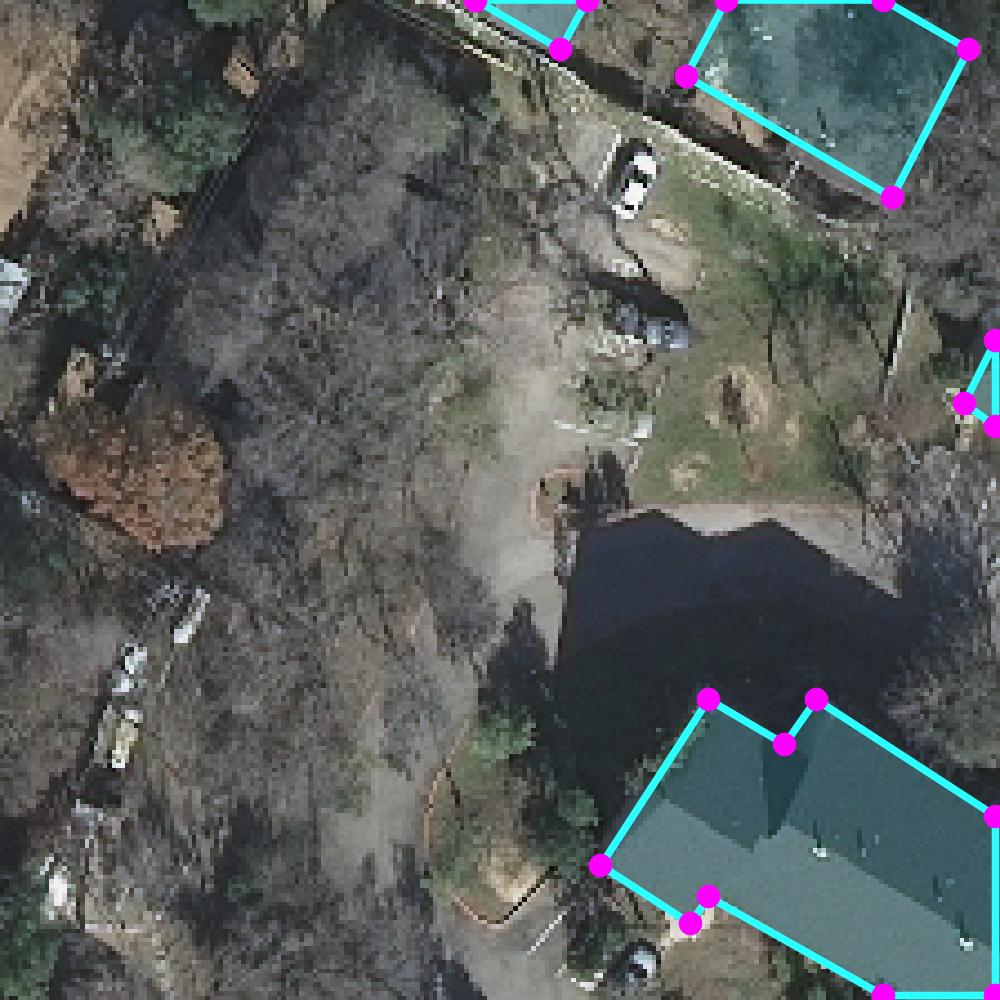} \\

\makebox[\datasetcompimgwidth][c]{Ground truth} &
\makebox[\datasetcompimgwidth][c]{Trained on WHU} &
\makebox[\datasetcompimgwidth][c]{Trained on \dsname}\\
\end{tabular}

        \caption{Inria~\citep{inria_dataset}}\label{fig:dataset_comparison_inria}
    \end{subfigure}
    \begin{subfigure}[t]{0.49\textwidth}
        \centering
        \newcommand{\datasetcompimgwidth}{0.32\linewidth}
\newcommand{\datasetcompcolsep}{1mm}
\begin{tabular}{@{}c@{\hspace{\datasetcompcolsep}}c@{\hspace{\datasetcompcolsep}}c@{}}
\scriptsize

\includegraphics[width=\datasetcompimgwidth,mytrim]{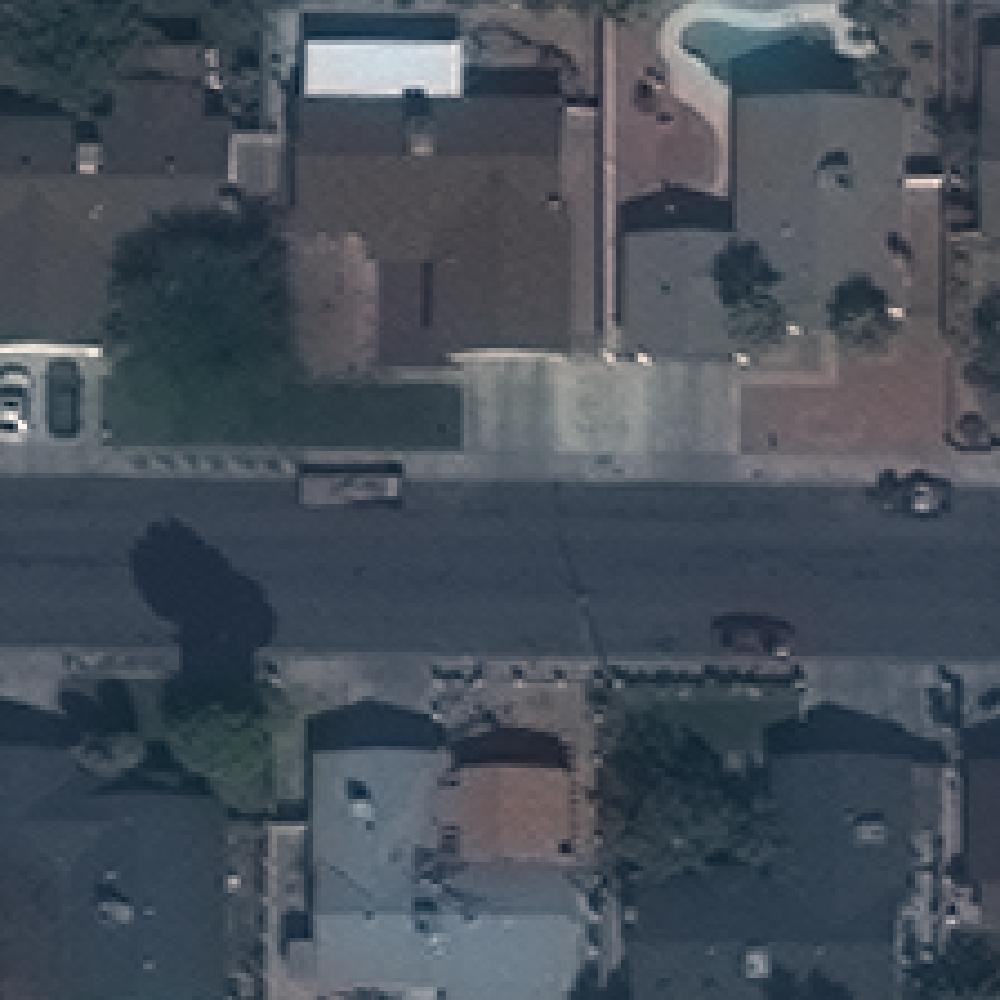} &
\includegraphics[width=\datasetcompimgwidth,mytrim]{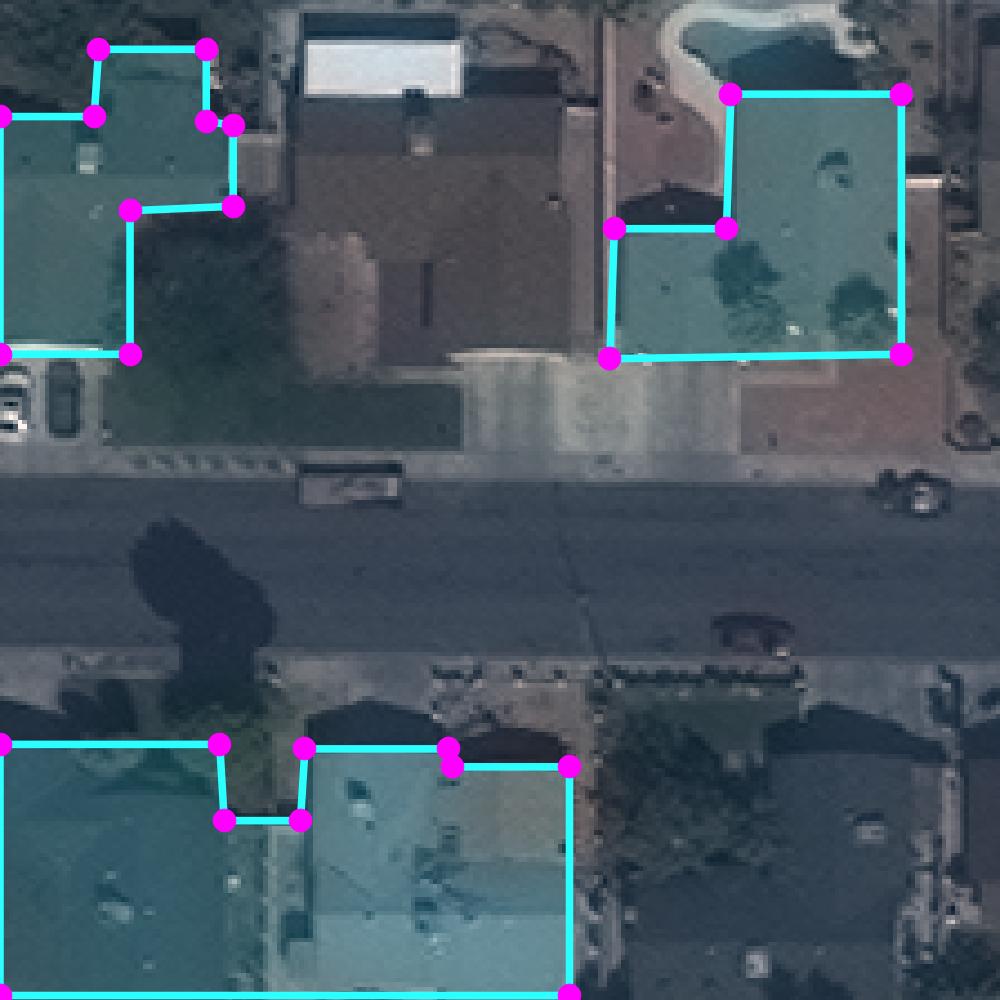} &
\includegraphics[width=\datasetcompimgwidth,mytrim]{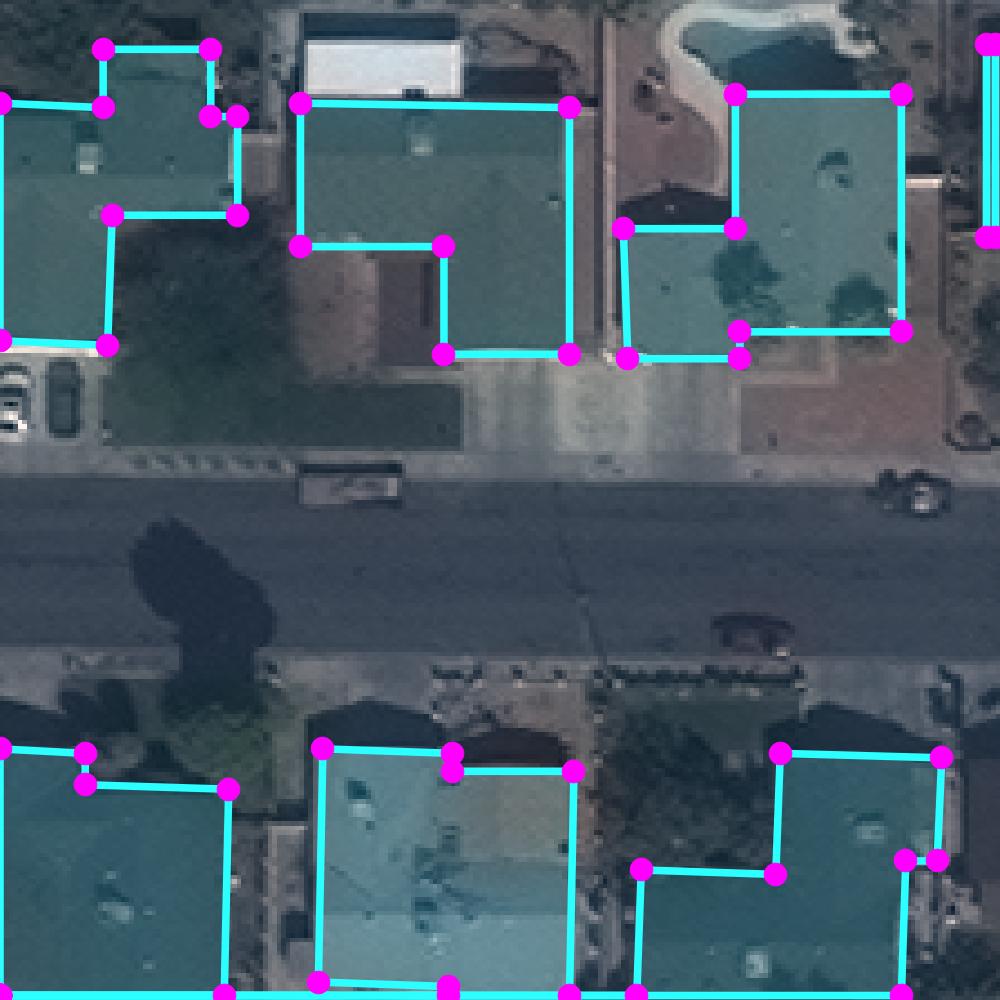} \\

\includegraphics[width=\datasetcompimgwidth,mytrim]{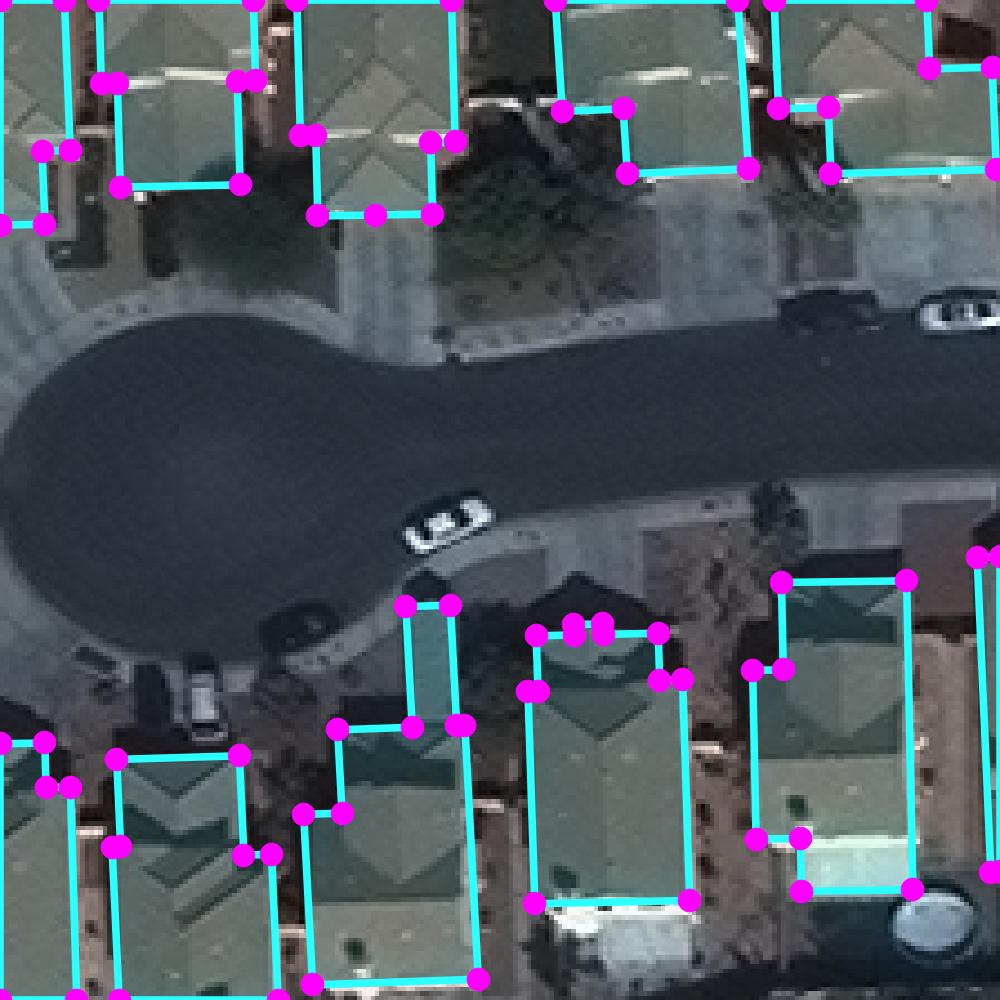} &
\includegraphics[width=\datasetcompimgwidth,mytrim]{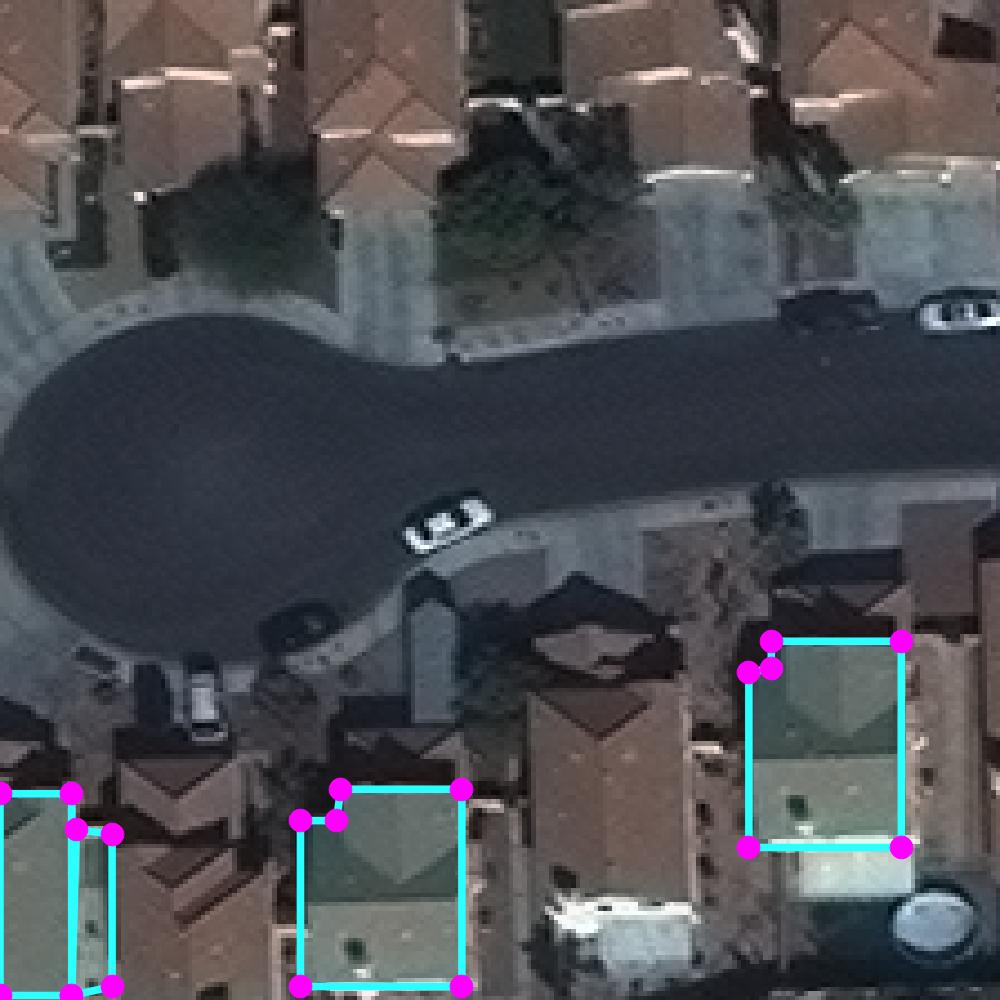} &
\includegraphics[width=\datasetcompimgwidth,mytrim]{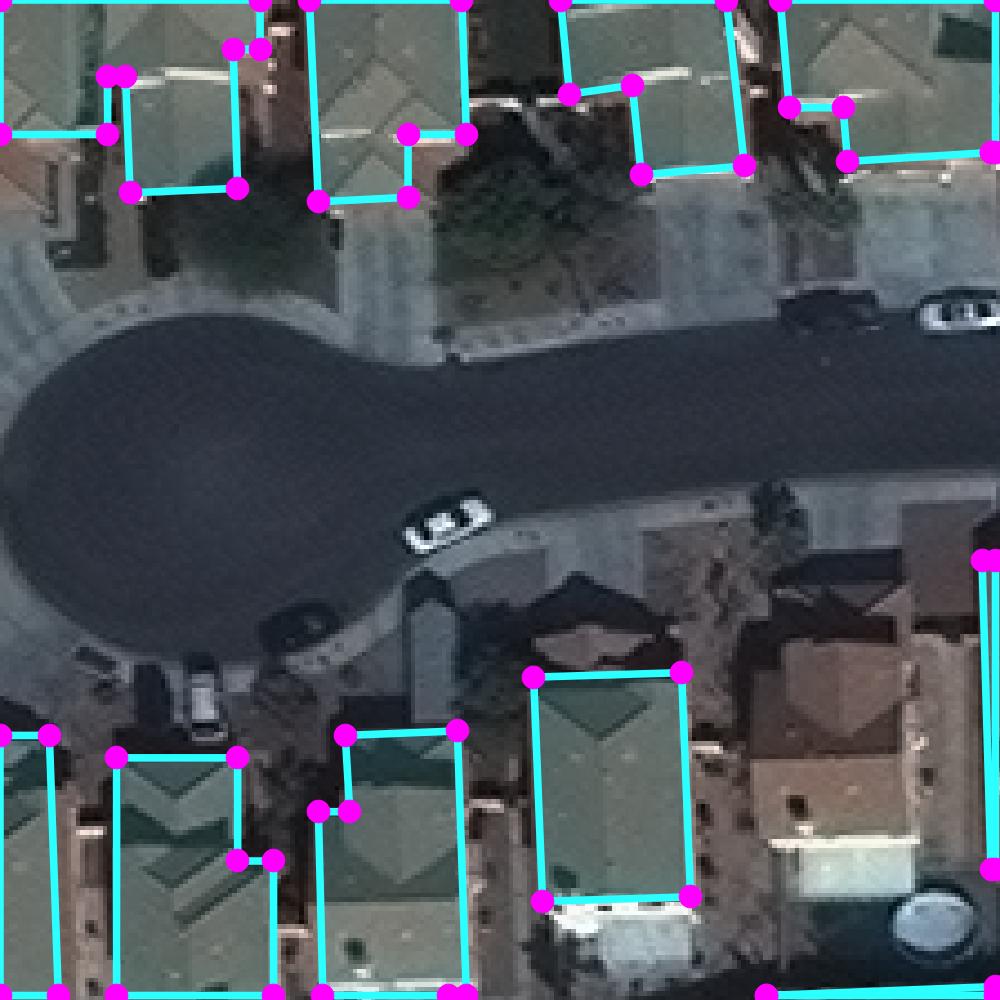} \\

\makebox[\datasetcompimgwidth][c]{Ground truth} &
\makebox[\datasetcompimgwidth][c]{Trained on WHU} &
\makebox[\datasetcompimgwidth][c]{Trained on \dsname}\\
\end{tabular}

        \caption{SpaceNet2~\citep{spacenet}}\label{fig:dataset_comparison_spacenet2}
    \end{subfigure}

    \caption{\textbf{Related dataset comparison.} We train two Pix2Poly models, one on the WHU and one on the \dsname dataset. We then predict polygons on the WHU (\ref{fig:dataset_comparison_whu}), P$^3$ (\ref{fig:dataset_comparison_p3}), Inria (\ref{fig:dataset_comparison_inria}), and SpaceNet2 (\ref{fig:dataset_comparison_spacenet2}) test sets. The polygons predicted by the \dsname trained model are generally more accurate, more complete and more regular on the out-of-distribution data.}\label{fig:dataset_comparison}
\end{figure}

\paragraph{WHU} {The model which was trained on the WHU dataset achieves a high IoU score of 94.1\% on the WHU test set, while the model trained on \dsname achieves a lower IoU score of 81.3\%.
\figref{fig:dataset_comparison_whu} shows the differences between the predictions of the two models on the WHU dataset. The model trained on the WHU dataset generally predicts the full outline of the building perimeter, as well as outlines of the two shipping containers. The model trained on \dsname predicts slighlty smaller polygons and does not predict polygons on the shipping containers.}

\paragraph{P3} {On the \dsname test set, the model trained on \dsname achieves an IoU score of 69.5\%, while the model trained on the WHU dataset achieves a lower IoU score of 65.3\%. \figref{fig:dataset_comparison_p3} shows that the model trained on WHU missing all or parts of the buildings in the example images while the model trained on \dsname predicts accurate and complete polygons.}

\paragraph{Inria} {On the Inria dataset, the model trained on \dsname achieves an IoU score of 70.8\%, while the model trained on the WHU dataset achieves a lower IoU score of 66.5\%. \figref{fig:dataset_comparison_inria} shows that the model trained on \dsname predicts more complete and also more regular polygons compared to the model trained on the WHU dataset.}

\paragraph{SpaceNet2} {On the satellite image dataset SpaceNet2, both models achieve relatively low IoU scores. This can be attributed to the fact that the satellite images have a lower resolution and a different perspective compared to the aerial images of ours and the WHU dataset. Additionally, as shown in \figref{fig:dataset_comparison_spacenet2}, some of the ground truth building polygons in the SpaceNet2 dataset are missing, which falsly contributes to the low IoU scores. Nonetheless, \figref{fig:dataset_comparison_spacenet2} shows that the model trained on \dsname is able to predict an almost complete set of building polygons.}

\section{Discussion, limitations and future work}
\label{sec:conclusion}

\paragraph{Multimodal polygon prediction.}

In our experiments, we show that building outline prediction is still a challenging problem to date. New methods are developing rapidly, and there is still room for improvement in terms of accuracy, efficiency and geometric quality of the predicted polygons. 

Traditional CNN-based methods produce complete, but complex polygons with redundant vertices or inaccurate boundaries. Furthermore, the predicted polygons usually do not include the regularity and symmetry often found in the building stock. 

The novel autoregressive sequence prediction method Pix2Poly produces simpler and more regular polygons, but sometimes misses buildings or parts of buildings entirely. Furthermore, Pix2Poly is significantly slower at inference time compared to the CNN-based methods. Another remaining challenge for end-to-end polygon prediction is the handling of buildings split over multiple tiles. Current sequence prediction methods do not support overlapping strategies, which limits their applicability in dense urban areas.

Future work for building polygon prediction should focus on regularity priors and efficient and boundary free end-to-end polygon prediction methods to allow true applicability on real-world scenarios at scale. Our dataset provides a solid foundation for future research in this direction, \eg by providing large-scale continuously annotated regions including diverse training samples with various building styles and urban layouts.

Furthermore, our experiments show that multimodal building polygon prediction is a promising research direction. The combination of aerial images and LiDAR point clouds leads to more accurate and complete building polygons compared to unimodal predictions. However, we also observe that the two modalities can sometimes provide conflicting information, \eg due to temporal differences in data acquisition or occlusions in the image modality. Future work could address this issue by developing models that can learn to handle such inconsistencies leveraging our multimodal dataset. 

\paragraph{Dataset.}

Currently, our dataset includes data from Switzerland, the USA and New Zealand. Aerial LiDAR data remains costly to acquire, and global coverage is still limited. Unfortunately, many of the least developed countries are not yet represented in available datasets. However, our dataset already includes a variety of urban landscapes in developed countries, such as residential, industrial, historical, and modern downtown areas. Furthermore, our dataset is easy to extend once new data becomes available. We are already in contact with national mapping agencies to expand our dataset to other countries in the near future.


Another limitation of our dataset is that it only includes building outlines, and does not provide 3D annotations. 
While 3D building reconstruction using LiDAR and aerial imagery represents an important long-term goal, high-quality 2D building polygon extraction already has substantial and immediate practical value while not being fully solved. Manual quality assurance and control is still required to correct errors and to ensure a certain level of fidelity  (see \eg \figureref{fig:modality_ablation_sm1}). Applications for 2D building polygons include settlement mapping, urban change monitoring, and post-disaster damage assessment. Furthermore, we maintain that most large-scale 3D building reconstruction from LiDAR require 2D polygons as input \citep{simplicity,Liu_2025_CVPR} and thus accurate 2D building polygons provide a strong foundation for large-scale 3D building reconstruction. As such, our focus on 2D annotations represents a necessary and impactful step toward achieving robust 3D building models. Furthermore, our dataset is partially composed of manual annotations. Manually generating 3D annotations is far more time consuming and such annotations are only available for small regions, \eg on city scale. 
Creating 3D annotations in a reliable manner is a challenge that we plan to tackle in the future to increase the application spectrum of our dataset.

Another application for future work could be the use our dataset for other tasks beyond building polygon extraction, such as multimodal change detection or depth estimation.

\section{Conclusion}

We present a new dataset and benchmark for large-scale building vectorization from high resolution aerial images and LiDAR pointclouds.
We show that (i) LiDAR point clouds are a valuable source for building outline prediction, that (ii) building outline prediction is challenging enough to require multimodal approaches (iii) and that current methods still have limitations either in terms of efficiency or regularity of predicted polygons.
We hope that P$^3$ jumpstarts research in architectures capable of using both aerial imagery and LiDAR data for precise mapping. To this end, we also introduce a baseline method for early fusion and add three different approaches to the leaderboard. Our approach shows that the combination of image and LiDAR point clouds leads to more accurate and complete polygons compared to unimodal predictions. We believe our baseline is a step forward in multimodal building polygon prediction and we hope our dataset will serve as a valuable resource for researchers aiming to further improve performance on this task.

\begin{figure}[t]
    \centering
	\definetrim{mytrim}{0 0 0 0}
	\newcommand{\mywidth}{0.25\linewidth}
	\newcommand{\myfontsize}{\scriptsize}
	\setlength{\tabcolsep}{0mm}
    \newcommand{\baseimg}{./images/all_countries_figure/sm}
    \newcommand{\imgpath}[1]{\baseimg/#1} 

\newcommand{\inputImages}[4]{%
    &
    \includegraphics[width=\mywidth,mytrim]{\imgpath{#1_test_#2_#3_#4.jpg}} &
    \includegraphics[width=\mywidth,mytrim]{\imgpath{#1_test_#2_pred_lidar.jpg}} &
    \includegraphics[width=\mywidth,mytrim]{\imgpath{#1_test_#2_pred_both.jpg}} &
    \includegraphics[width=\mywidth,mytrim]{\imgpath{#1_test_#2_gt.jpg}} \\
}
\newcommand{\inputPred}[4]{%
    \includegraphics[width=\mywidth,mytrim]{\imgpath{#1_test_#2_#3_#4.jpg}}
}
\newcommand{\inputGt}[3]{%
    \includegraphics[width=\mywidth,mytrim]{\imgpath{#1_test_#2_gt_#3}}
}

\newcommand{\tileIDa}{7886}
\newcommand{\tileIDb}{19926}
\newcommand{\tileIDc}{29553}

\centering
\resizebox{0.98\textwidth}{!}{

\begin{tabular}{@{}l@{\hspace{3pt}}cccc@{}}

\rotatebox{90}{\hspace{13mm}Switzerland} &
\inputGt{all}{\tileIDa}{both} &
\inputPred{all}{\tileIDa}{ffl}{v0_all_bs4x16} &
\inputPred{all}{\tileIDa}{hisup}{v0_all_bs4x16} &
\inputPred{all}{\tileIDa}{pix2poly}{v0_all_bs4x16} \\

\rotatebox{90}{\hspace{12mm}New Zealand} &
\inputGt{all}{\tileIDb}{both} &
\inputPred{all}{\tileIDb}{ffl}{v0_all_bs4x16} &
\inputPred{all}{\tileIDb}{hisup}{v0_all_bs4x16} &
\inputPred{all}{\tileIDb}{pix2poly}{v0_all_bs4x16} \\

\rotatebox{90}{\hspace{17mm}USA} &
\inputGt{all}{\tileIDc}{both} &
\inputPred{all}{\tileIDc}{ffl}{v0_all_bs4x16} &
\inputPred{all}{\tileIDc}{hisup}{v0_all_bs4x16} &
\inputPred{all}{\tileIDc}{pix2poly}{v0_all_bs4x16} \\

&\makebox[\mywidth][c]{Ground truth} &
\makebox[\mywidth][c]{FFL \citep{ffl}} &
\makebox[\mywidth][c]{HiSup \citep{hisup}} &
\makebox[\mywidth][c]{Pix2Poly \citep{pix2poly}}\\

\end{tabular}
}

\caption{\textbf{Multimodal polygon prediction 2.} We show ground truth and predicted building outlines of P$^3$, for Swizerland, New Zealand\protect\footnotemark[1]~and the USA (from top to bottom), with LiDAR point clouds superimposed on aerial images. 
The first column shows the ground truth reference polygons, while the second to forth column show polygons predicted using both input modalities. In these three challenging examples, none of the tested methods can produce accurate building outlines.}
\label{fig:all_countries_sm1}
\end{figure}

\begin{figure}[t]
    \centering
	\definetrim{mytrim}{0 0 0 0}
	\newcommand{\mywidth}{0.25\linewidth}
	\newcommand{\myfontsize}{\scriptsize}
	\setlength{\tabcolsep}{0mm}
    \newcommand{\baseimg}{./images/all_countries_figure/sm}
    \newcommand{\imgpath}[1]{\baseimg/#1} 

\newcommand{\inputImages}[4]{%
    &
    \includegraphics[width=\mywidth,mytrim]{\imgpath{#1_test_#2_#3_#4.jpg}} &
    \includegraphics[width=\mywidth,mytrim]{\imgpath{#1_test_#2_pred_lidar.jpg}} &
    \includegraphics[width=\mywidth,mytrim]{\imgpath{#1_test_#2_pred_both.jpg}} &
    \includegraphics[width=\mywidth,mytrim]{\imgpath{#1_test_#2_gt.jpg}} \\
}
\newcommand{\inputPred}[4]{%
    \includegraphics[width=\mywidth,mytrim]{\imgpath{#1_test_#2_#3_#4.jpg}}
}
\newcommand{\inputGt}[3]{%
    \includegraphics[width=\mywidth,mytrim]{\imgpath{#1_test_#2_gt_#3}}
}

\newcommand{\tileIDa}{9042}
\newcommand{\tileIDb}{16495}
\newcommand{\tileIDc}{30131}

\centering
\resizebox{0.98\textwidth}{!}{

\begin{tabular}{@{}l@{\hspace{3pt}}cccc@{}}

\rotatebox{90}{\hspace{9mm}Switzerland} &
\inputGt{all}{\tileIDa}{both} &
\inputPred{all}{\tileIDa}{ffl}{v0_all_bs4x16} &
\inputPred{all}{\tileIDa}{hisup}{v0_all_bs4x16} &
\inputPred{all}{\tileIDa}{pix2poly}{v0_all_bs4x16} \\

\rotatebox{90}{\hspace{8mm}New Zealand} &
\inputGt{all}{\tileIDb}{both} &
\inputPred{all}{\tileIDb}{ffl}{v0_all_bs4x16} &
\inputPred{all}{\tileIDb}{hisup}{v0_all_bs4x16} &
\inputPred{all}{\tileIDb}{pix2poly}{v0_all_bs4x16} \\

\rotatebox{90}{\hspace{14mm}USA} &
\inputGt{all}{\tileIDc}{both} &
\inputPred{all}{\tileIDc}{ffl}{v0_all_bs4x16} &
\inputPred{all}{\tileIDc}{hisup}{v0_all_bs4x16} &
\inputPred{all}{\tileIDc}{pix2poly}{v0_all_bs4x16} \\

&\makebox[\mywidth][c]{Ground truth} &
\makebox[\mywidth][c]{FFL \citep{ffl}} &
\makebox[\mywidth][c]{HiSup \citep{hisup}} &
\makebox[\mywidth][c]{Pix2Poly \citep{pix2poly}}\\

\end{tabular}
}
\caption{\textbf{Multimodal polygon prediction 3.} We show ground truth and predicted building outlines of P$^3$, for Swizerland, New Zealand\protect\footnotemark[1]~and the USA (from top to bottom), with LiDAR point clouds superimposed on aerial images. 
The first column shows the ground truth reference polygons, while the second to forth column show polygons predicted using both input modalities. The two CNN-based methods HiSup and FFL produce a more complete polygon set compared to Pix2Poly, which struggles to capture all buildings and interior cutouts.}
\label{fig:all_countries_sm2}
\end{figure}

\begin{figure}
\centering

	\definetrim{mytrim}{0 0 0 0}
	\newcommand{\mywidth}{0.25\linewidth}
	\newcommand{\myfontsize}{\scriptsize}
	\setlength{\tabcolsep}{0mm}
    \newcommand{\baseimg}{./images/modality_ablation_figure}
    \newcommand{\imgpath}[1]{\baseimg/#1} 

\newcommand{\inputImages}[4]{%
    &
    \includegraphics[width=\mywidth,mytrim]{\imgpath{#1_test_#2_#3_#4.jpg}} &
    \includegraphics[width=\mywidth,mytrim]{\imgpath{#1_test_#2_pred_lidar.jpg}} &
    \includegraphics[width=\mywidth,mytrim]{\imgpath{#1_test_#2_pred_both.jpg}} &
    \includegraphics[width=\mywidth,mytrim]{\imgpath{#1_test_#2_gt.jpg}} \\
}
\newcommand{\inputPred}[4]{%
    \includegraphics[width=\mywidth,mytrim]{\imgpath{#1_test_#2_#3_#4.jpg}}
}
\newcommand{\inputGt}[3]{%
    \includegraphics[width=\mywidth,mytrim]{\imgpath{#1_test_#2_gt_#3}}
}

    \newcommand{\tileID}{1615}

    \resizebox{0.98\textwidth}{!}{%
        \begin{tabular}{@{}c@{}c@{}c@{}c@{\hspace{2pt}}l@{}}

    \inputGt{Switzerland}{\tileID}{image} &
    \inputPred{Switzerland}{\tileID}{ffl}{v4_image_bs4x16} &
    \inputPred{Switzerland}{\tileID}{hisup}{v3_image_vit_cnn_bs4x12} &
    \inputPred{Switzerland}{\tileID}{pix2poly}{v4_image_vit_bs4x16} &
    \rotatebox{-90}{\hspace{-25mm}Pred. Image} \\
    
    \inputGt{Switzerland}{\tileID}{lidar} &
    \inputPred{Switzerland}{\tileID}{ffl}{v5_lidar_bs2x16_mnv64} &
    \inputPred{Switzerland}{\tileID}{hisup}{lidar_pp_vit_cnn_bs2x16_mnv64} &
    \inputPred{Switzerland}{\tileID}{pix2poly}{lidar_pp_vit_bs2x16_mnv64} &
    \rotatebox{-90}{\hspace{-25mm}Pred. LiDAR} \\
    
    \inputGt{Switzerland}{\tileID}{both} &
    \inputPred{Switzerland}{\tileID}{ffl}{v4_fusion_bs4x16_mnv64} &
    \inputPred{Switzerland}{\tileID}{hisup}{early_fusion_vit_cnn_bs2x16_mnv64} &
    \inputPred{Switzerland}{\tileID}{pix2poly}{early_fusion_bs2x16_mnv64} &
    \rotatebox{-90}{\hspace{-25mm}Pred. Fusion} \\

        \makebox[\mywidth][c]{Ground truth} &
        \makebox[\mywidth][c]{FFL \citep{ffl}} &
        \makebox[\mywidth][c]{HiSup \citep{hisup}} &
        \makebox[\mywidth][c]{Pix2Poly \citep{pix2poly}} & \\

\end{tabular}
    }
\caption{\textbf{Modality ablation 2.} We show an example tile of predicted and ground truth building polygons from the Switzerland subset. The first column shows ground truth polygons, and following columns show predicted polygons of the baseline models trained on different modalities, \ie images only (first row), LiDAR only (second row), and both image and LiDAR data (third row). The tile shows a complex building due to a cutout and open roof structure. FFL and HiSup struggle to predict the building outline, while Pix2Poly is able to predict the building outline almost perfectly using image and LiDAR data.}
\label{fig:modality_ablation_sm1}
\end{figure}

\begin{figure}
\centering
	\definetrim{mytrim}{0 0 0 0}
	\newcommand{\mywidth}{0.25\linewidth}
	\newcommand{\myfontsize}{\scriptsize}
	\setlength{\tabcolsep}{0mm}
    \newcommand{\baseimg}{./images/modality_ablation_figure}
    \newcommand{\imgpath}[1]{\baseimg/#1} 

\newcommand{\inputImages}[4]{%
    &
    \includegraphics[width=\mywidth,mytrim]{\imgpath{#1_test_#2_#3_#4.jpg}} &
    \includegraphics[width=\mywidth,mytrim]{\imgpath{#1_test_#2_pred_lidar.jpg}} &
    \includegraphics[width=\mywidth,mytrim]{\imgpath{#1_test_#2_pred_both.jpg}} &
    \includegraphics[width=\mywidth,mytrim]{\imgpath{#1_test_#2_gt.jpg}} \\
}
\newcommand{\inputPred}[4]{%
    \includegraphics[width=\mywidth,mytrim]{\imgpath{#1_test_#2_#3_#4.jpg}}
}
\newcommand{\inputGt}[3]{%
    \includegraphics[width=\mywidth,mytrim]{\imgpath{#1_test_#2_gt_#3}}
}

    \newcommand{\tileID}{1274}

    \resizebox{0.98\textwidth}{!}{%
        \begin{tabular}{@{}c@{}c@{}c@{}c@{\hspace{2pt}}l@{}}

    \inputGt{Switzerland}{\tileID}{image} &
    \inputPred{Switzerland}{\tileID}{ffl}{v4_image_bs4x16} &
    \inputPred{Switzerland}{\tileID}{hisup}{v3_image_vit_cnn_bs4x12} &
    \inputPred{Switzerland}{\tileID}{pix2poly}{v4_image_vit_bs4x16} &
    \rotatebox{-90}{\hspace{-25mm}Pred. Image} \\
    
    \inputGt{Switzerland}{\tileID}{lidar} &
    \inputPred{Switzerland}{\tileID}{ffl}{v5_lidar_bs2x16_mnv64} &
    \inputPred{Switzerland}{\tileID}{hisup}{lidar_pp_vit_cnn_bs2x16_mnv64} &
    \inputPred{Switzerland}{\tileID}{pix2poly}{lidar_pp_vit_bs2x16_mnv64} &
    \rotatebox{-90}{\hspace{-25mm}Pred. LiDAR} \\
    
    \inputGt{Switzerland}{\tileID}{both} &
    \inputPred{Switzerland}{\tileID}{ffl}{v4_fusion_bs4x16_mnv64} &
    \inputPred{Switzerland}{\tileID}{hisup}{early_fusion_vit_cnn_bs2x16_mnv64} &
    \inputPred{Switzerland}{\tileID}{pix2poly}{early_fusion_bs2x16_mnv64} &
    \rotatebox{-90}{\hspace{-25mm}Pred. Fusion} \\

        \makebox[\mywidth][c]{Ground truth} &
        \makebox[\mywidth][c]{FFL \citep{ffl}} &
        \makebox[\mywidth][c]{HiSup \citep{hisup}} &
        \makebox[\mywidth][c]{Pix2Poly \citep{pix2poly}} & \\

\end{tabular}
    }
\caption{\textbf{Modality ablation 3.} We show an example tile of predicted and ground truth building polygons from the Switzerland subset. The first column shows ground truth polygons, and following columns show predicted polygons of the baseline models trained on different modalities, \ie images only (first row), LiDAR only (second row), and both image and LiDAR data (third row). All models produce the best polygon predictions when trained on both modalities.}
\label{fig:modality_ablation_sm2}
\end{figure}

\FloatBarrier



\appendix

\section{Evaluation metrics}
\label{sm:evaluation_metrics}

The evaluation metrics used in our benchmark are defined as follows.

\subsection{Intersection over union (IoU)}

The \ac{iou} metric is defined as the ratio between the intersection and the union of the predicted and ground truth polygons. Let $P^{g}$ be the set of ground truth polygons and $P^{d}$ the set of predicted polygons of a single tile, then the \ac{iou} is defined as: 
\begin{equation}
    \text{IoU}(P^{g}, P^{d}) = \frac{|P^{g} \cap P^{d}|}{|P^{g} \cup P^{d}|} .
\end{equation}

\subsection{Average precision (AP) and recall (AR)}

The average precision (AP) and average recall (AR) metrics are computed using the \ac{iou} metric averaged over different thresholds \citep{mscoco}. For a fixed IoU threshold $\tau$, a predicted polygon $p^d$ is considered a true positive (TP) if it matches a ground truth polygon $p^g$ with $\mathrm{IoU} \geq \tau$. Otherwise, it is a false positive (FP). Ground truth polygons not matched by any prediction are counted as false negatives (FN). Then, precision and recall are defined as

\begin{align}
    \mathrm{Precision} &= \frac{\mathrm{TP}}{\mathrm{TP} + \mathrm{FP}}
\end{align}
and
\begin{align}
    \mathrm{Recall} &= \frac{\mathrm{TP}}{\mathrm{TP} + \mathrm{FN}} .
\end{align}

AP and AR are computed and averaged over 10 IoU thresholds $\tau \in \{0.50, 0.55, ..., 0.95\}$ as

\begin{equation}
    \mathrm{AP} = \frac{1}{10} \sum_{\tau} \mathrm{Precision}_{\tau}
\end{equation}
and
\begin{equation}
    \mathrm{AR} = \frac{1}{10} \sum_{\tau} \mathrm{Recall}_{\tau} .
\end{equation}


\subsection{POLIS}
The POLIS metric \citep{polis} is computed for ground truth and predicted polygons $p^g$ and $p^d$ matched with a minimum \ac{iou} threshold of 0.5.
Let $V$ be the vertices of a polygon $p$. Then, the POLIS metric is defined as the symmetric average distance between each predicted vertex $v^{d} \in p^d$ 
and its closest point $x^{g}$ on boundary of the matched ground truth polygon $\partial p^{g}$, and vice versa, \ie  
\begin{equation}
    \begin{aligned}
    {\rm POLIS}(p^{g}, p^{d}) =& \frac{1}{2|v^{g}|}\sum_{v^{g}\in p^{g}}\min_{a\in \partial p^{d}}||v^{g}-a||
        & +\frac{1}{2|v^{d}|}\sum_{v^{d}\in p_{d}}\min_{b\in \partial p^{g}}||v^{d}-b|| .
    \end{aligned}
\end{equation}

\subsection{Chamfer distance (CD)}

The Chamfer distance (CD) is defined similarly to the POLIS distance. However, instead of using the polygon vertices, we sample points on the polygon boundaries to approximate the boundary distance of $\partial p^{g}$ and $\partial p^{d}$. Let $S_g$ and $S_d$ be 2D point sets sampled with a maximum spacing of 10~cm on the polygon boundaries $\partial p^g$ and $\partial p^d$ for all polygons $p^g \in P^g$ and $p^d \in P^d$ of a tile. Then, the CD is defined as:

\begin{equation}
    \begin{aligned}
    {\rm CD}(S^{g}, S^{d}) =& \frac{1}{2|S^{g}|}\sum_{s^{g}\in S^{g}}\min_{s^{d}\in S^{d}}||s^{g}-s^{d}||_2
        & +\frac{1}{2|S^{d}|}\sum_{s^{d}\in S^{d}}\min_{s^{g}\in S_{g}}||s^{d}-s^{g}||_2 .
    \end{aligned}
\end{equation}

\subsection{Hausdorff distance (HD)}

The Hausdorff distance (HD) is defined as the maximum distance between the predicted and ground truth polygon boundaries. Using the same sampled points on the polygon boundaries as before, the HD is defined as

\begin{equation}
    \begin{aligned}
    {\rm HD}(S^{g}, S^{d}) =& \max_{s^{g}\in S^{g}}\min_{s^{d}\in S^{d}}||s^{g}-s^{d}||_2
        & +\max_{s^{d}\in S^{d}}\min_{s^{g}\in S^{g}}||s^{d}-s^{g}||_2 .
    \end{aligned}
\end{equation}

\subsection{Maximum tangent angle (MTA)}
The \ac{mta}~\citep{ffl} is also computed for ground truth and predicted polygons matched with a minimum \ac{iou} threshold of 0.5. We sample an ordered sequence of points $T^d$ on the predicted polygons boundary and project it to the matched ground truth polygons, \ie for each $t^{d}$ we find the closest point $x^{g}$ on the ground truth polygons boundary. For both sequences of points $T^d$ and $X^g$, ommiting the superscript, corresponding normed tangent directions are computed as
\begin{equation}
    \begin{aligned}
\rm Tanget(t_i) = \frac{t_{i+1} - t_i}{\Vert t_{i+1} - t_i \Vert} \;\;\;\text{and}\;\;\; \rm Tangent(x_i) = \frac{x_{i+1} - x_i}{\Vert x_{i+1} - x_i \Vert} \,.
\end{aligned}
\end{equation}

The angle differences between the two are computed from the scalar product

\begin{equation}
    \begin{aligned}
\Delta\theta_i = cos^{-1}( \langle \rm Tangent(t_i), \rm Tangent(x_i) \rangle ) \,.
\end{aligned}
\end{equation}

Before computing the maximum angle error MTA, we filter out tangents whose projection is stretched more than a factor of 2, \eg due to mismatches between ground truth and predicted segments. 
We keep all $\Delta\theta_j, \forall j \in A$ where $A = \{j \;|\; j \in [1 .. n] , \frac{1}{2} < \frac{\Vert x_{i+1} - x_i \Vert}{\Vert t^{i+1} - t_i \Vert} < 2 \} $. The final maximum tangent angle error is then computed as

\begin{equation}
    \begin{aligned}
\rm MTA = \max_{j \in A} \Delta\theta_j \,.
\end{aligned}
\end{equation}

\subsection{Vertex number ration (NR)}

The relative difference of predicted and ground truth number of vertices $\left|V^d\right|$ and $\left|V^g\right|$ per tile is defined as
\begin{equation}
    \begin{aligned}
\rm NR = \frac{\left| \left| V^d \right| - \left| V^g \right| \right|}{\left| V^d \right| + \left| V^g \right|}
\end{aligned}
\end{equation}

\subsection{Complexity IoU (C-IoU)}

The complexity \ac{iou} (C-IoU) is defined as the product of the \ac{iou} and the vertex ratio (NR): 

\begin{equation}
    \text{C-IoU} = \text{IoU} \cdot \text{NR} .
\end{equation}

\subsection{Line degree of freedom (DoF)}

The line-degree of freedom score (DoF) \citep{lineFit_eccv2024} is defined as the ratio of the degree of freedom $k_{L}$ of a configuration of line segments $L$ to twice the number of lines $|L|$ in the configuration.

\begin{equation}
\text{DoF}(L)=\frac{k_L}{2|L|}
\end{equation}

The degree of freedom $k_L$ is computed via the counting of the number of line-segments in different regularization groups, \ie for parallelism, orthogonality and co-linearity as regularities, the degree of freedom is given by
\begin{equation}
 k_L= 2|L|  - \sum_{i=1}^{K^{\parallel,\perp}} (k_i^{\parallel,\perp} + 1) - \sum_{i=1}^{K^{\parallel}} (k_i^{\parallel} + 1) - \sum_{i=1}^{K^{--}} (k_i^{--} + 1) 
\end{equation}
where $K^{\parallel,\perp}$, $K^{\parallel}$ and $K^{--}$ correspond to the number of parallel groups which are also mutually orthogonal, the number of parallel groups which are not orthogonal to other groups, and the number of co-linear sub-groups inside all the parallel groups respectively. $k_i^{\parallel,\perp}$ is the number of line segments in the $i^{th}$ group of $K^{\parallel,\perp}$, $k_i^{\parallel}$ the number of line segments in the $i^{th}$ group of $K^{\parallel}$, and  $k_i^{--}$ the number of line segments in the $i^{th}$ group of $K^{--}$.

Having a DoF score of 1 means that all line segments of the configuration are independent and not subject to regularity constraints. Note that a DoF score of $0$ is not possible as, even in the case where all the line segments were mutually co -linear, at least one of the line-segment must be described by two parameters. In such a case, the minimal score would be $1/|L|$.

We compute the DoF score with a threshold of 0.5$^\circ$ for the predicted polygons. We filter out polygon edges on the tile boundary as they are not subject to regularity constraints of a polygon.

\end{document}